\documentclass{article} 
\usepackage{iclr2027_conference,times}

\usepackage{latexsym}
\usepackage{tcolorbox}
\usepackage{graphicx}
\usepackage{subcaption}
\usepackage{booktabs}
\usepackage{multirow}
\usepackage{cuted}
\usepackage[hidelinks]{hyperref}
\usepackage{doi}
\usepackage{amsmath}
\usepackage{mleftright}
\usepackage[capitalize,nameinlink]{cleveref}
\usepackage{etoolbox}
\robustify\bfseries

\usepackage{tikz}
\usetikzlibrary{shapes.geometric,shapes.symbols}

\usepackage{fontawesome5}

\usepackage{makecell}
\newcommand{\valci}[3]{\makecell[c]{#1\\[-1pt]{\tiny[#2, #3]}}}
\usepackage{siunitx}
\usepackage[T1]{fontenc}
\usepackage[utf8]{inputenc}
\usepackage{microtype}
\usepackage{inconsolata}
\usepackage{graphicx}
\usepackage{tipa} 

\usepackage{enumitem}
\setlist{itemsep=2pt, topsep=0pt, parsep=0pt,leftmargin=1em}

\usepackage{svg}

\newcommand{\Train}{Train}                             
\newcommand{\Val}{Val}                                 
\newcommand{\Test}{Test}                               
\newcommand{\condZ}{Zero-shot}                         
\newcommand{\condC}{Choice-only}                       
\newcommand{\condG}{+Gaze}                             
\newcommand{\condA}{+AvgGaze}                          
\newcommand{\condS}{+Gaze+Side}                        
\newcommand{\humanagr}{human-choice agreement}         
\newcommand{\catagr}{intended-category agreement}               
\newcommand{\dta}{decision-token attention}            
\newcommand{\rhoin}{within-image gaze correlation}     
\newcommand{\gazeKL}{gaze KL}                          
\newcommand{\dKL}{KL gain over train-mean}             
\newcommand{\sideJS}{between-image JS}                 
\newcommand{\cattn}{choice--attention side agreement}  
\newcommand{\gattn}{gaze--attention side agreement}    
\newcommand{\gemma}{Gemma4-E4B}                        
\newcommand{\qwen}{Qwen3-VL-8B}
\newcommand{\molmo}{Molmo2-8B}

\usepackage{newtxtext}
\usepackage{newtxmath}

\iclrfinalcopy
\title{Similar Choices, Different Attention: \\ Cross-Modal Associations \\ in Humans and Vision--Language Models}

\author{Sumin Hong, Katsumi Ibaraki\thanks{Equal contribution.} \\
University of Notre Dame\\
\texttt{\{shong6,kibaraki\}@nd.edu}
\\
\And
Renee Shi\\
University of Notre Dame\\
\texttt{rshi2@nd.edu}
\\
\And
David Chiang\\
University of Notre Dame\\
\texttt{dchiang@nd.edu}
\\
\And
Toby Jia-Jun Li\\
University of Notre Dame\\
\texttt{toby.j.li@nd.edu}
}

\begin{document}
\maketitle
\lhead{} 

\begin{abstract}
Cross-modal associations are systematic pairings of features across modalities, such as the association of `bouba' with round shapes and `kiki' with sharp shapes.
Prior work has compared humans and vision--language models (VLMs) on such associations, but often using different stimuli or tasks between humans and models.
Here, we ask whether VLMs align with humans not only in choices, but also in where they look when making those choices.
We study both VLMs and humans ($N=53$), presenting them with the same stimuli, a pseudo-word and two images, and record participants' choices and eye movements, which we release\footnote{Dataset: https://anonymous.4open.science/r/Sound-Symbolism-Dataset/README.md}.
We find choice alignment in a few larger VLMs, but their saliency matches human gaze less closely than a center-bias baseline, a fixed Gaussian at the center of each image.
Fine-tuning small VLMs on human choices brings their choice alignment to the level of a human majority-vote reference on unseen words and images, yet their attention still matches human gaze less closely than this baseline.
Training model attention on human gaze raises attention--gaze correlation without improving choice alignment, and a single average gaze map per image position raises it by a similar amount.
Matching human choices, or even human gaze patterns, is therefore not sufficient evidence of human-aligned cross-modal processing.

\end{abstract}


\section{Introduction}
\looseness=-1 Humans systematically map information across sensory modalities. 
A common example is the bouba--kiki effect: When asked to match the pseudo-words \textit{bouba} and \textit{kiki} to rounded and spiky shapes, participants overwhelmingly pair \textit{bouba} with the rounded shape and \textit{kiki} with the spiky shape~\citep{ramachandran2001synaesthesia}. 
This sound--shape correspondence has been replicated across languages and writing systems, and appears to reflect cross-modal correspondences between speech sounds and visual properties \citep{cwiek2022bouba}.
It emerges even in pre-literate children~\citep{maurer2006shape} and infants~\citep{ozturk2013sound},
making it a standard testbed for studying human cross-modal association. 

\looseness=-1 The rise of vision--language models (VLMs) raises a natural question: Do models trained on visual and linguistic data develop cross-modal associations comparable to those of humans? 
The bouba--kiki effect is a useful test case because it challenges the assumption that how a word sounds is arbitrary with respect to what it refers to.
If VLMs align with humans on novel pseudo-words, this would suggest that their representations capture not only lexical semantics, but also non-arbitrary associations between word form and perceived meaning.
%
Aligning with humans in their choices, however, is not enough to answer this question. 
The similar choices can rest on different attention: a model may select the round image because of its shape, the property the association concerns, or because of other properties that co-vary with roundness in the stimuli. 
Choice agreement alone cannot tell these apart.
We therefore also ask where models attend, by comparing their saliency with human gaze on the same trials.
%
%

Prior work has reached conflicting conclusions. 
\Citet{alper2023kiki} evaluated CLIP~\citep{radford2021learning} and Stable Diffusion~\citep{rombach2022high} and claimed to find cross-modal association in their representations.
\Citet{kouwenhovencross}, in contrast, reported that such associations were absent from CLIP models. 
Critically, however, prior studies often stop short of a direct, matched comparison: they evaluate humans and VLMs on different stimuli or different tasks, or analyze them at different levels.

\begin{figure}[t]
\centering
\begin{subfigure}[b]{0.3\textwidth}
    \centering
    \includegraphics[width=\linewidth]{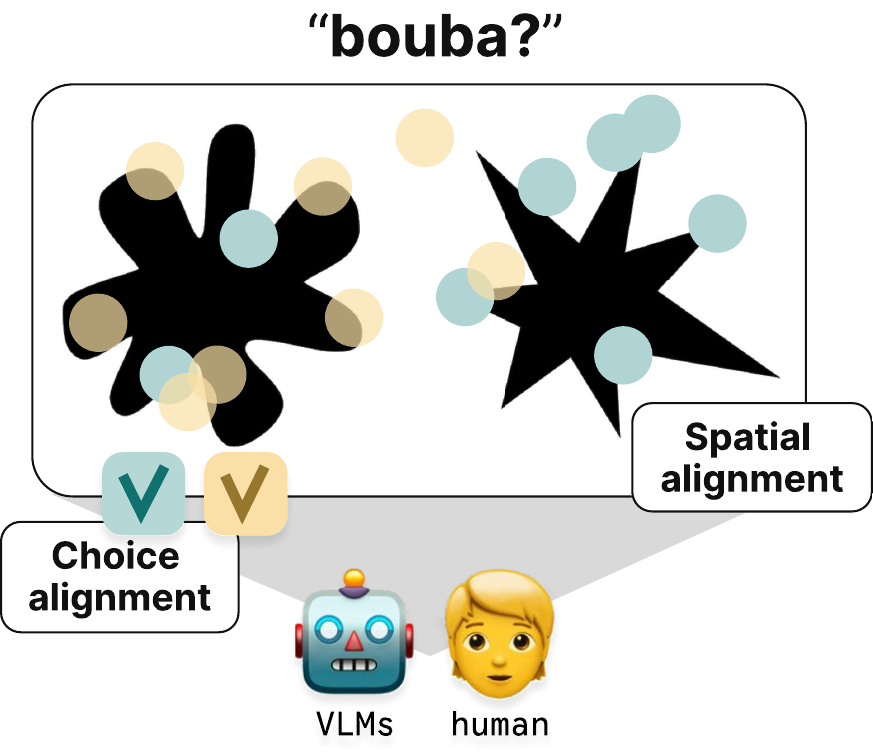}
    \caption{Overview}
    \label{fig:teaser}
\end{subfigure}
\hfill
\begin{subfigure}[b]{0.3\textwidth}
    \centering
    \includegraphics[width=0.6\linewidth]{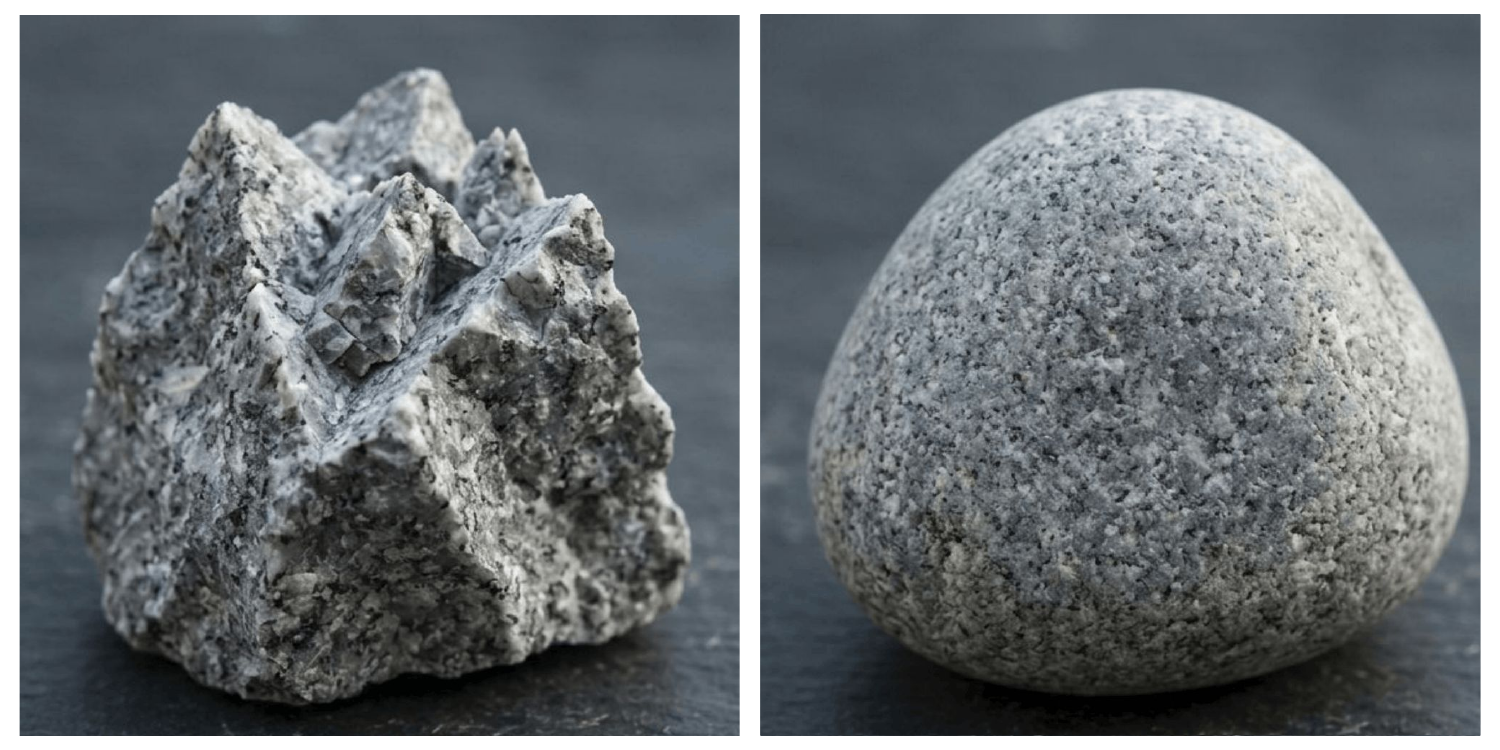}
    \\
    \includegraphics[width=0.6\linewidth]{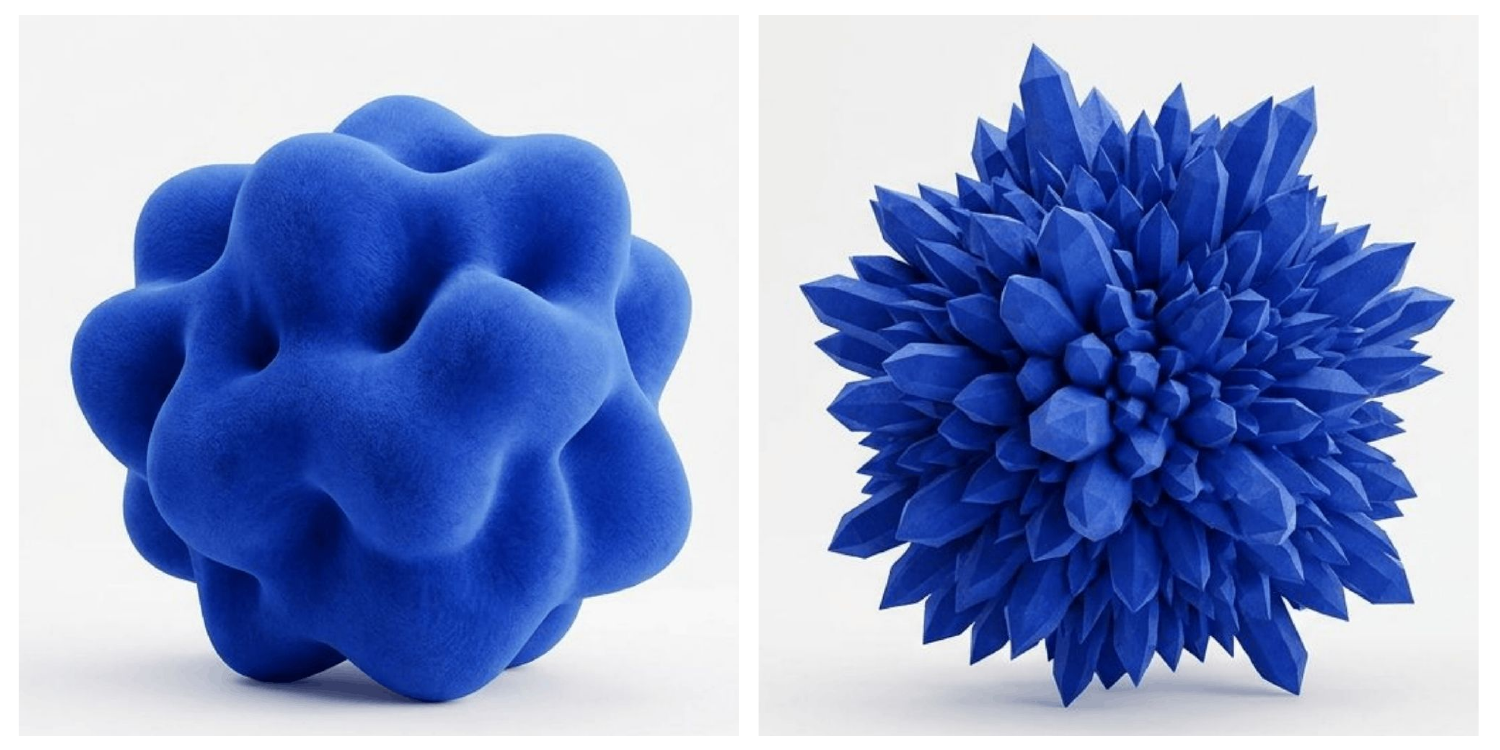}
    \caption{Gemini-generated pairs}
    \label{fig:gemini_images}
\end{subfigure}
\hfill
\begin{subfigure}[b]{0.3\textwidth}
    \centering
    \includegraphics[width=0.6\linewidth]{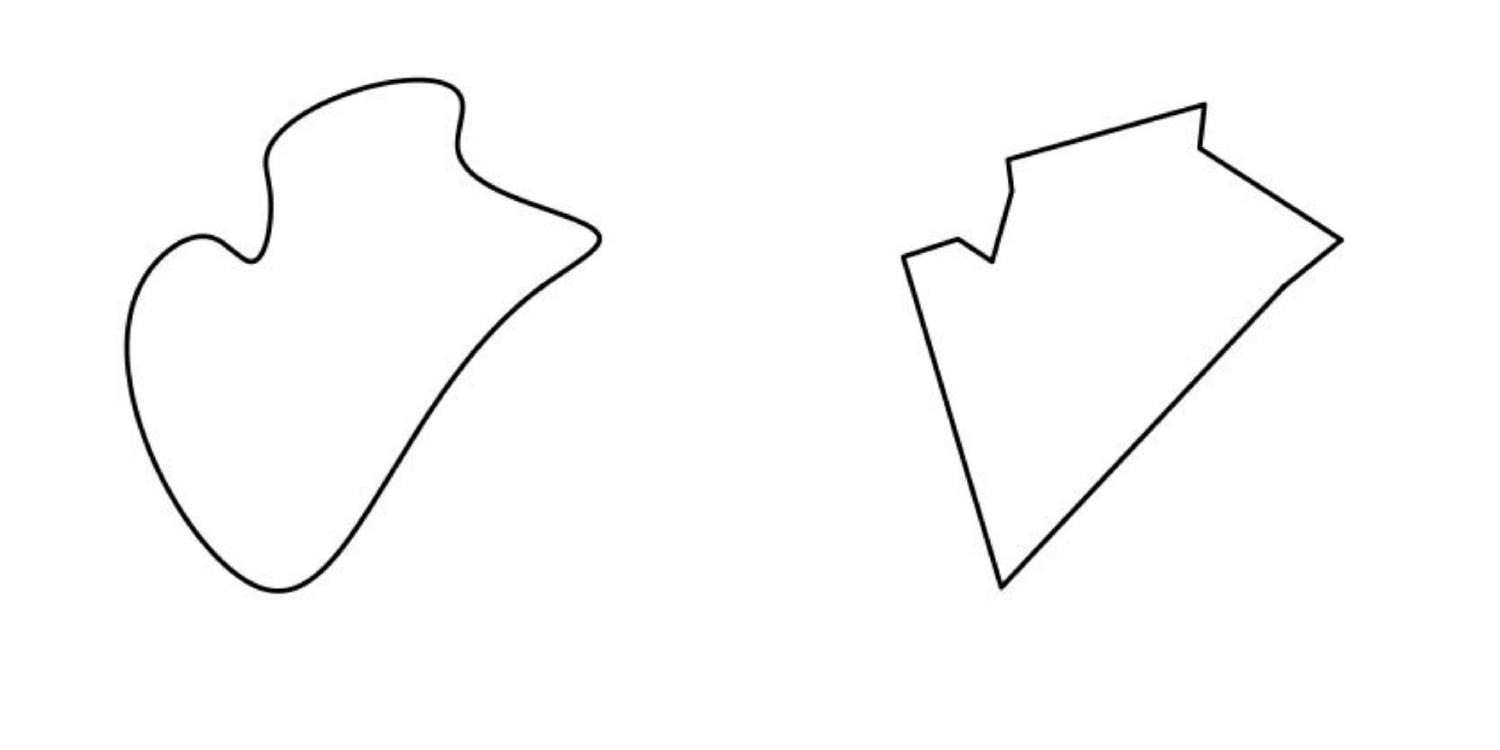}
    \\
    \includegraphics[width=0.6\linewidth]{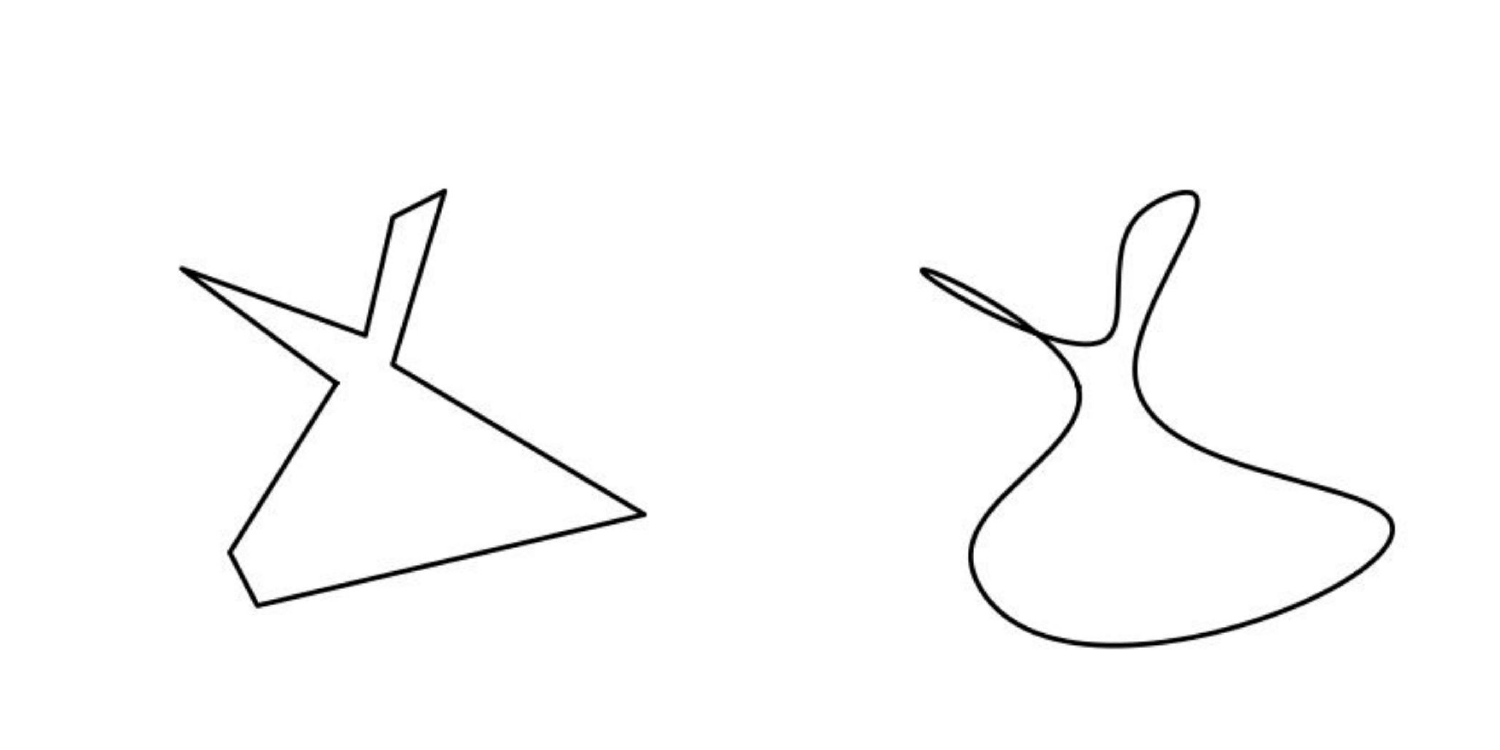}
    \caption{Line-based minimal pairs}
    \label{fig:nielsen_images}
\end{subfigure}
\caption{(a) Humans and VLMs receive the same stimulus, a pseudo-word (\emph{bouba}) and a round--sharp image pair; we compare which shape they choose (checks) and which regions they attend to (circles). Here, both choose the round shape but attend to different regions. (b, c) Example image pairs: abstract pairs generated with Gemini \citep{team2023gemini} and line-based minimal pairs following \citet{nielsen2013parsing}.}
\label{fig:main_fig}
\end{figure}

We address this gap by comparing humans and VLMs on identical stimuli (\cref{fig:teaser}), using choices and eye-tracking from 53 participants (8{,}208 trials), which we release.
We consider three questions:

\begin{itemize}
    \item \textbf{Do VLMs choose as humans do?} \textit{Choice alignment} asks whether, on trials matched to the human study, a model selects the same image. We find that it does mainly in a subset of larger VLMs; smaller models match adjectives but are near chance and side-locked on pseudo-words.
    \item \textbf{Do they attend where humans look?} \textit{Spatial alignment} compares VLM saliency (attention maps or Grad-CAM~\citep{selvaraju2020grad}) with the same participant's gaze on the same trial, using gaze as a reference for where people look when choosing~\citep{pani2025gazeVLMneurips,lai2020understanding, DAS201790}.
    We find that a simple center-bias baseline, a fixed Gaussian at the center of each image that uses no model and no information about the word or trial, aligns with human gaze better than the saliency of even the models whose choices track human judgments.
    \item \textbf{Are the two levels linked?} We fine-tune three small VLMs on human choices and ask whether, once their choices are human-aligned, their attention becomes more gaze-aligned, and conversely whether, once attention is more gaze-aligned, their choices become more human-aligned.
    Neither holds. 
    Training on ${\sim}2$k choices reaches human-level choice agreement that transfers to unseen words and images, but the center-bias baseline still aligns with human gaze better than the models' attention does.
    Adding human gaze as a training target for attention brings attention closer to human gaze but has no detectable improvement on agreement with human choices.
\end{itemize}

\section{Related work}
\subsection{Cross-modal associations in humans}
Word--meaning mappings are largely arbitrary \citep{hockett1960origin}, but \textit{sound symbolism} is a well-documented exception \citep{kohler1929gestalt, kohler1947gestalt}: most people match \textit{bouba} to a round shape and \textit{kiki} to a sharp one \citep{ramachandran2001synaesthesia, maurer2006shape, cwiek2022bouba}, and specific phonological properties correlate with roundness and sharpness~\citep{nielsen2013parsing,mccormick2015sound}.
Related associations exist for magnitude \citep{sapir1929study, shinohara2010cross, haynie2014sound, winter2021size} and texture \citep{greenberg1966studies, sakamoto2018bouba, kumagai2020pluripotentiality, sidhu2022higher, winter2022trilled}.


\subsection{Cross-modal associations in VLMs}





Prior work on VLMs has focused on round--sharp symbolism, with mixed conclusions.
\citet{alper2023kiki} defined a round--sharp direction in CLIP \citep{radford2021learning} embedding space from adjective embeddings and found that pseudo-words and images of the same category (round/sharp) project to the same side of this axis;
\citet{iida2024investigating} could not replicate this in Japanese VLMs, suggesting dependence on the English-based setup, including the adjective sets.
\citet{verhoef-etal-2024-kiki} and \citet{kouwenhovencross} instead compared pseudo-word probabilities per image, the latter adding multiple prompts, an adjective baseline, and Grad-CAM \citep{selvaraju2020grad} analysis, and found no consistent evidence of the association in CLIP models despite successful adjective matching.
Beyond CLIP, studies of multimodal LLMs \citep{loakman2024ears,jeong2026language} and audio--visual models \citep{tseng2024measuring} extend the models and semantic dimensions tested, with mixed findings.
Across these studies, humans and models are typically evaluated on different stimuli or tasks, and compared only at the output level; whether VLMs match humans under matched conditions, and whether they attend to the same image regions, remains open.

\section{Dataset and human study}
\label{sec:dataset}
We construct a set of pseudo-words paired with abstract round--sharp image pairs and evaluate humans and VLMs under the same forced-choice paradigm, recording each participant's choice and gaze on every trial.
Gaze data serves two purposes: it extends the comparison beyond choices to the allocation of attention (\cref{sec:zeroshot}), and provides a per-trial supervision signal (\cref{sec:finetune}). 
Throughout, labels and images have an \textit{intended} round or sharp category, based on tendencies in prior human studies rather than objective ground truth.

\subsection{Stimuli}
\label{sec:stimuli}

\paragraph{Linguistic stimuli.}
\label{sec:linguistic_stimuli}
Linguistic inputs come from three sources: the canonical pairs \textit{takete--maluma} and \textit{bouba--kiki} \citep{kohler1929gestalt, ramachandran2001synaesthesia}; the 20 English round--sharp adjectives of \citet{alper2023kiki} as a semantic baseline (Appendix~\ref{sec:app_stimuli}); and 162 constructed two-syllable pseudo-words.
Following prior human studies, in which sonorant consonants and rounded vowels are associated with roundness, and unvoiced plosive consonants and unrounded vowels with sharpness \citep{nielsen2013parsing, mccormick2015sound}, we use
$V_{\text{sharp}}=\{\text{EE},\text{AY},\text{UH}\}$,
$V_{\text{round}}=\{\text{OO},\text{OH},\text{AH}\}$,
$C_{\text{sharp}}=\{\text{T},\text{K},\text{P}\}$, and
$C_{\text{round}}=\{\text{M},\text{N},\text{L}\}$.
These patterns have also informed the stimuli used in prior VLM evaluations \citep{alper2023kiki, kouwenhovencross, verhoef-etal-2024-kiki}. 
We combine vowels and consonants only within the same intended category, yielding 18 one-syllable words (9 round and 9 sharp); combining them into two-syllable words yields 162 pseudo-words (e.g., \text{KEEKEE}, \text{MOHMAH}, \text{PUHTAY}), for 186 labels in total. 
All labels are presented in English orthography to standardize intended pronunciation, following the convention established in \citet{nielsen2013parsing}.
Participants read each label aloud before choosing (\cref{sec:procedure}), whereas models receive only the written form, so we treat the correspondence as one between written word form and shape, and we do not separate orthographic from phonological contributions, which are known to be intertwined in literate participants \citep{cuskley2017phonological}.

\paragraph{Visual stimuli.}
\label{sec:visual_stimuli}
We use 184 round--sharp image pairs: 17 from prior work \citep{kohler1929gestalt, kohler1947gestalt, maurer2006shape, ramachandran2001synaesthesia, westbury2005implicit, kouwenhovencross} (Appendix~\ref{sec:appendix_images}); 75 line-based minimal pairs generated by connecting random points with curved or straight lines \citep{nielsen2013parsing}, also used in \citet{verhoef-etal-2024-kiki} and \citet{kouwenhovencross}; and 92 more realistic abstract pairs generated with Gemini \citep{team2023gemini} (prompt in Appendix~\ref{sec:appendix_prompt}), which better resemble images VLMs encounter in training (\cref{fig:main_fig}b,c).
Every pair is prepared in both left--right orientations, and label--pair assignments and orientations are randomized per participant (Appendix~\ref{sec:app_human_study_details}), so no label is tied to an image pair or side, except for the canonical labels which keep their original images.

\subsection{Human study procedure}
\label{sec:procedure}
Following the human study structure of \citet{cwiek2022bouba}, participants saw a written label followed by a pair of images, read the label aloud, and verbally selected the better-matching image by saying \textit{A} (left) or \textit{B} (right); no audio was played. 
Gaze data was recorded with a Tobii Pro Fusion monitor-mounted eye-tracker, and screen/audio recordings are used for analysis.
The study had three sessions, each preceded by eye-tracker calibration; after each session, participants reported pseudo-words that reminded them of real words or names, and those items were excluded (Appendix~\ref{sec:app_human_study_details}).
We recruited 55 participants, and after retaining trials with a complete image presentation, a valid choice, and at least two in-image fixations with the gaze recording covering $\geq$80\% of the time the images were displayed, the cleaned dataset contains 8{,}208 trials from 53 participants (7{,}141 constructed pseudo-word, 910 adjective, 157 canonical); cleaning changed agreement rates by less than one point (Appendix~\ref{sec:appendix_participants_and_cleaning}).

\paragraph{Human choices.}
\label{sec:human_choices}
\looseness=-1 \emph{Intended-category agreement (ICA)}  is the rate at which a response matches the label's intended round--sharp category, which is based on prior human studies (e.g., \text{LOONOO} with the round image or  \text{KEETUH} with the sharp image).
ICA is highest for English adjectives (94.84\%), while canonical pseudo-words reach 84.71\%, replicating the standard bouba--kiki effect. 
Constructed pseudo-words reach 73.98\% agreement, substantially above chance, indicating reliable sound--shape associations in our pseudo-word set. 
Roughly one response in four departs from the intended category, with stronger consensus for round-associated phonemes (\cref{fig:human-phoneme}).
In pseudo-words, word-level round-choice rates are reproducible across participants (mean split-half Pearson $r=.892$; Spearman--Brown-corrected reliability $=.943$ over 500 random within-word splits), so we compare models both with the intended category and with the actual response of the participant on the same trial.


\paragraph{Gaze--choice coupling.}
\label{sec:gaze_choice}
The image a participant looked at more (by gaze samples or summed fixation duration) is the one they chose on 77--80\% of trials (80.58\% for constructed pseudo-words, by gaze samples), whereas the first fixation agrees at chance (51\%; Appendix~\ref{sec:app_gaze_choice}): gaze accumulates on the eventual choice \citep{shimojo2003gaze, krajbich2010visual}.
This association, not a causal relation, is what makes per-trial gaze informative as a supervision signal in \cref{sec:finetune}.




\subsection{Splits and behavioral references}
\label{sec:splits}

\paragraph{Fixed splits.}
Fine-tuning and evaluation use only the 7{,}141 constructed pseudo-word trials.
Adjectives are in-vocabulary words whose shape associations are encoded in their lexical meaning, and the canonical pairs are discussed throughout the pretraining corpora of current VLMs, so neither can test a sound--shape association on novel forms.
We partition the pseudo-word trials by word and by image pair into a training set (\Train; 1{,}932 trials), a validation set (\Val; 721 trials), and a held-out test set (\Test; 1{,}030 trials, 62 words, 70 image pairs, 53 participants) in which both the word and the image pair are unseen during training.
Participants are shared across splits, and participant identity is never a model input; the evaluation therefore measures generalization to new stimuli, not to new people.
 
\paragraph{Human references.}
\label{sec:references}
To provide a reference point for the fine-tuning results, we report two predictors on \Test{}; these values serve as references, not ceilings. 
Always choosing the intended category agrees with participants' actual choices on 72.82\% of \Test{} trials, 
and choosing each word's majority response among the other participants (leave-one-out) agrees on 72.86\%; the best attainable agreement cannot be estimated, because 915 of the 972 word$\times$pair$\times$orientation conditions in \Test{} were seen by only one participant.
For within-image alignment, a \emph{center-bias baseline} places an isotropic Gaussian ($\sigma=0.125$ in normalized composite-image coordinates) at the center of each image, and a \emph{population-mean map} averages the gaze of the other \Test{} trials separately for the left and right image positions, excluding the evaluated trial.
The first asks whether saliency beats a generic center-bias; the second asks how much alignment a shared gaze pattern reaches without trial-specific information.
Both are model-free maps that we integrate over each model's spatial cells and compare with the participant's gaze exactly as model saliency is compared (Section~\ref{sec:protocol}), so their values depend on the model's grid.
The center-bias baseline is also computed on all pseudo-word trials for the zero-shot comparison (\cref{tab:human_model_attention_alignment_combined}); the population-mean map is used on \Test{} only (\cref{sec:finetune}).



\section{Cross-modal associations in pretrained VLMs}
\label{sec:zeroshot}

\subsection{Evaluation protocol}
\label{sec:protocol}

We evaluate six embedding models (CLIP \citep{radford2021learning}, SigLIP \citep{zhai2023sigmoid}, and SigLIP2 \citep{tschannen2025siglip}) and six generative VLMs (Qwen3-VL \citep{bai2025qwen3}, Gemma4 \citep{team2026gemma}, and Molmo2 \citep{molmo2openweightsdata}) on the human forced-choice task.
Embedding models choose the image with the higher image--text cosine similarity, averaged over ten fixed caption templates (Appendix~\ref{sec:appendix_image_caption_analysis}). 
Generative models receive the original forced-choice prompt (Appendix~\ref{sec:appendix_prompt_forced}), with the composite image panel resized to $512\times512$ before each model's standard image preprocessing 
(Appendix~\ref{sec:appendix_human_model_alignment_methods}).
Their choices and attention come from one eager forward pass per input, with A/B probabilities from an FP32 projection of the final hidden state (Gemma's output softcap retained) and exact ties counted as half.
The original prompt names the label type and asks for ``visual/sound-symbolic fit,'' making it a favorable test of whether the model can express the association.
Because the cue could inflate agreement, we rerun with a neutral prompt that mirrors the participants' instruction (``Choose the image that better matches this word'') only for Gemma4-26B and Qwen3-VL-30B, the two models whose choices track human judgments; the smaller models are near chance and side-locked even with the cue.



For generative VLMs, we read out \dta{}, attention from the position that predicts the A/B answer to the image tokens, averaged over heads and over the last half of full-attention decoder layers~\citep{liu2026seeing}
(Gemma: global-attention layers only; Molmo: global and high-resolution tokens). 
For embedding models, we follow \citet{kouwenhovencross} and compute Grad-CAM maps~\citep{selvaraju2020grad}, using the fixed caption ``The label for this image is \{word\}'' on the $512\times512$ composed image input, keeping each model's native spatial cells without upsampling.
Note that this map explains the label's similarity to the whole panel, not the choice between the two images.
We then compare each model's map with the participant's gaze from the same trial.
We sum the participant's smoothed gaze within each of the model's cells and compute Spearman $\rho$ between gaze and model saliency separately for image A and image B; we then average across the two images and all trials.
Because $\rho$ is computed within each image, it does not depend on how attention is split between the two images; allocation between images is analyzed separately (Section~\ref{sec:finetune}).
As a reference, we apply the same computation to the center-bias baseline of \cref{sec:references}; because Grad-CAM and attention differ in resolution and smoothness, we compare each model with its own center reference rather than ranking models by raw $\rho$.
Fine-tuning is evaluated with the same computation.



\subsection{Zero-shot results}
\label{sec:zeroshot_results}

\paragraph{Choice alignment.}


Table~\ref{tab:model_results} shows that embedding models match adjectives well but generalize little to pseudo-words.
CLIP RN50 and ViT-B/32 reach ICA of 0.75 and 0.76 on adjectives, suggesting that CLIP representations encode lexical semantics, but pseudo-word ICA is near chance for most models. 
Only SigLIP-so400m shows a clear signal (ICA $=0.57$, Maj.\ $=0.72$, Spearman $=0.56$).

Most generative VLMs are near ceiling on adjectives, but pseudo-word alignment varies with size.
Gemma4-26B achieves the highest pseudo-word ICA (0.62), followed by Qwen3-VL-30B (0.57).
%
Human distribution metrics show that both models partially capture human judgments on pseudo-words, rather than merely achieving above-chance ICA: majority match 0.73 and 0.68, Pearson 0.58 and 0.57, and Spearman 0.58 and 0.56, respectively.
\molmo{} shows weaker positive signals, while \gemma, \qwen, and Molmo2-4B models show limited correlation with human judgments despite high adjective ICA.
Their zero-shot answers are largely side-locked (on \Test{}, \qwen{} answers B on 96\% of trials; Appendix~\ref{app:swap}), so chance-level ICA in the small models reflects a positional default rather than a label-dependent choice.

\paragraph{Instruction robustness.}
\label{sec:neutral_prompt}
Removing the sound-symbolic framing does not remove the association.
Correlation with human pseudo-word round-choice rates stays positive (Gemma4-26B: $r=.580\rightarrow.534$; Qwen3-VL-30B: $.571\rightarrow.535$; 95\% intervals above zero; paired changes include zero), and agreement with the participant's choice on the same trial does not decrease (57.30$\rightarrow$57.74\%; 55.73$\rightarrow$57.72\%, a gain of $+1.99$ percentage points (pp) [0.76, 3.22] for Qwen3-VL-30B).
Individual decisions do depend on the instruction: 33.23\% and 20.55\% of inputs receive a different choice, Gemma4-26B's ICA falls by 3.99\,pp [$-5.29$, $-2.74$], and majority match decreases (0.73$\rightarrow$0.70; 0.68$\rightarrow$0.57; unadjusted intervals, Appendix~\ref{sec:appendix_full_stats}).
The graded, human-like pattern is therefore not an artifact of the prompt,
although individual decisions depend on the instruction.

\begin{table*}[t]
\centering\footnotesize
\setlength{\tabcolsep}{.8pt}
\sisetup{
  table-number-alignment = center,
  round-mode = places,
  round-precision = 2,
  detect-weight = true,
  detect-family = true
}
\caption{VLM--human behavioral alignment on the 8{,}208 cleaned trials.
Maj. = majority-choice match per label; $r$, $\rho$ = Pearson and Spearman correlations between human and model label-level round-choice rates; ICA = intended-category agreement.
Embedding rows pool ten captions per trial; generative rows use the joint forward pass of Section~\ref{sec:protocol}; neutral-prompt rows, \cref{sec:neutral_prompt}.
An asterisk (*) means that nominal, unadjusted label-level $p<.05$; 95\% participant-cluster intervals are in Appendix~\ref{sec:appendix_full_stats}.}
\begin{tabular}{
  @{}
  l
  @{\hspace{10pt}}
  S[table-format=1.2] S[table-format=-1.2,table-space-text-post={*}]
  S[table-format=-1.2,table-space-text-post={*}] S[table-format=-1.2]
  @{\hspace{10pt}}
  S[table-format=1.2] S[table-format=-1.2,table-space-text-post={*}] S[table-format=-1.2,table-space-text-post={*}] S[table-format=-1.2]
  @{\hspace{10pt}}
  S[table-format=1.2] S[table-format=-1.2,table-space-text-post={*}] S[table-format=-1.2,table-space-text-post={*}] S[table-format=-1.2]
  @{}
}
\toprule
& \multicolumn{4}{c}{Overall} 
& \multicolumn{4}{c}{Pseudo-word} 
& \multicolumn{4}{c}{Adjective} \\
\cmidrule(lr){2-5}
\cmidrule(lr){6-9}
\cmidrule(lr){10-13}
Model 
& \multicolumn{1}{c}{Maj.} 
& \multicolumn{1}{c}{$r$} 
& \multicolumn{1}{c}{$\rho$} 
& \multicolumn{1}{c}{ICA}
& \multicolumn{1}{c}{Maj.} 
& \multicolumn{1}{c}{$r$} 
& \multicolumn{1}{c}{$\rho$} 
& \multicolumn{1}{c}{ICA}
& \multicolumn{1}{c}{Maj.} 
& \multicolumn{1}{c}{$r$} 
& \multicolumn{1}{c}{$\rho$} 
& \multicolumn{1}{c}{ICA} \\
\midrule
CLIP RN50 & 0.618 & 0.452$^{*}$ & 0.424$^{*}$ & 0.548 & 0.580 & 0.285$^{*}$ & 0.274$^{*}$ & 0.520 & 0.950 & 0.879$^{*}$ & 0.715$^{*}$ & 0.750 \\

CLIP ViT-B/32 & 0.597 & 0.375$^{*}$ & 0.344$^{*}$ & 0.534 & 0.549 & 0.116 & 0.132 & 0.505 & \bfseries 1.000 & 0.918$^{*}$ & 0.752$^{*}$ & 0.760 \\

SigLIP-base & 0.613 & 0.381$^{*}$ & 0.372$^{*}$ & 0.548 & 0.574 & 0.209$^{*}$ & 0.197$^{*}$ & 0.521 & 0.950 & 0.917$^{*}$ & 0.847$^{*}$ & 0.762 \\

SigLIP-so400m & 0.731 & 0.534$^{*}$ & 0.603$^{*}$ & 0.574 & 0.716 & 0.560$^{*}$ & 0.561$^{*}$ & 0.565 & 0.900 & 0.819$^{*}$ & 0.685$^{*}$ & 0.657 \\

SigLIP2-base & 0.495 & 0.324$^{*}$ & 0.162$^{*}$ & 0.531 & 0.438 & -0.095 & -0.121 & 0.497 & 0.950 & 0.923$^{*}$ & 0.756$^{*}$ & 0.786 \\

SigLIP2-so400m & 0.478 & 0.018 & -0.043 & 0.502 & 0.500 & 0.011 & -0.019 & 0.503 & 0.350 & -0.022 & -0.032 & 0.497 \\

\midrule

Gemma4-E4B & 0.565 & 0.447$^{*}$ & 0.326$^{*}$ & 0.554 & 0.512 & 0.153 & 0.085 & 0.511 & \bfseries 1.000 & 0.992$^{*}$ & 0.854$^{*}$ & 0.909 \\

Gemma4-26B & \bfseries 0.758 & 0.696$^{*}$ & \bfseries 0.695$^{*}$ & \bfseries 0.663 & \bfseries 0.728 & \bfseries 0.580$^{*}$ & \bfseries 0.579$^{*}$ & \bfseries 0.622 & \bfseries 1.000 & \bfseries 0.994$^{*}$ & 0.775$^{*}$ & \bfseries 0.970 \\

Qwen3-VL-8B & 0.613 & 0.527$^{*}$ & 0.456$^{*}$ & 0.566 & 0.562 & 0.262$^{*}$ & 0.242$^{*}$ & 0.516 & \bfseries 1.000 & 0.990$^{*}$ & 0.829$^{*}$ & 0.942 \\

Qwen3-VL-30B & 0.715 & \bfseries 0.707$^{*}$ & 0.693$^{*}$ & 0.613 & 0.679 & 0.571$^{*}$ & 0.562$^{*}$ & 0.566 & 0.950 & 0.985$^{*}$ & \bfseries 0.905$^{*}$ & 0.937 \\

Molmo2-4B & 0.570 & 0.417$^{*}$ & 0.349$^{*}$ & 0.530 & 0.525 & 0.131 & 0.143 & 0.507 & 0.950 & 0.866$^{*}$ & 0.817$^{*}$ & 0.716 \\

Molmo2-8B & 0.656 & 0.622$^{*}$ & 0.580$^{*}$ & 0.584 & 0.605 & 0.413$^{*}$ & 0.390$^{*}$ & 0.532 & \bfseries 1.000 & 0.987$^{*}$ & 0.825$^{*}$ & 0.936 \\

\midrule
\multicolumn{13}{@{}l}{\emph{Neutral prompt}} \\

Gemma4-26B & 0.726 & 0.678$^{*}$ & 0.656$^{*}$ & 0.625 & 0.704 & 0.534$^{*}$ & 0.516$^{*}$ & 0.582 & 0.950 & 0.982$^{*}$ & 0.771$^{*}$ & 0.945 \\

Qwen3-VL-30B & 0.624 & 0.665$^{*}$ & 0.644$^{*}$ & 0.609 & 0.574 & 0.535$^{*}$ & 0.524$^{*}$ & 0.566 & 0.950 & 0.983$^{*}$ & 0.869$^{*}$ & 0.945 \\

\bottomrule
\end{tabular}

\label{tab:model_results}
\end{table*}

\paragraph{Spatial alignment.}

\Cref{tab:human_model_attention_alignment_combined} compares each model's spatial $\rho$ with its center-bias baseline (\cref{sec:references}), a fixed Gaussian at each image center scored against the same participant's gaze on the same grid.
A model with lower $\rho$ than this baseline predicts gaze worse than a map that ignores the input.
Every evaluated model falls below its center-bias baseline.
Among embedding models, pseudo-word Grad-CAM--gaze correlations range from $-.091$ to $.588$; even the highest, CLIP ViT-B/32 ($\rho=.588$), is below its center-bias baseline ($.743$).
Among generative models, Gemma4-E4B and Gemma4-26B partly track gaze ($\rho=.340$ and $.265$; center $=.637$), whereas Qwen3-VL-8B, Qwen3-VL-30B ($.022$, $-.003$; center $.637$), Molmo2-4B, and Molmo2-8B ($-.074$, $.004$; center $=.601$) are near zero or slightly negative.
%
Because the two families require different saliency methods, this comparison cannot distinguish between model-level and readout-level differences.
Section~\ref{sec:finetune} therefore holds the model and the readout fixed and changes only the training signal.

\begin{table*}[t]
\centering\footnotesize
\setlength{\tabcolsep}{2.6pt}
\sisetup{
  table-number-alignment = center,
  round-mode = places,
  round-precision = 2,
  detect-weight = true,
  detect-family = true
}
\caption{Zero-shot choice and spatial alignment on the 7{,}141 pseudo-word trials. Choice $r$ = pseudo-word Pearson $r$ (Table~\ref{tab:model_results}). Spatial $\rho$ = within-image Spearman correlation between model saliency and the same participant's gaze on the model's native grid; Center = the center-bias baseline (\cref{sec:references}) on the same grid, the reference for each model. For grids and denominators, see Appendix~\ref{sec:appendix_human_model_alignment_methods}.}
\label{tab:human_model_attention_alignment_combined}
\resizebox{\linewidth}{!}{\begin{tabular}{@{}l*{6}{S[table-format=-1.3]}@{\hspace{10pt}}*{6}{S[table-format=-1.3]}@{}}
\toprule
 & \multicolumn{6}{c}{Embedding (Grad-CAM)} & \multicolumn{6}{c}{Generative (\dta{})} \\
\cmidrule(lr){2-7}\cmidrule(l){8-13}
 & \shortstack{CLIP\\RN50} & \shortstack{CLIP\\ViT-B/32} & \shortstack{SigLIP\\base} & \shortstack{SigLIP\\so400m} & \shortstack{SigLIP2\\base} & \shortstack{SigLIP2\\so400m}
 & \shortstack{Gemma4\\E4B} & \shortstack{Gemma4\\26B} & \shortstack{Qwen3-VL\\8B} & \shortstack{Qwen3-VL\\30B} & \shortstack{Molmo2\\4B} & \shortstack{Molmo2\\8B} \\
\midrule
Choice $r$   & 0.285 & 0.116 & 0.209 & 0.560 & -0.095 & 0.011 & 0.153 & 0.580 & 0.262 & 0.571 & 0.131 & 0.413 \\
Spatial $\rho$ & 0.312 & 0.588 & -0.053 & -0.091 & -0.056 & -0.065 & 0.340 & 0.265 & 0.022 & -0.003 & -0.074 & 0.004 \\
\midrule
Center       & 0.743 & 0.743 & 0.635 & 0.614 & 0.635 & 0.596 & 0.637 & 0.637 & 0.637 & 0.637 & 0.601 & 0.601 \\
\bottomrule
\end{tabular}}
\end{table*}


\section{Fine-tuning on human choices and gaze}
\label{sec:finetune}
The zero-shot results leave the link between choices and attention unclear. The small models' side-locked answers may hide an association or reflect its absence, the large models' choices are partly human-aligned, and saliency is read out differently for each model family.
Fine-tuning helps address those gaps: 
we ask (i) whether the association is learnable from a small set and generalizes to unseen words and images, (ii) whether human-aligned choices are then accompanied by attention closer to human gaze, and (iii) conversely, whether adding human gaze as a training target for attention makes choices more human-aligned.
All conditions train LoRA adapters on the \Train{} trials with the observed participant choice as the label and select checkpoints by \Val{} choice agreement, using the prompt and $512\times512$ input of Appendix~\ref{sec:appendix_prompt_forced}.
The \dta{} readout serves both as the gaze-loss target and as the attention measure.

\subsection{Training conditions}
\label{sec:conditions}

\textbf{\condC} trains on the participant's A/B choice alone.
\textbf{\condG} adds, for each training trial, the within-image gaze map of the participant who supplied that trial's choice; this is the main comparison against \condC, since it changes exactly one thing---one added loss term.
\textbf{\condA} adds the same loss with a single fixed target: one average gaze map per image position (left or right), estimated only from \Train{} gaze and applied to every training trial.
It is the control for \condG: whatever alignment \condA{} reaches is what a target that carries no trial-specific information can produce.
\textbf{\condS} additionally supervises how much attention falls on each of the two images through a jointly normalized gaze loss (Appendix~\ref{app:training_details}).
It is a robustness condition rather than the main comparison: it changes the normalization and image weighting at once, has no matched constant-target control, and its side target is coupled to the choice label in 80\% of trials (\cref{tab:gaze_choice}), so a choice gain under it would not isolate gaze information (Appendix~\ref{app:training_details}).%
\footnote{Because the image receiving more gaze is usually the chosen one, \condS{} partly re-supplies the choice label through the gaze channel. A choice gain under it would therefore not isolate gaze information beyond the label; the absence of such a gain is the more informative result, and \cref{sec:res_gaze_choice} reports \condS{} in that role.}

\subsection{Training objective}
\label{sec:objective}
All conditions minimize the same per-trial loss: the full-vocabulary cross-entropy of the participant's answer token $y_t$, plus a gaze term with weight $\lambda$ that is the only element varied across conditions,
\begin{equation}
\mathcal L_t=\operatorname{CE}(y_t)
+\lambda\,\frac{\sum_{i\in\{A,B\}} m_{t,i}\operatorname{KL}\mleft[H_{t,i}\mathbin{\Vert}A_{t,i}\mright]}{\sum_i m_{t,i}},
\label{eq:training_loss}
\end{equation}
where $m_{t,i}$ marks a valid human gaze image $i\in\{A,B\}$ on trial $t$, $H_{t,i}$ is the normalized gaze target defined below, and $A_{t,i}$ is the \dta{} readout normalized within that image after adding $10^{-6}$ to model token masses.
\condC{} uses $\lambda=0$; \condG, \condA, and \condS{} use $\lambda=1$, and we do not sweep $\lambda$.
The gaze term is zero if no image is valid; the choice loss is retained, and the training pool has at least one valid image per trial.
KL uses natural logarithms during training, with zero human bins contributing zero. 
Because each image is normalized separately, \cref{eq:training_loss} does not supervise the relative total attention mass on $A$ versus $B$.
Gaze targets are sample-density maps (Gaussian $\sigma=40$ screen pixels) integrated over each model's image-token regions; for \condA{} the target is replaced by the fixed \Train{} average for the same image position (Appendix~\ref{sec:optim}).


\subsection{Alignment measures}
\label{sec:measures}
On \Test{}, we report \humanagr{} and mean human-response NLL, $-\log_2 p(y_{\mathrm{human}})$ averaged over trials (uniform prediction: 1 bit); within-image $\rho$ and gaze KL as in \cref{sec:protocol}, with KL in bits summarized as
\begin{equation}
\Delta\operatorname{KL}=\operatorname{KL}(H\,\Vert\,H_{\text{train-mean}})-\operatorname{KL}(H\,\Vert\,A),
\label{eq:delta_kl}
\end{equation}
so that positive values indicate lower divergence than the fixed training-mean map; and allocation between images, via \sideJS{} and two side-agreement rates (\cattn{} and \gattn{}) reported as gaps over the agreement expected from their marginal A/B rates.
We compare within-image $\rho$ with the center-bias baseline and population-mean map of \cref{sec:references}, which are .637/.637/.603 and .731/.731/.680 on the token grids of \gemma/\qwen/\molmo.
Intervals are paired word$\times$pair$\times$participant cluster bootstraps conditional on the fitted seed and checkpoints; Holm correction is applied separately to the 18 \condG--\condA{} comparisons and the nine choice-NLL comparisons (Appendix~\ref{app:training_details}).

\subsection{Results}
\label{sec:results}
We first assess whether human-choice alignment generalizes to unseen words and image pairs, then examine how adding human gaze as a training target for attention changes choices, within-image attention, and allocation between images. 
Table~\ref{tab:main} reports all five conditions on the same \Test{} trials, using selected checkpoints from a single training seed; values below follow the order \gemma/\qwen/\molmo{}.

\begin{table}[t]
\centering
\footnotesize
\setlength{\tabcolsep}{2.5pt}
\caption{Matched \Test{} evaluation on 1{,}030 trials with unseen words and image pairs. 
\emph{Choice alignment.} 
ICA: intended category agreement; 
Human agr.: agreement with the actual participant choice;
NLL: average negative log-likelihood of the participant's choice (lower is better, and assigning 50\% to each image gives 1 bit).
\emph{Within-image alignment.} 
$\rho$: Spearman correlation between model attention and the participant's gaze inside each image;
$\Delta$KL: how much closer attention is to the participant's gaze than the average training gaze map is (\cref{eq:delta_kl}); positive means closer;
$^{*}$: \condG{} has a larger $\Delta$KL than \condA{} after Holm correction ($p<.05$; Appendix~\ref{app:gaze_mean}).
\emph{Allocation between images.}
Side JS: divergence between the shares of attention and of gaze falling on each image; lower is closer to humans.
Choice--attn: how often the model chooses the side it attends to more.
Gaze--attn: how often the model attends more to the side the participant looked at more.
Gap: observed agreement minus that expected if the two were independent, given how often each favors image B; a positive Gap indicates association, not that attention causes the choice.
Zero-shot rows are the models before fine-tuning, with choices read out as for the fine-tuned models.
}
\label{tab:main}
\sisetup{round-mode=none, table-align-text-post=false}
\resizebox{\linewidth}{!}{%
\begin{tabular}{@{}ll
  S[table-format=2.2]
  S[table-format=2.2]
  S[table-format=1.3]
  S[table-format=1.3]
  S[table-format=+1.3, table-space-text-post=\textsuperscript{*}]
  S[table-format=1.4]
  S[table-format=2.2]
  S[table-format=+2.2]
  S[table-format=2.2]
  S[table-format=+2.2]
@{}}
\toprule
 & & \multicolumn{3}{c}{Choice alignment} & \multicolumn{2}{c}{Within-image alignment} & \multicolumn{5}{c}{Allocation between images} \\
\cmidrule(lr){3-5}\cmidrule(lr){6-7}\cmidrule(lr){8-12}
 & & & & & & & & \multicolumn{2}{c}{Choice--attn} & \multicolumn{2}{c}{Gaze--attn} \\
\cmidrule(lr){9-10}\cmidrule(lr){11-12}
Model & Cond. & {ICA} & {Human agr.} & {NLL $\downarrow$} & {$\rho$ $\uparrow$} & {$\Delta$KL $\uparrow$} & {Side JS $\downarrow$} & {Agr.} & {Gap} & {Agr.} & {Gap} \\
 & & {(\%)} & {(\%)} & {(bits)} & & {(bits)} & {(nats)} & {(\%)} & {(pp)} & {(\%)} & {(pp)} \\
\midrule
\gemma & \condZ & 53.69 & 52.91 & 1.310 & 0.349 & -2.398 & 0.0285 & 78.54 & +17.68 & 54.37 & +2.45 \\
 & \condC & 92.72 & 72.72 & 0.817 & 0.308 & -2.655 & 0.0325 & 93.11 & +43.12 & 64.27 & +14.28 \\
 & \condG & 92.04 & 72.82 & 0.858 & 0.733 & +0.083\textsuperscript{*} & 0.0281 & 92.14 & +42.11 & 64.37 & +14.20 \\
 & \condA & 94.37 & 73.01 & 0.848 & 0.733 & +0.002 & 0.0312 & 53.20 & +3.83 & 46.21 & +1.32 \\
 & \condS & 94.56 & 72.62 & 0.814 & 0.730 & +0.070 & 0.0244 & 91.07 & +40.87 & 64.85 & +14.31 \\
\midrule
\qwen & \condZ & 49.90 & 48.54 & 1.835 & 0.032 & -3.776 & 0.0447 & 65.34 & +2.45 & 46.50 & -1.95 \\
 & \condC & 94.08 & 72.91 & 0.823 & 0.065 & -3.694 & 0.0469 & 59.32 & +9.59 & 49.51 & +0.77 \\
 & \condG & 92.43 & 72.82 & 0.813 & 0.729 & +0.078\textsuperscript{*} & 0.0253 & 88.64 & +38.13 & 63.40 & +12.77 \\
 & \condA & 89.61 & 71.75 & 0.814 & 0.731 & +0.004 & 0.0258 & 82.82 & +32.51 & 61.36 & +12.23 \\
 & \condS & 90.97 & 71.94 & 0.833 & 0.726 & +0.068 & 0.0247 & 90.49 & +40.36 & 64.37 & +14.48 \\
\midrule
\molmo & \condZ & 53.50 & 50.00 & 1.054 & 0.010 & -5.548 & 0.0667 & 80.68 & -1.86 & 48.93 & +0.44 \\
 & \condC & 92.52 & 71.94 & 0.883 & 0.062 & -5.457 & 0.0703 & 47.48 & +0.51 & 48.74 & +0.43 \\
 & \condG & 89.03 & 73.11 & 0.819 & 0.673 & +0.070 & 0.0293 & 86.41 & +35.43 & 64.37 & +14.15 \\
 & \condA & 89.71 & 72.43 & 0.819 & 0.679 & -0.007 & 0.0293 & 87.57 & +36.49 & 65.44 & +15.18 \\
 & \condS & 89.22 & 72.72 & 0.815 & 0.672 & +0.059 & 0.0285 & 89.61 & +39.19 & 65.24 & +15.14 \\
\bottomrule
\end{tabular}}
\end{table}

\paragraph{Learning from human choices.}
\label{sec:res_learnable}
Training on 1{,}932 observed human choices raises \humanagr{} from 52.91/48.54/50.00\% under \condZ{} to 72.72/72.91/71.94\% under \condC{}.
Under \condC{}, ICA is 92.72/94.08/92.52\%, indicating that the models choose the intended shape category more often than they match the participant's actual response.
Mean human-response NLL also decreases, from 1.310/1.835/1.054 to 0.817/0.823/0.883 bits, below the uniform-choice reference of 1 bit.
Alignment with human responses is thus learnable from this small training set and transfers to new words and image pairs, although the comparison does not establish whether the association was already present before fine-tuning.
The observed human-choice agreement under \condC{} is similar to the 72.86\% human majority-vote reference (\cref{sec:references}).
On the 280 trials where participants chose against the intended category, \condC{} agrees with them on only 13.21/11.07/12.14\% of trials.
The learned choices are also robust to image position, so the model chooses the same shape regardless of whether it is shown on the left or right (Appendix~\ref{app:swap}).

\paragraph{Effects of adding human gaze as a training target for attention.}
\label{sec:res_gaze_choice}
Relative to \condC{}, adding trial-specific within-image gaze changes \humanagr{} by $+0.10$ [$-3.87$, $+4.49$], $-0.10$ [$-5.19$, $+4.84$], and $+1.17$ [$-3.98$, $+6.62$]\,pp (individual 95\% intervals), and all nine NLL contrasts among \condC, \condG, and \condA{} have intervals containing zero (Holm-adjusted $p=1.000$).
The probability-based evaluation gives the same qualification. 
Although gaze accumulates on the chosen image in 80.58\% of pseudo-word trials (\cref{tab:gaze_choice}), supervising allocation between images with \condS{} improves \sideJS{} relative to \condC{} for all three models (Appendix~\ref{app:gaze_side}) without a detectable additional choice gain~(72.62/71.94/72.72\%; all paired confidence intervals against \condC{} and \condG{} include zero).

\looseness=-1 \paragraph{Agreement without gaze alignment.}
\label{sec:res_choice_attn}
Despite the increase in \humanagr{}, \condC{} reaches within-image gaze correlations of only .308/.065/.062, below the center-bias baseline~(.637/.637/.603; \cref{sec:references}).
The changes from \condZ{} are also inconsistent in direction. 
Correlation decreases for \gemma{} and increases modestly for \qwen{} and \molmo{}.
Choice training therefore does not produce large, consistent gaze alignment at the decision-token readout used here.

\looseness=-1 \paragraph{Average-gaze control.}
\label{sec:res_avg}
\condG{} raises \rhoin{} to .733/.729/.673, but \condA{} reaches .733/.731/.679 while using the same fixed training-average gaze target on every trial.
High within-image correlation therefore does not by itself demonstrate learning of trial-specific gaze patterns.
The \condG{}$-$\condA{} difference in $\Delta$KL (\cref{tab:main}) is an estimated KL reduction of .074--.081 bits, meeting the Holm-adjusted .05 threshold for \gemma{} and \qwen{}, but not \molmo{} (Appendix~\ref{app:gaze_mean}).

\paragraph{Allocation between images.}
\looseness=-1 Similar choice scores also coexist with different links between choice and attention. 
Under \condC{}, \humanagr{} differs by less than one point across models, yet the gap between \cattn{} and its value expected under independence is $+43.12$/$+9.59$/$+0.51$ pp~(Table~\ref{tab:main}).
A near-zero gap suggests that an agreement reflects shared side preferences rather than a link between attention and choice (Appendix~\ref{app:gaze_side}).
This is an association between the attended side and the answer, not evidence that attention causes the choice.





\section{Discussion}
\looseness=-1 Our results show a gap between adjective matching, which most models handle, and pseudo-word sound--shape alignment, which emerges mainly in a subset of larger VLMs.
Molmo2-8B illustrates the distinction: near-perfect adjective ICA ($0.94$) with near-chance pseudo-word ICA ($0.53$).
Adjective performance can rest on learned associations between familiar word forms and their referents; the partial success of Gemma4-26B and Qwen3-VL-30B on pseudo-words cannot rest on such associations, since the words are novel; it more plausibly reflects statistical regularities linking written form, phonology, and visual features in training data.
%
Choice alignment does not imply human-aligned visual attention.
Gemma4-26B's pseudo-word choices correlate with human judgments at $r\approx.58$ while its attention correlates with gaze at only .27, and fine-tuning on human choices raises \humanagr{} by ${\sim}20$\,pp while \rhoin{} stays below a center-bias baseline.
These models likely reach human-aligned choices through holistic image representations rather than the local shape-informative features on which human fixations concentrate, consistent with findings that vision transformers integrate information globally from early layers onward~\citep{NEURIPS2021_652cf383}.
Judged by choice agreement alone, our fine-tuned models would appear to match the human majority; spatial alignment is the more diagnostic test.
%
%
Adding human gaze as a training target for attention raises within-image attention--gaze correlation from below the center-bias baseline to .67--.73, with no detectable change in \humanagr{} or NLL, and a single average gaze map per image position produces the same correlation gain (\cref{sec:res_avg}).
Prior work reports that gaze regularization improves attention alignment~\citep{pani2025gazeVLMneurips, Yu2017SupervisingNA, NEURIPS2020_460191c7}; 
our results agree on attention, but the gain does not carry over to choices and is matched by a constant average gaze map, echoing VQA gains from human-attention supervision that random maps reproduce~\citep{shrestha-etal-2020-negative}.

\paragraph{Limitations.}
Fine-tuning uses three models, one optimization seed,
one gaze-loss weight, and the original prompt.
Attention and Grad-CAM are layer-dependent correlational readouts, not causal explanations.
Written labels confound phonology and orthography; participants were English-speaking members of one institution and shared across splits.

\section{Conclusion}
We presented a direct human--VLM comparison of sound--shape associations, evaluating both choice alignment and spatial alignment on the same stimuli.
Many models matched English adjectives to corresponding visual shapes, but pseudo-word alignment appears mainly in a subset of generative VLMs, and even there, model saliency diverges from human gaze.
%
Fine-tuning closes the choice gap on unseen words and images without closing the spatial gap.
Choice agreement alone is therefore insufficient evidence of human-aligned cross-modal processing; both choices and their visual attention must be evaluated.

\section*{AI use statement}
In this work, we used generative AI tools for generating synthetic datasets, as detailed in Section~\ref{sec:visual_stimuli}.
Additionally, we used generative AI tools to polish writing to improve readability.
We have not used generative AI tools to help develop theoretical models or conceptual frameworks, formulate mathematical claims, provide critical ingredients for proving mathematical claims, assist in the writing of proofs, propose or refine hypotheses, design or provide feedback on research methodology or experiments, implement methods, assist with translation, clean and reformat datasets, support qualitative and thematic data analysis, or interpret results. 
We take responsibility for the final content of this work, including text, claims, or artifacts produced with the aid of generative AI.

\section*{Ethics statement}
This work adheres to the ICLR Code of Ethics.
The protocol for human subject research in this work was approved by the IRB at our institution. Participants gave informed consent, including consent to release de-identified choice and gaze data. Audio and video recordings were used only to verify responses and will not be released; released data contain no names, audio, video, or other personal identifiable information.

\section*{Reproducibility statement}
We will provide all code after publication.
We will release all datasets used, including the stimuli and gaze data.

\bibliography{iclr2027_conference}
\bibliographystyle{iclr2027_conference}

\appendix

\section{Stimuli}
\label{sec:app_stimuli}
Below is the full list of adjectives from \citet{alper2023kiki}, used as a baseline:

\begin{itemize}
    \item $A_{\text{sharp}} = $ \{sharp, spiky, angular, jagged, hard, edgy, pointed, prickly, rugged, uneven\}
    \item  $A_{\text{round}} = $ \{round, circular, soft, fat, chubby, curved, smooth, plush, plump, rotund\}.
\end{itemize}

\paragraph{Note on AH.}
{AH is not a rounded vowel in every English dialect, but we include it in the round-associated set following prior human studies
\citep{nielsen2013parsing, maurer2006shape}, where labels such as \textit{bouba/BOOBAH} have been used as round-associated stimuli.}

\section{Prompt for Image Generation}
\label{sec:appendix_prompt}
The prompt template used to generate images using Google's Gemini \citep{team2023gemini} in Section~\ref{sec:visual_stimuli} is provided in Figure~\ref{fig:llm-prompt-box}.

\begin{figure}[ht]
    \centering
    \begin{tcolorbox}[colback=gray!5!white, colframe=gray!75!black]
        Generate a distinct image pair of \{\textit{$material$}\}, \{\textit{$adjective_1$}\} vs \{\textit{$adjective_2$}\} abstract objects, where both objects are \{\textit{$color$}\}, do not include text, no frames, ensure it is different from all previous images, and use a white background with no separation.
    \end{tcolorbox}
    \caption{Prompt used to generate images.}
    \label{fig:llm-prompt-box}
\end{figure}

For all prompts, $adjective_1$ and $adjective_2$ are the 20 adjectives from \citet{alper2023kiki}, listed in Section~\ref{sec:linguistic_stimuli}.
Each pair of adjectives is always from different sets (i.e., one adjective from $A_{\text{sharp}}$ and one from $A_{\text{round}}$).
Some prompts do not include $material$, while others specify materials such as \textit{plastic-like} or \textit{metal-like}.

\begin{figure}
    \centering
    \includegraphics[width=\linewidth]{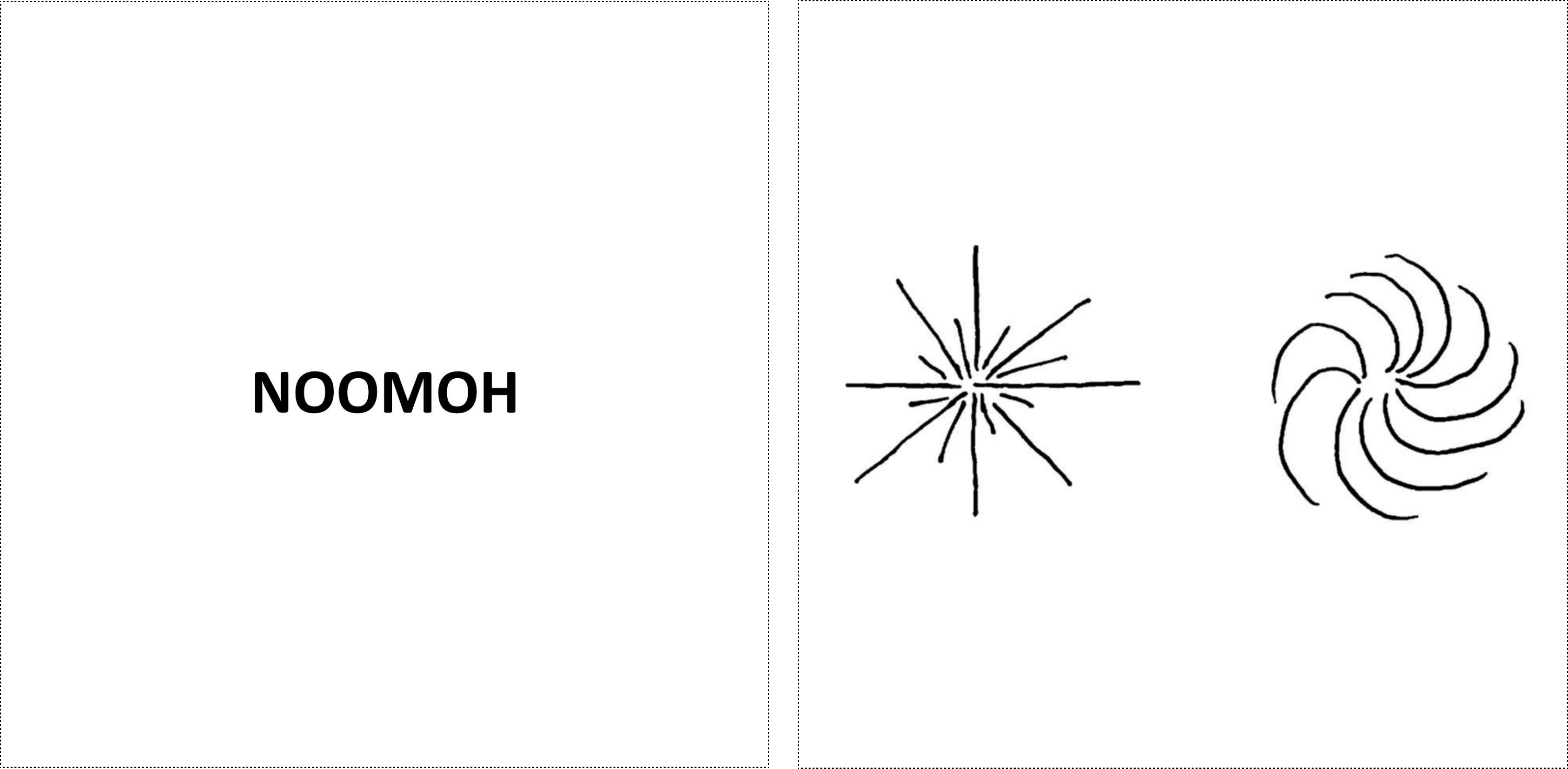}
    \caption{Stimuli pair in the human study. Word stimuli come first, and following image stimuli come second.}
    \label{fig:stimuli-human-study}
\end{figure}

\section{Prompt for Forced Choice Analysis}
\label{sec:appendix_prompt_forced}
 
The prompt used for forced choice analysis of generative VLMs is in Figure \ref{fig:llm-prompt-forced}.
\begin{figure}[ht]
    \centering
    \begin{tcolorbox}[colback=gray!5!white, colframe=gray!75!black]
        You are doing a forced-choice visual alignment task.\\
 
You will see one image containing two options:
 
- Image A is on the LEFT.
 
- Image B is on the RIGHT.\\
 
Label: "{label}"\\
Label type: {label\_type}\\
 
Task:
Choose which image, A or B, better aligns with the label.\\
 
Important rules:\\
\begin{enumerate}
    \item You must choose exactly one option: A or B.
    \item Do not answer "both", "neither", "unclear", or "tie".
    \item If the label is a pseudoword, do not try to define it. Judge only by visual/sound-symbolic fit.
    \item Base your answer only on the visual appearance of A and B.
    \item Return only valid JSON with this exact schema:
\end{enumerate}
{{"choice":"A"}}
 
or
 
{{"choice":"B"}}
    \end{tcolorbox}
    \caption{Prompt used for forced choice analysis.}
    \label{fig:llm-prompt-forced}
\end{figure}
 
The neutral prompt used for the instruction-robustness check (Section~\ref{sec:neutral_prompt}) follows the participants' instruction, ``Choose the image that better matches this word'' and is shown in Figure~\ref{fig:llm-prompt-forced-neutral}. 
Information about the label type and sound symbolism is removed, as well as the rules, except for returning a JSON.

\section{Prior Visual Stimuli}
\label{sec:appendix_images}

\begin{figure*}[ht]
    \centering
    \includegraphics[width=0.6\textwidth]{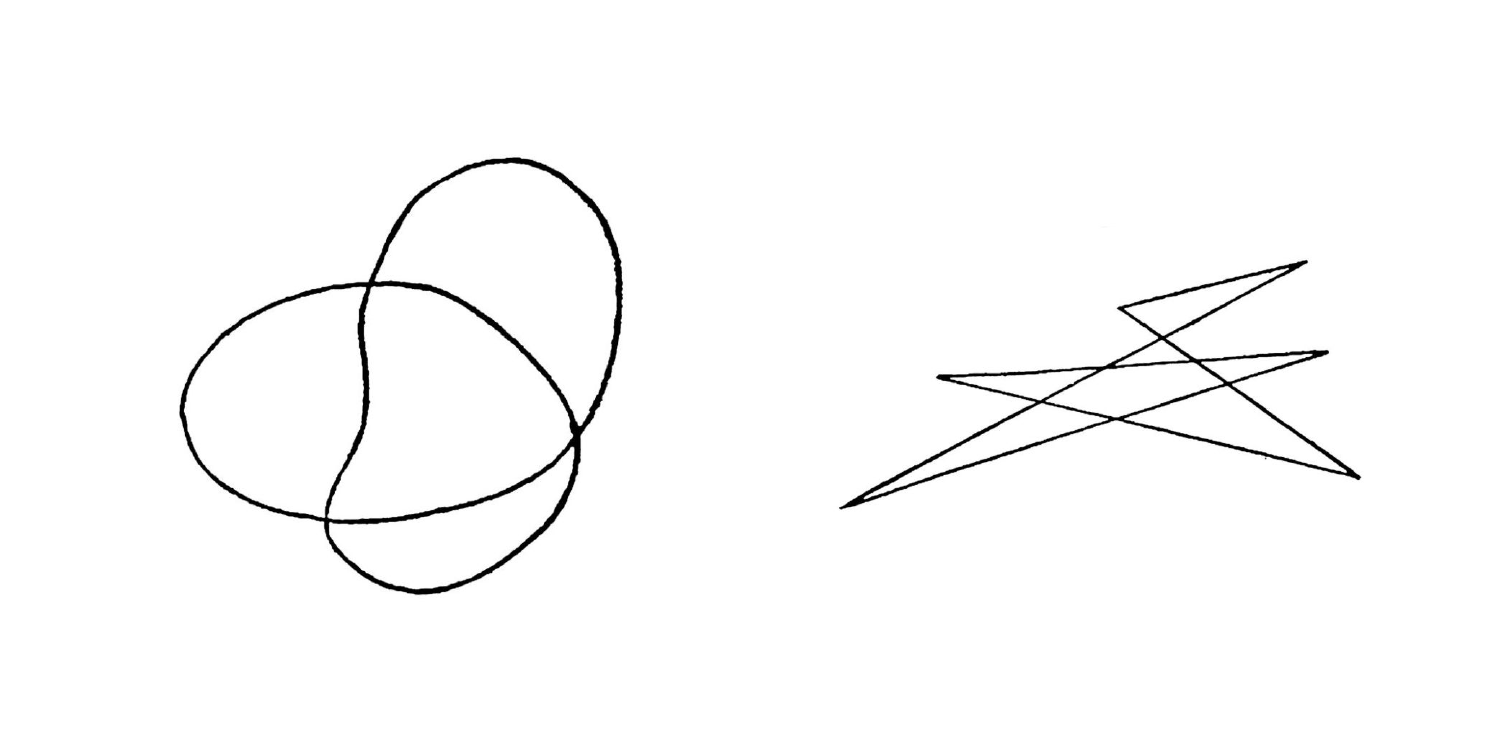}
    
    \caption{Image pairs from \citet{kohler1929gestalt, kohler1947gestalt}}
    \label{kohler_img}
\end{figure*}

\begin{figure*}[ht]
    \centering
    \includegraphics[width=0.4\textwidth]{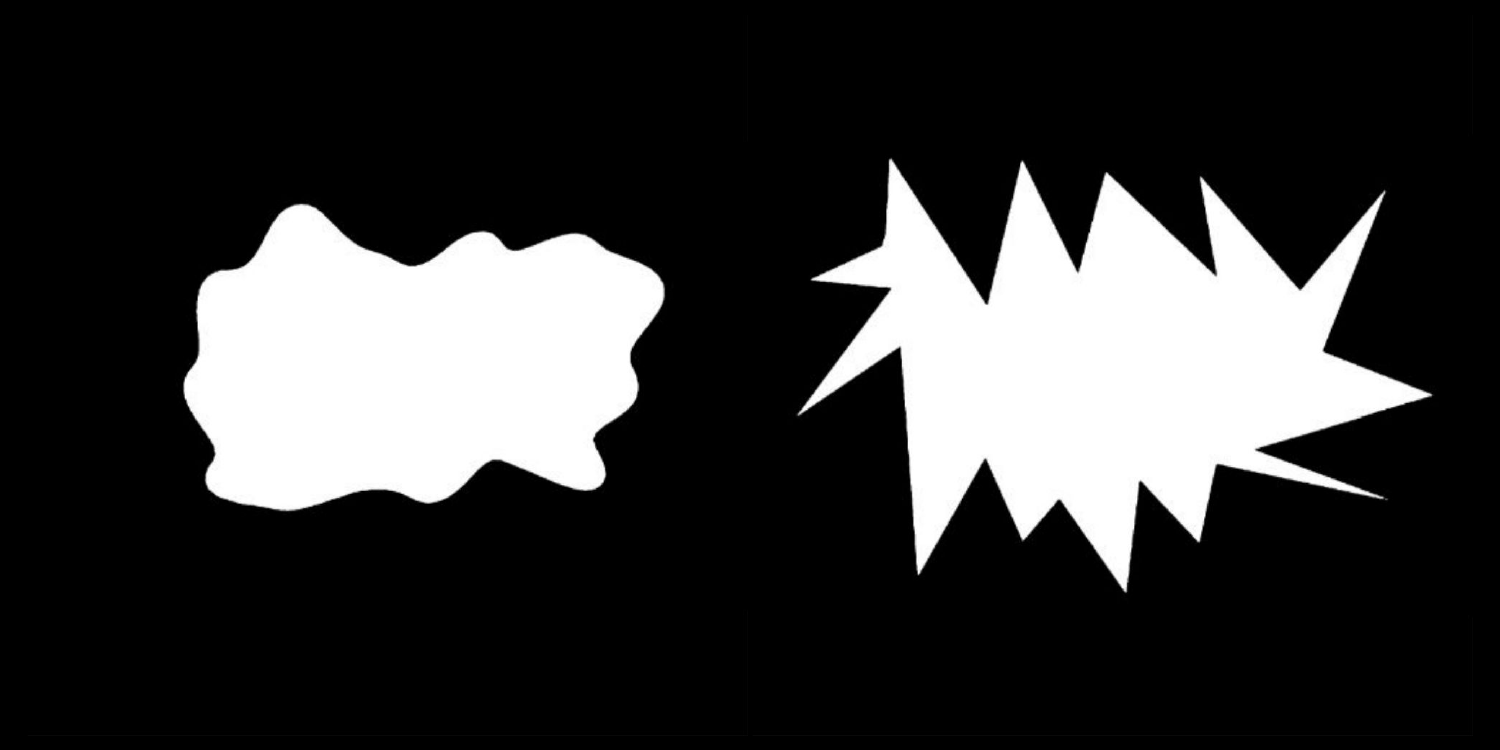}
    \includegraphics[width=0.4\textwidth]{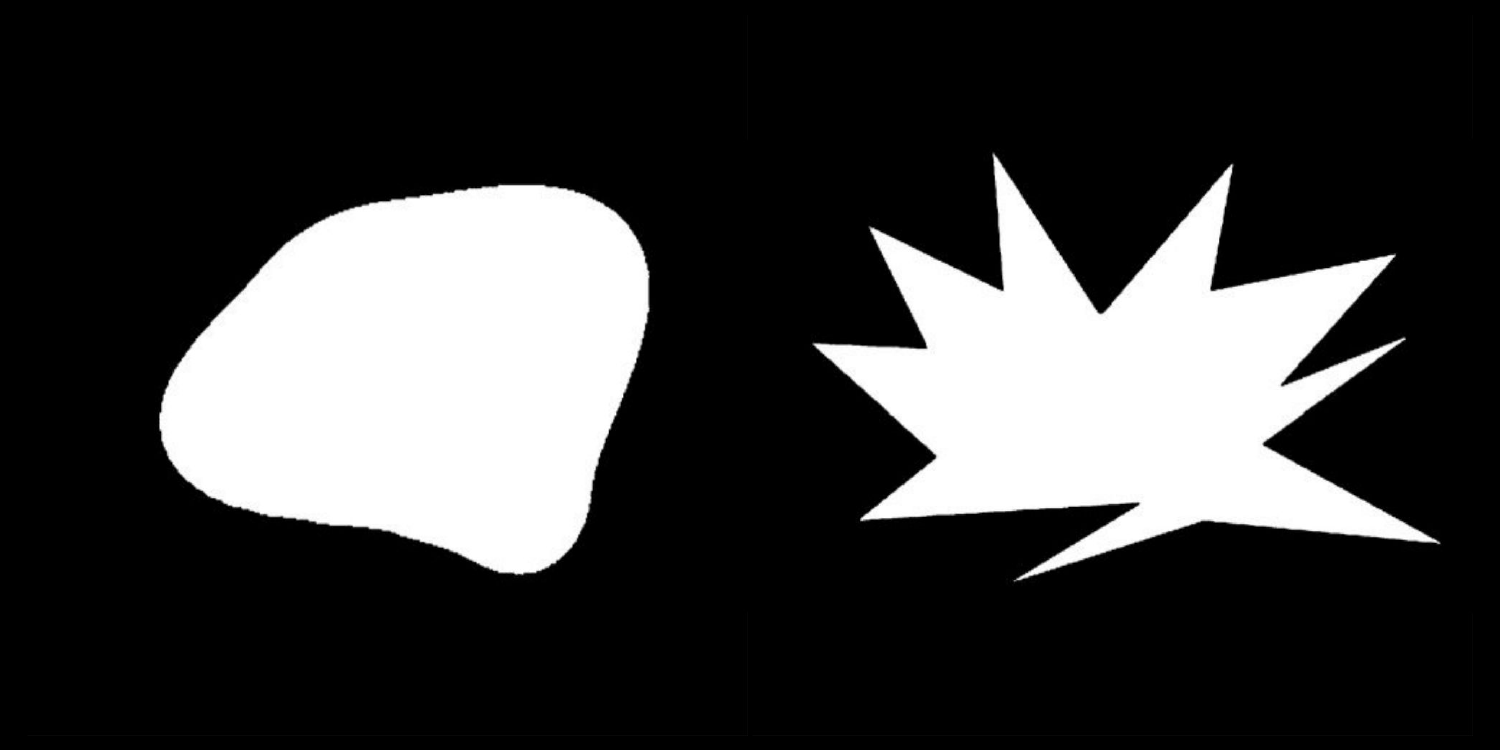}
    \includegraphics[width=0.4\textwidth]{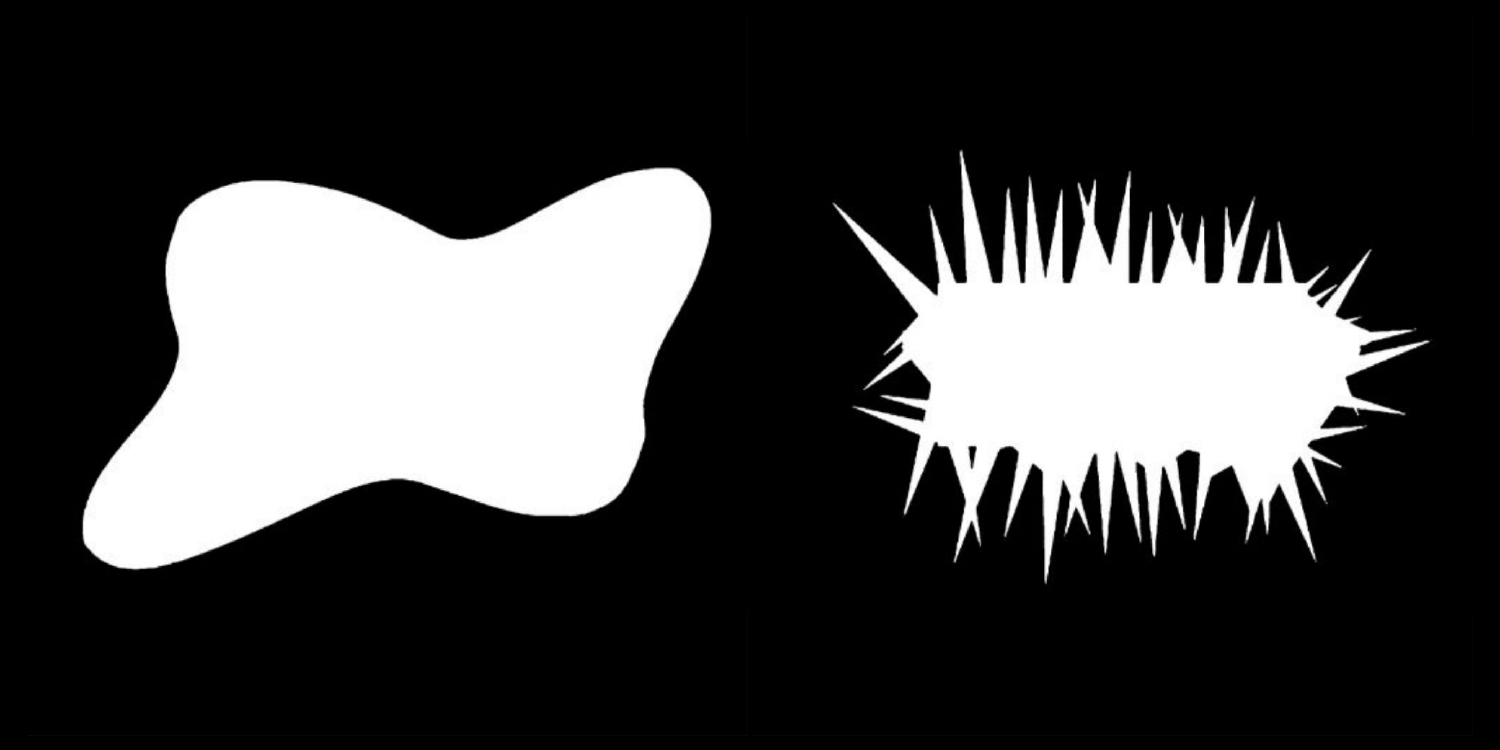}
    \includegraphics[width=0.4\textwidth]{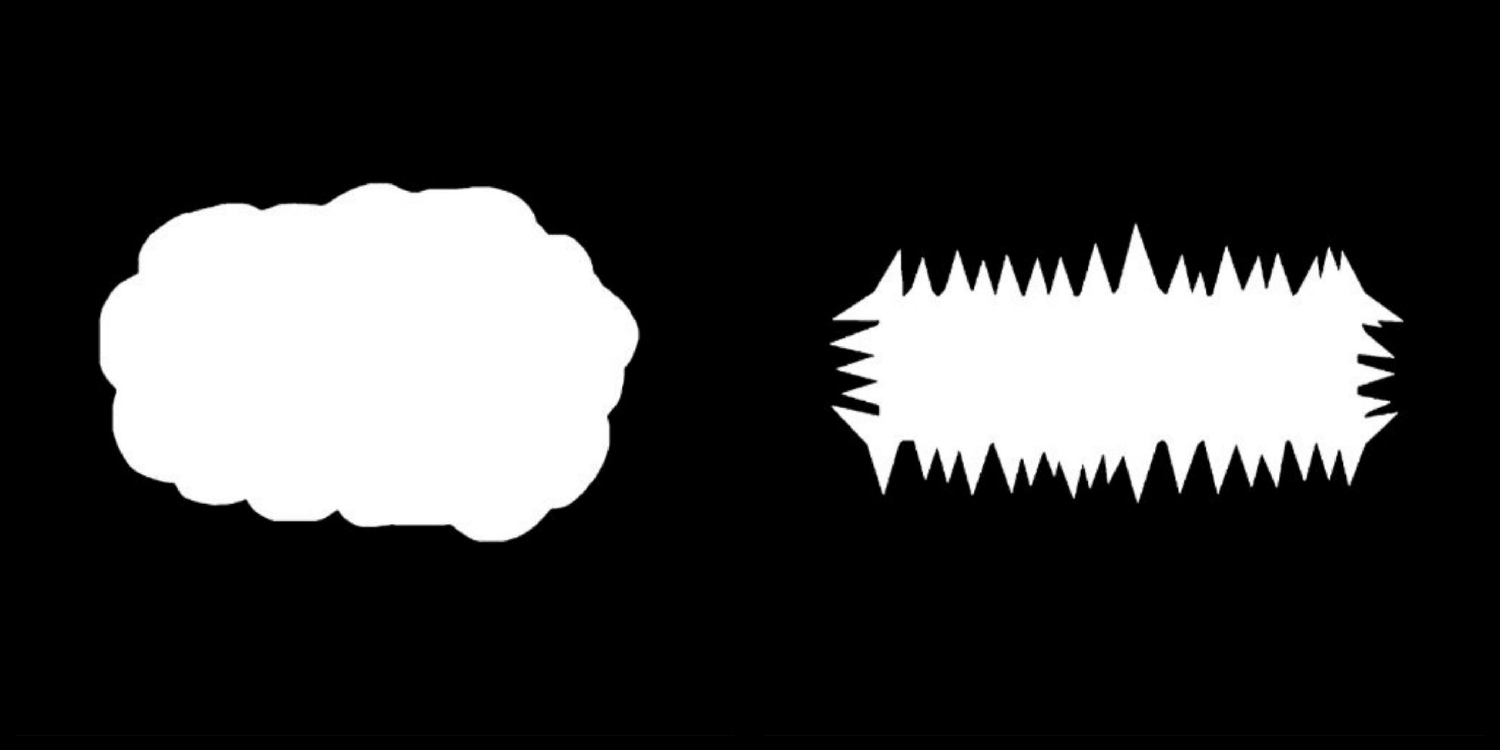}
    \caption{Image pairs from \citet{westbury2005implicit}}
    \label{west_img}
\end{figure*}

\begin{figure*}[ht]
    \centering
    \includegraphics[width=0.4\textwidth]{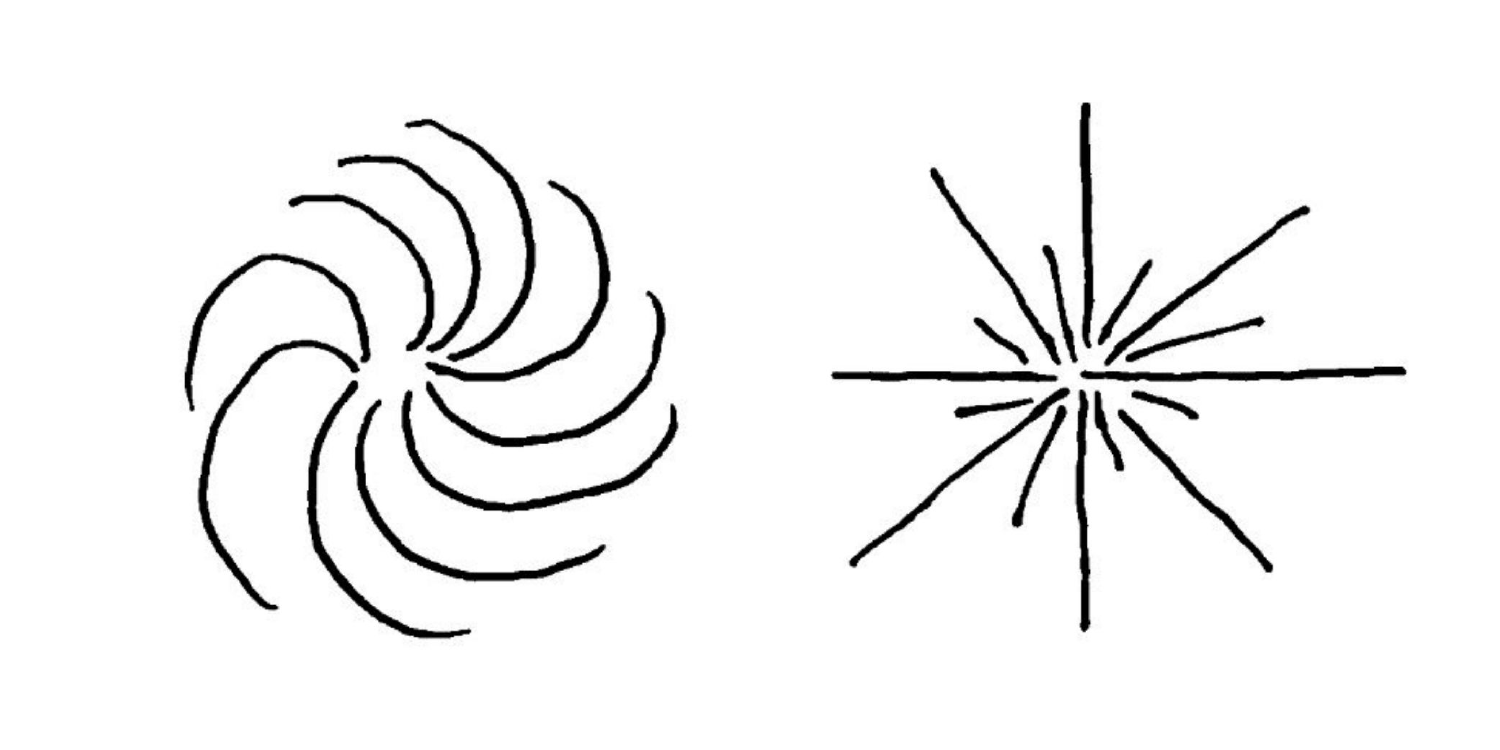}
    \includegraphics[width=0.4\textwidth]{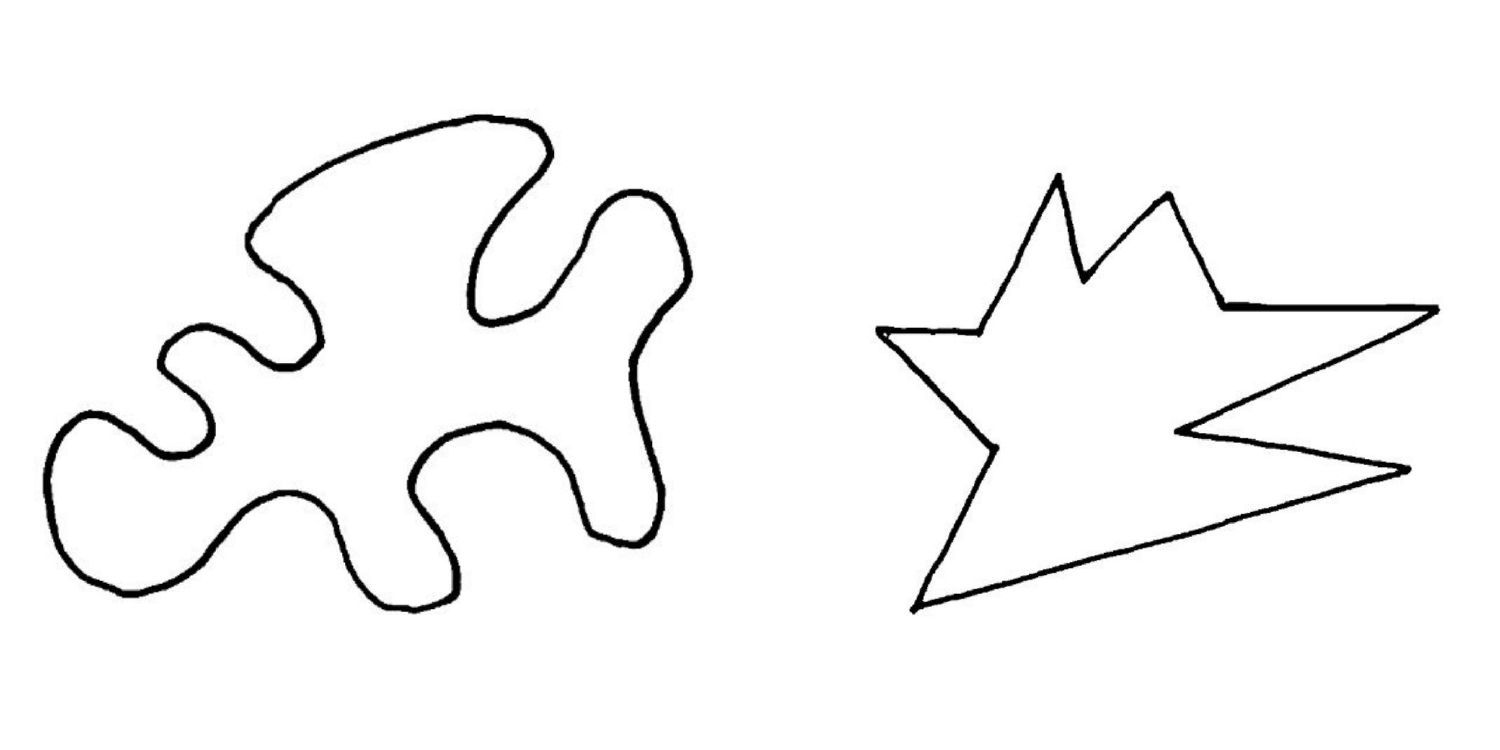}
    \includegraphics[width=0.4\textwidth]{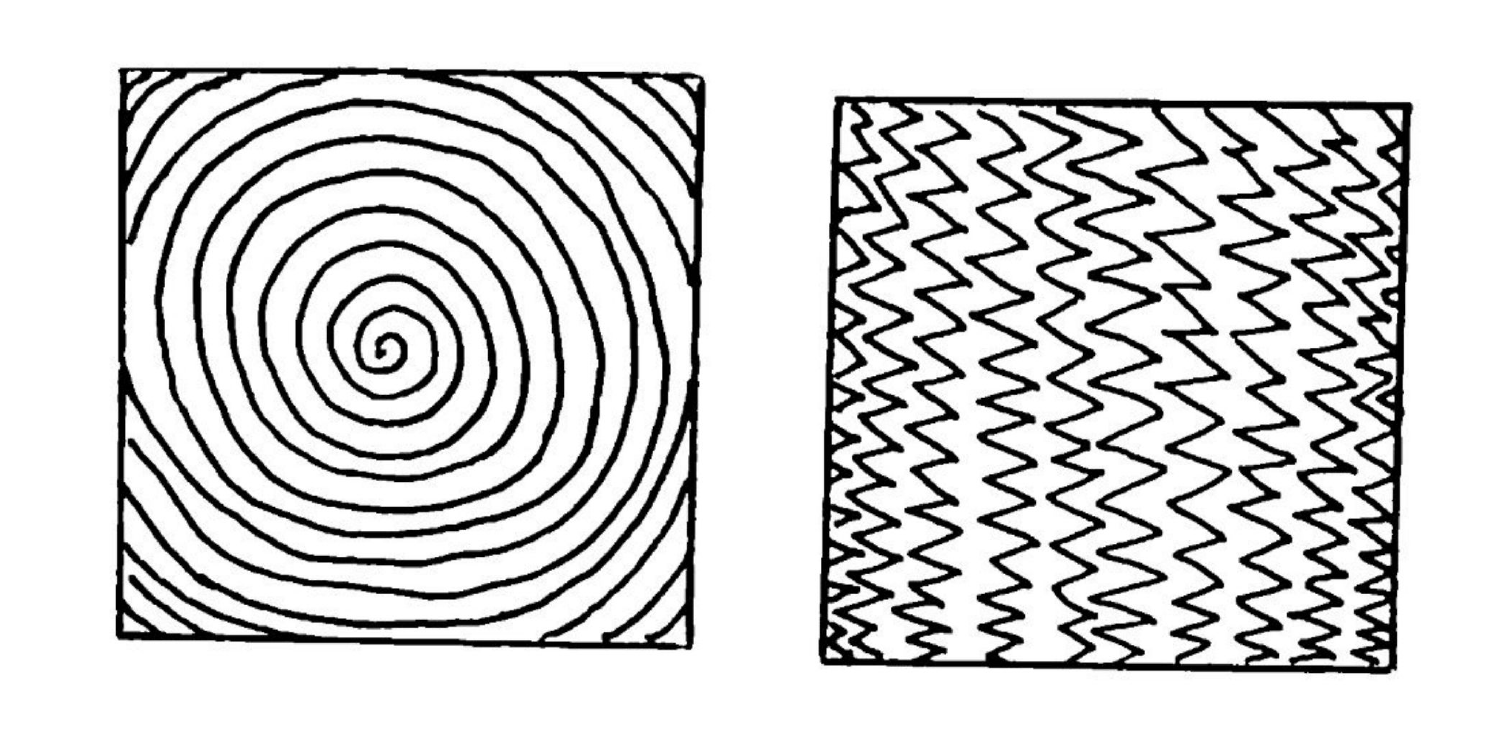}
    \includegraphics[width=0.4\textwidth]{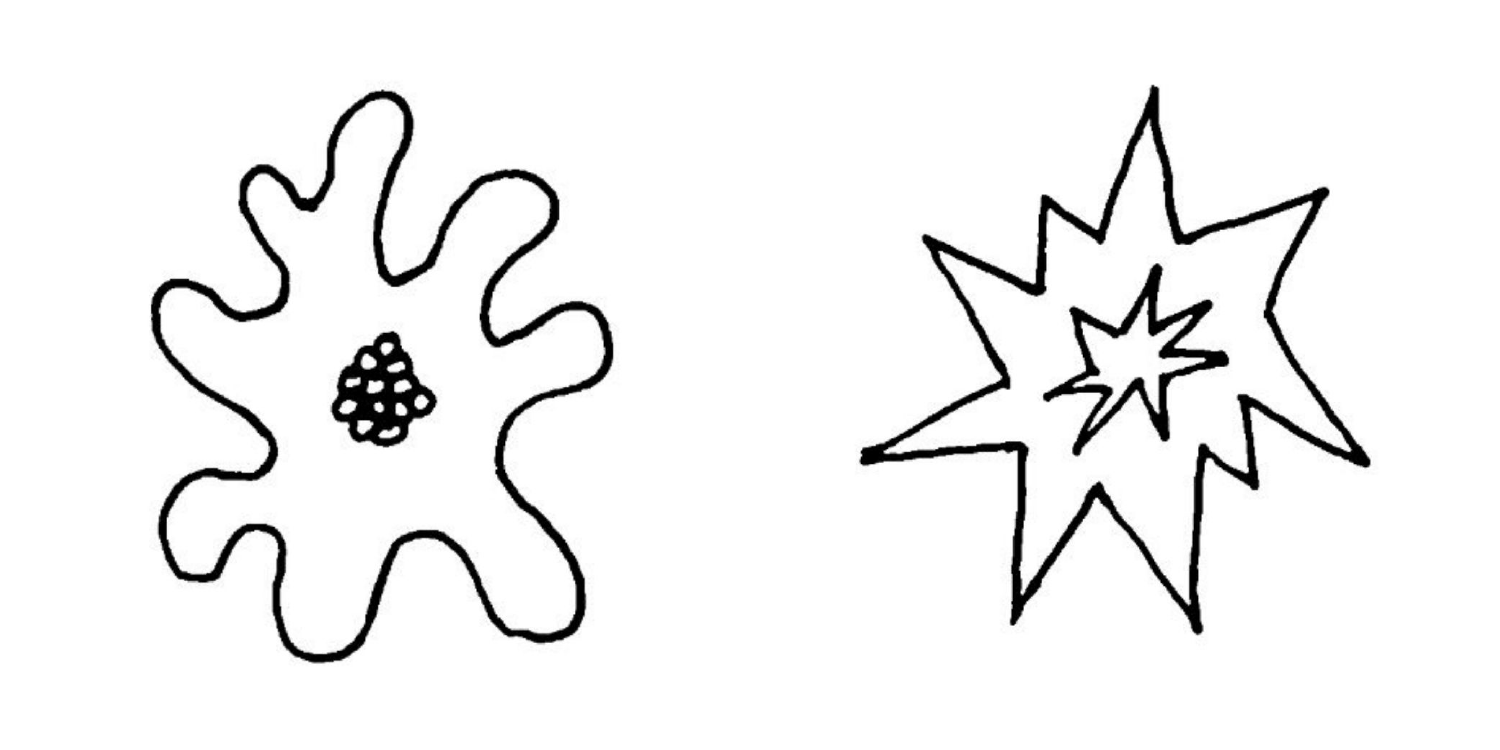}
    \caption{Image pairs from \citet{maurer2006shape}}
    \label{maurer_img}
\end{figure*}

\begin{figure*}[ht]
    \centering
    \includegraphics[width=0.25\textwidth]{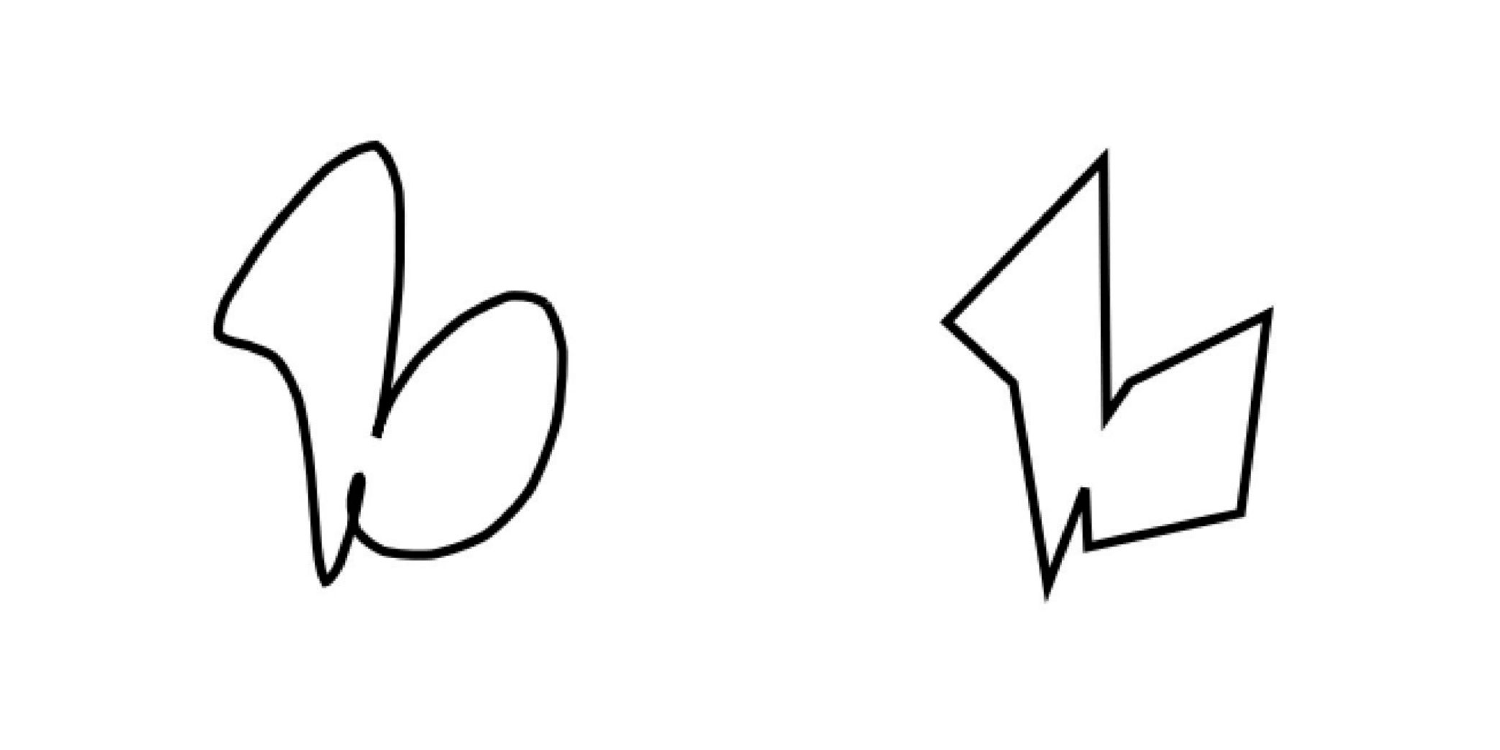}
    \includegraphics[width=0.25\textwidth]{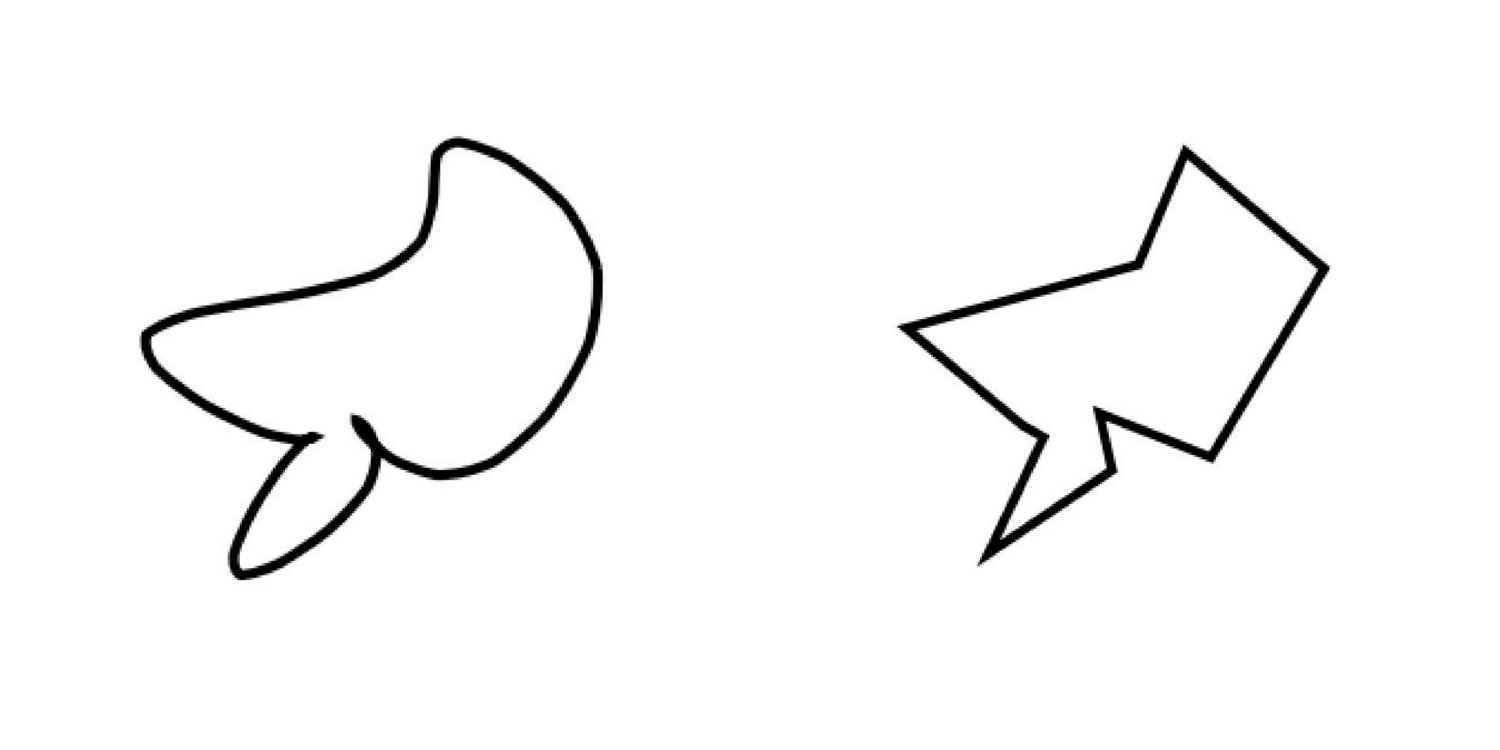}
    \includegraphics[width=0.25\textwidth]{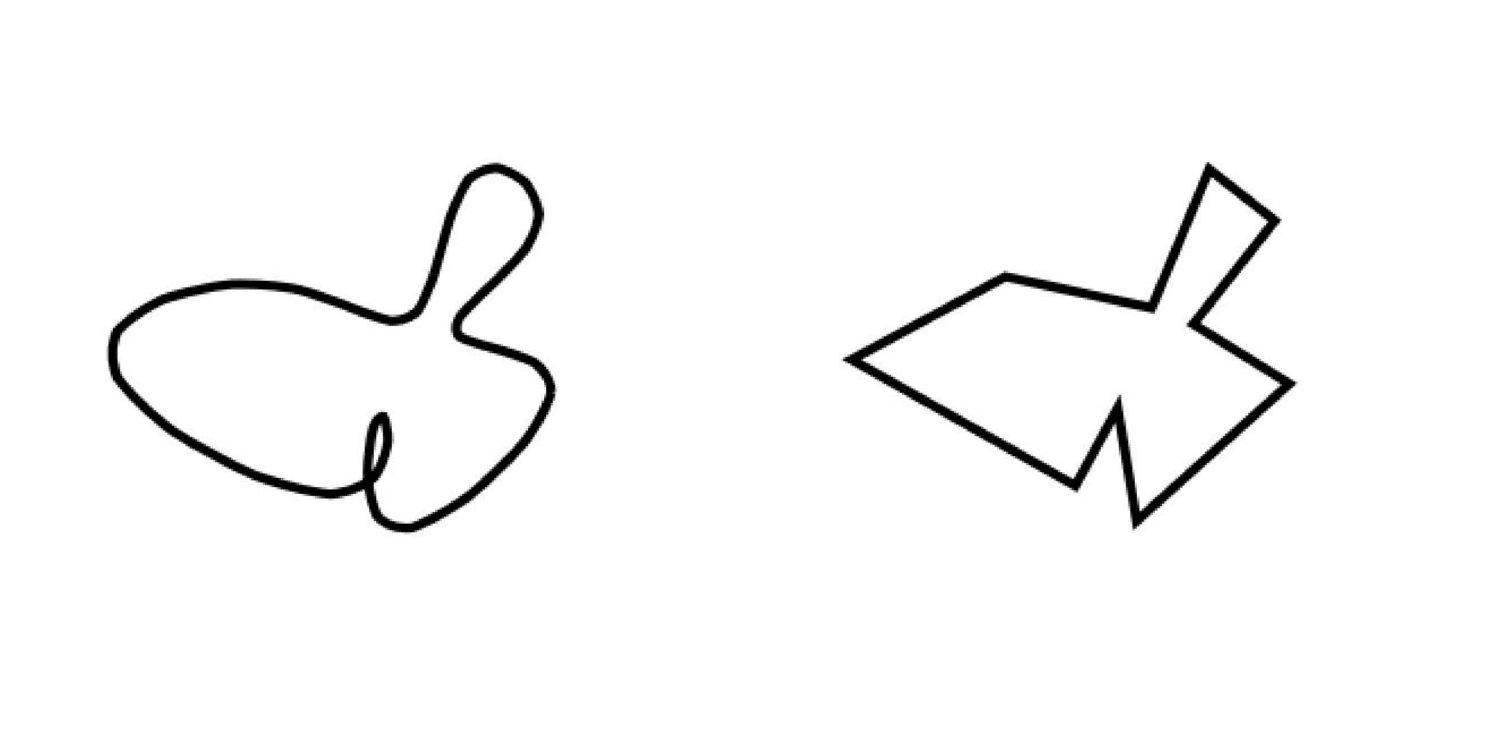}
    \includegraphics[width=0.25\textwidth]{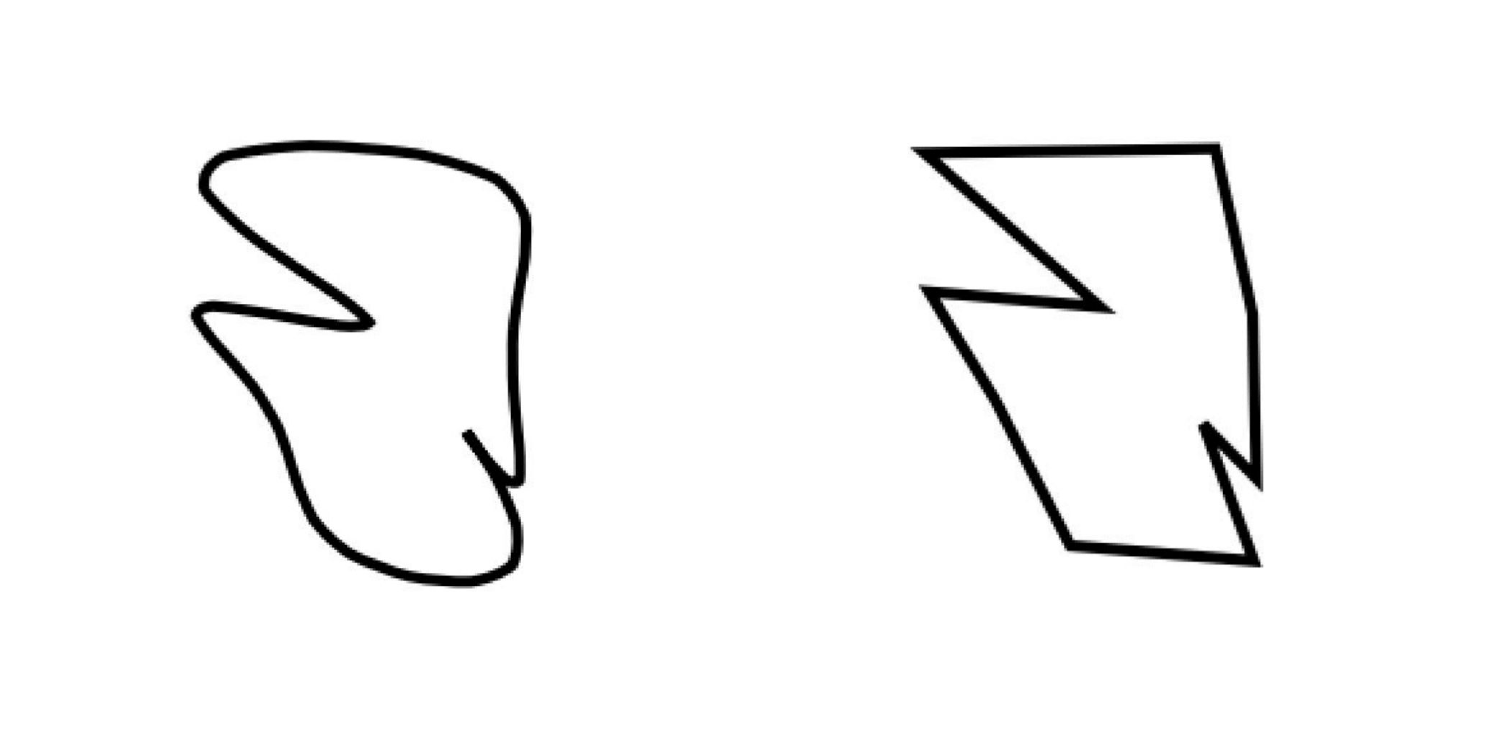}
    \includegraphics[width=0.25\textwidth]{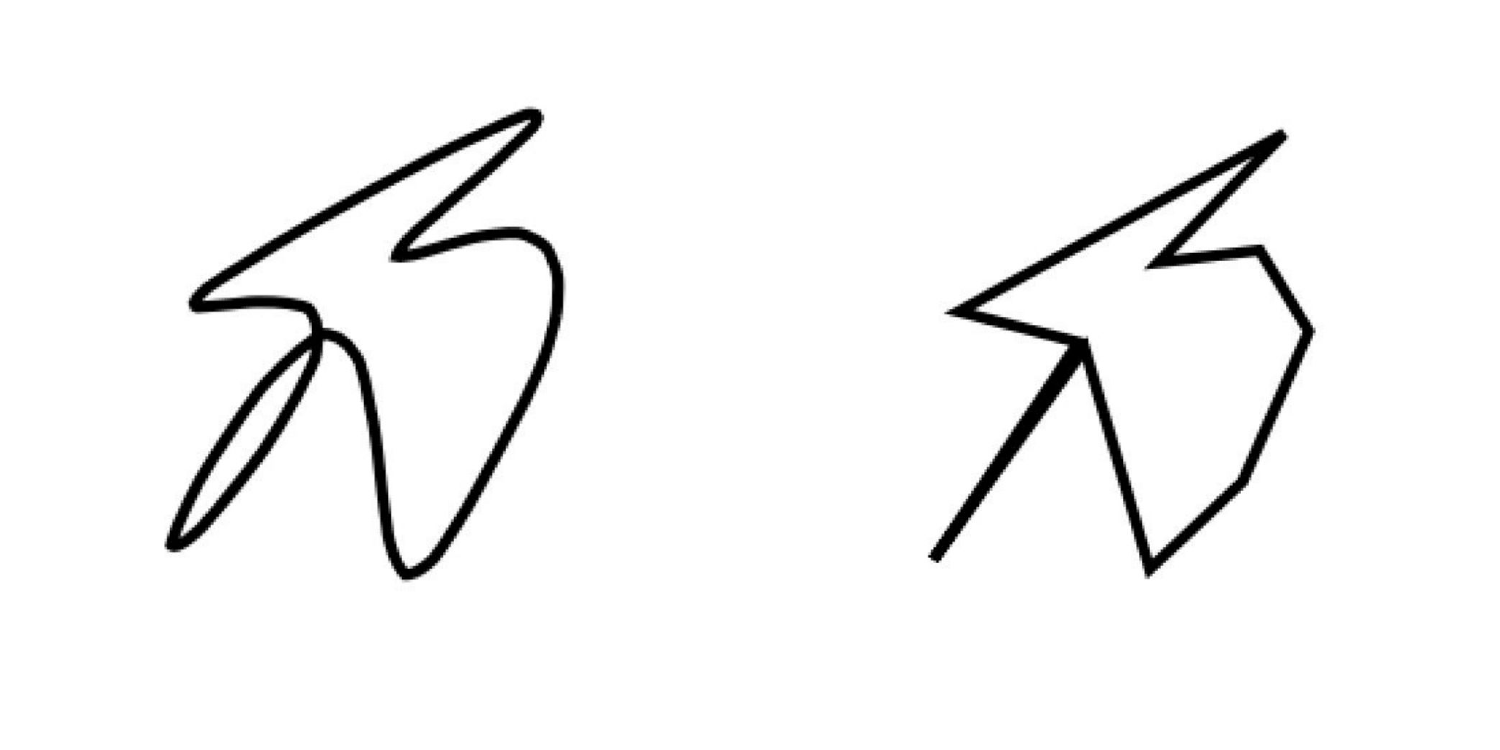}
    \includegraphics[width=0.25\textwidth]{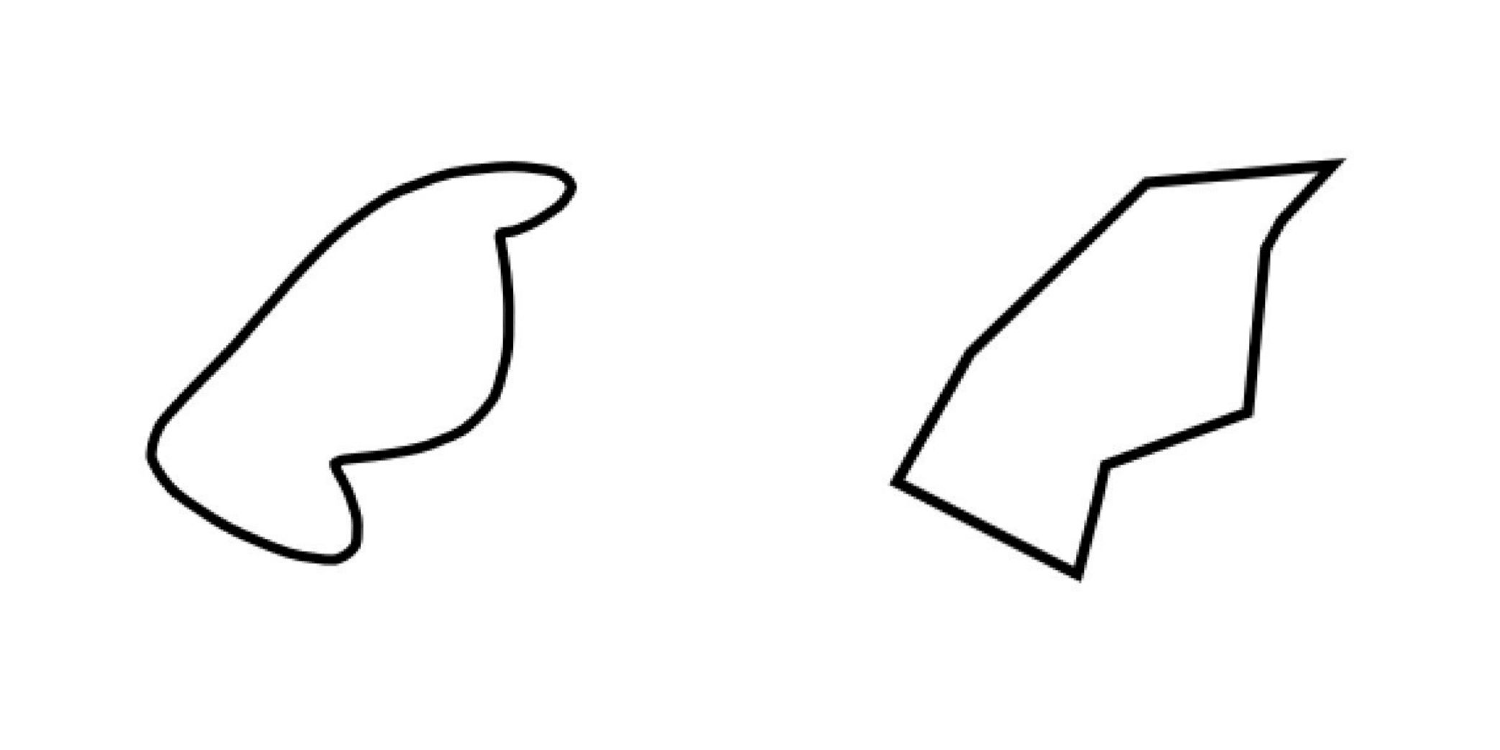}
    \includegraphics[width=0.25\textwidth]{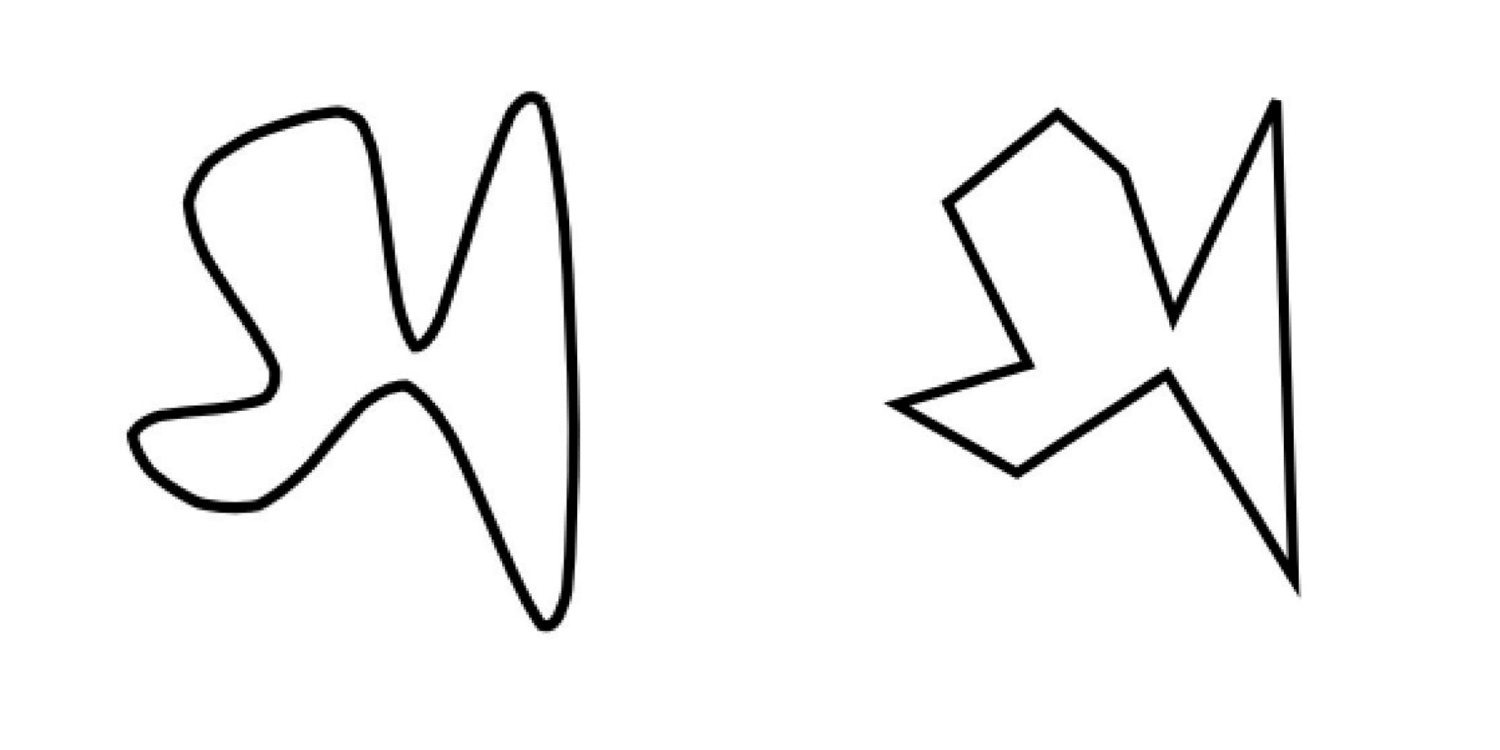}
    \includegraphics[width=0.25\textwidth]{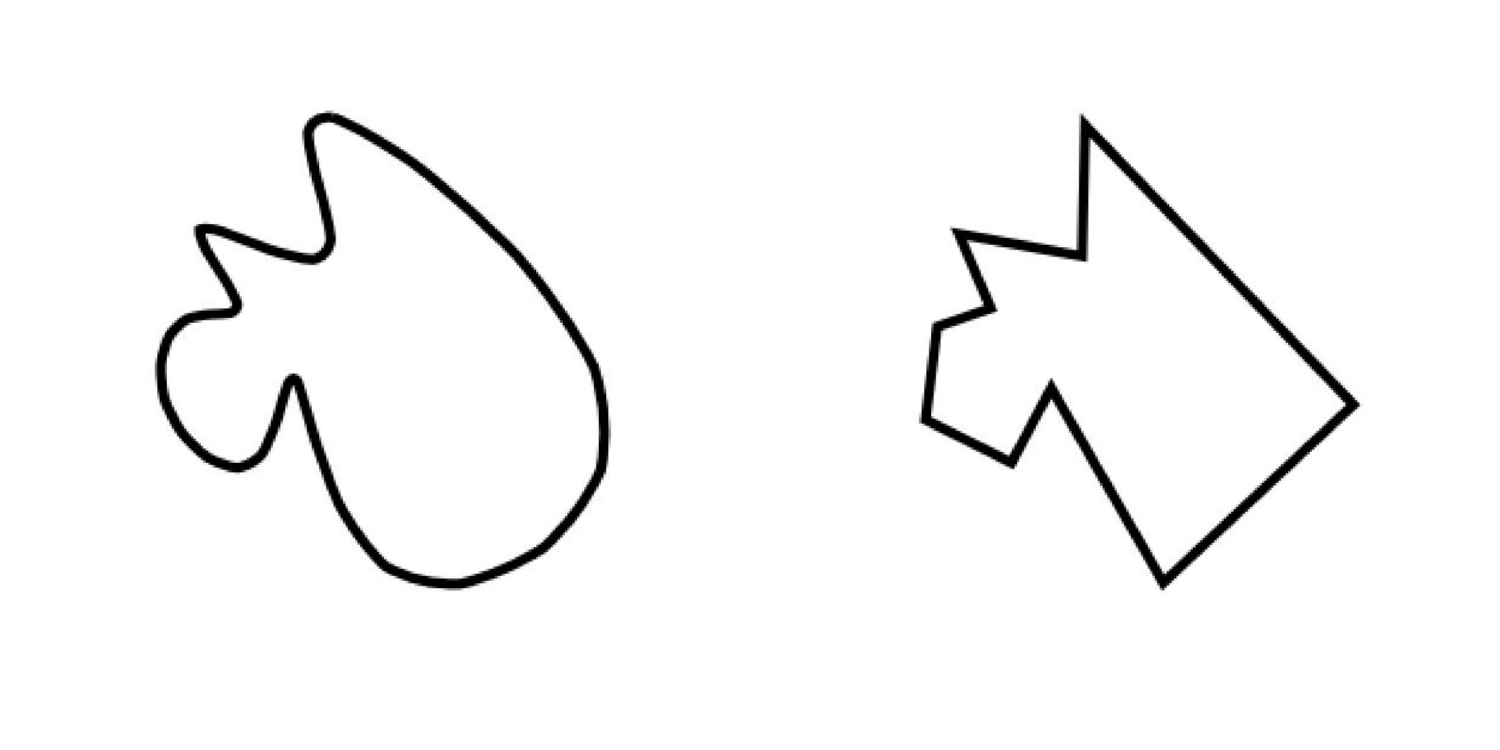}
    \caption{Image pairs from \citet{kouwenhovencross}}
    \label{kouwen_img}
\end{figure*}

Visual stimuli from prior works are in Figures \ref{kohler_img}, \ref{west_img}, \ref{maurer_img}, and \ref{kouwen_img}.

\section{Human Study details}
\label{sec:app_human_study_details}

\begin{figure*}
    \centering
    \includegraphics[width=0.9\linewidth]{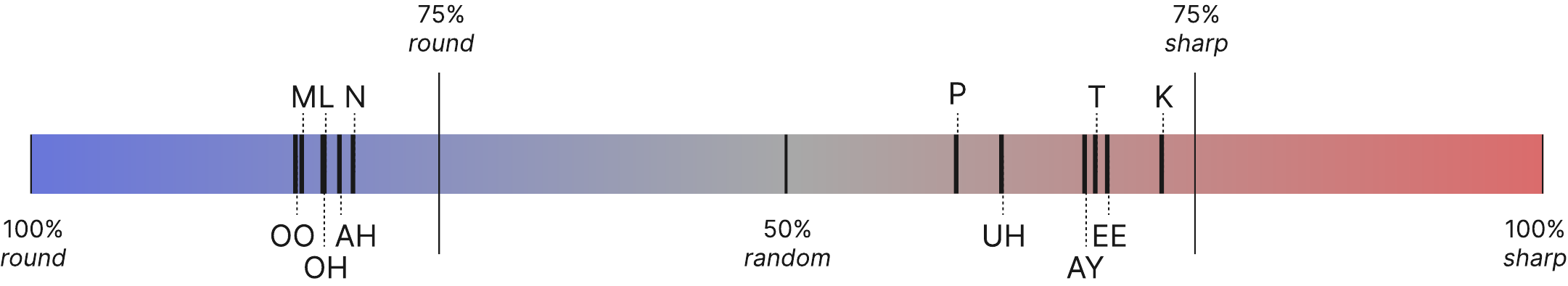}
    \caption{Human intended-category agreement by phoneme.}
    \label{fig:human-phoneme}
\end{figure*}

\subsection{Recruitment}
Participants were recruited through word of mouth and email lists, and included undergraduate and graduate students, and staff at our institution. 
Each participant was compensated with a \$15 gift card for the full study (30 minutes), as approved by our institution's IRB.

\subsection{Methodology}
We use a slide deck comprised of pairs of labels and images (shown in Figure~\ref{fig:stimuli-human-study}).
We provide a word label for each image pair and ask participants to choose the image that they thought better matched with the label.
Participants received the following verbal instructions prior to the task: 
``You will see a word followed by two images. 
Choose the image that best matches the word by saying A (left) or B (right).''
No additional written instruction screen was shown beyond the procedure described above.
For all labels, we ask them to read out loud and then verbally express their choice.

The study was divided into three equal-length sessions, with eye-tracker calibration before each session. 
At the end of each session, participants reported whether any pseudo-words in that session reminded them of real words or names; these items were excluded from analysis to minimize the influence of semantic associations. 
The study lasted approximately 30 minutes, participants were compensated with a \$15 gift card, and the protocol was approved by our institutional IRB.

During the studies, two authors listened to participants' comments and observed no offensive content. 
After data collection, all personally identifying information was removed, as we processed only user choice and gaze data. Audio recordings were collected only from participants who provided explicit consent on the consent form; participants who declined audio recording were not recorded. Gaze data and choice responses were retained in anonymized form, linked only to participant identifiers.

\subsection{Stimuli}
\label{sec:appendix_human_stimuli}
Each stimulus is presented via a slide-based format in which a word appears on one slide, followed immediately by the corresponding image pair on the next. As described in Section~\ref{sec:linguistic_stimuli}, the word set comprises 4 canonical pseudo-words, 162 phonemically constructed pseudo-words, and 20 adjectives. Stimuli are presented in the following order: pseudo-words first (canonical and phonemic combined, $n=166$), followed by adjectives ($n=20$). Within each block, the order of words is randomized across participants to counterbalance side bias. As described in Section~\ref{sec:visual_stimuli}, the image-pair set comprises 75 minimal line-drawing pairs, 92 language-model-generated pairs, and 17 pairs drawn from prior work. Two of these 17 original pairs were each used twice: one paired with \textit{bouba} and \textit{kiki}, and one with \textit{takete} and \textit{maluma}, thereby replicating the stimulus conditions of the original sound symbolism studies for our canonical word set. All image pairs were created in both left--right orientations (round shape on the left or right), yielding $186 \times 2 = 372$ total image instances, of which $184 \times 2 = 368$ are distinct. The four canonical words were fixed to their designated image pairs. The remaining 182 phonemically constructed pseudo-words and adjectives were then randomly mapped to the 364 non-canonical image pairs, with each word assigned to one instance of an image pair per participant. An example of stimuli is in Figure~\ref{fig:stimuli-human-study}.

\subsection{Human Gaze--Choice Coupling}
\label{sec:app_gaze_choice}
\begin{table}[t]
\centering
\small
\caption{Agreement (\%) between the image a participant attended to more and the image they chose, on the cleaned dataset (53 participants; 8{,}208 trials: 7{,}141 constructed pseudo-words, 910 adjectives, 157 canonical trials). Parentheses show matches/valid trials; trials tied in a measure are excluded only from that measure's denominator. Gaze samples and fixation durations are summed inside each image over the full image-display interval; the first-fixation row uses the first detected in-image fixation. These direct image-region statistics use no Gaussian smoothing or model token pooling. 95\% participant-cluster bootstrap intervals (20{,}000 resamples): 80.21 [77.02, 83.00], 76.97 [73.95, 79.75], 51.07 [49.70, 52.41].}
\label{tab:gaze_choice}
\resizebox{\linewidth}{!}{%
\begin{tabular}{lrrrr}
\toprule
``Attended to more'' defined by & Total & Pseudo-words & Adjectives & Canonical \\
\midrule
More gaze samples inside the image      & 80.21 (6,568/8,189) & 80.58 (5,742/7,126) & 77.59 (703/906) & 78.34 (123/157) \\
Longer fixation duration inside the image & 76.97 (6,318/8,208) & 77.37 (5,525/7,141) & 73.85 (672/910) & 77.07 (121/157) \\
Image containing the first fixation     & 51.07 (4,192/8,208) & 51.09 (3,648/7,141) & 51.76 (471/910) & 46.50 (73/157) \\
\bottomrule
\end{tabular}}
\end{table}
Because gaze is recorded on the same trial as the choice, we can ask whether where a participant looked relates to what they chose.
Raw gaze points are converted into fixation with an Identification by Dispersion Threshold (I-DT) procedure~\citep{10.1145/355017.355028} using a minimum duration of 100\, ms and a dispersion of at most 50 screen pixels, with recording gaps over 100\, ms breaking continuous runs.
\cref{tab:gaze_choice} defines ``the image attended to more'' in three ways: the image containing more exported gaze samples, the image with the longer summed fixation duration (assigned by fixation centroid), and the image containing the first detected in-image fixation.
We report how often each agrees with the participant's choice.
Accumulated gaze aligns strongly with their choices, whereas early gaze does not.
Under either accumulation measure, the image a participant looked at more is the image they chose in 77--80\% of trials, while the first in-image fixation agrees at chance level: the coupling reflects where gaze settles, not an initial side bias.
This first fixation is defined within the recorded image interval and may continue a fixation that began before image onset; it need not be the first newly initiated fixation after onset.

The image-pair slide was advanced immediately after each verbal response.
These statistics describe gaze--choice association and do not establish its causal direction.
The coupling replicates the well-documented tendency for gaze to accumulate on the option that is eventually chosen \citep{shimojo2003gaze, krajbich2010visual}, and it matters here for one reason: a participant's gaze carries information about that participant's choice on that trial, which is the premise for using it as a supervision signal in \cref{sec:finetune}.
This association motivates gaze supervision in \cref{sec:finetune}, as participants tend to look more at the image they choose.
Gaze--choice agreement alone does not establish that round- and sharp-associated words produce distinct patterns.
For the model comparisons, we later smooth gaze samples into density maps (\cref{sec:objective}); the statistics in \cref{tab:gaze_choice} use raw samples and fixations only.

\subsection{Participants and cleaning}
\label{sec:appendix_participants_and_cleaning}
We recruited 55 participants (undergraduate, graduate students and staff at our institution), yielding 9{,}688 recorded trials.
We retain trials with a verified complete image presentation, a valid A/B choice, and at least two fixations within the two images combined.
Gaze recording must also cover at least 80\% of the time the images were displayed.
The cleaned dataset contains 8{,}208 trials from 53 participants: 7{,}141 constructed pseudo-words, 910 adjectives, and 157 canonical trials. 
One participant had a calibration failure on gaze recording, and another participant's data was lost.
Among pseudo-word trials with valid responses, \catagr{} was 70.14\% (881/1{,}256) in the excluded trials and 73.98\% (5{,}283/7{,}141) in the retained trials; it was 73.41\% (6{,}164/8{,}397) before cleaning.
Unless stated otherwise, all human statistics use the cleaned dataset.

\section{Image Type and Caption Type Analysis}
\label{sec:appendix_image_caption_analysis}

Tables~\ref{tab:clip_caption_overall_ci}--\ref{tab:clip_caption_adjective_ci} and \ref{tab:siglip_so400m_caption_overall_ci}--\ref{tab:siglip_so400m_caption_adjective_ci} show the results split by caption, for CLIP (RN50) and SigLIP-so400m.

\begin{table*}[t]
\centering
\small
\setlength{\tabcolsep}{3pt}
\renewcommand{\arraystretch}{1.15}
\caption{
CLIP (RN50) caption comparison for the overall subset. 
Each cell reports the point estimate with the 95\% confidence interval below it.
Maj. = majority-choice match per label; $r$ = Pearson correlation; $\rho$ = Spearman correlation; ICA = intended-category agreement.
Maj. and ICA use 95\% Clopper--Pearson confidence intervals; 
correlation intervals use 95\% bootstrap confidence intervals over labels.
Asterisks indicate $p<.05$ under two-sided tests of zero Pearson/Spearman correlation.
}
\begin{tabular}{lcccc}
\toprule
Caption & Maj. & $r$ & $\rho$ & ICA \\
\midrule
\{label\} 
& \valci{0.57}{0.50}{0.65} 
& \valci{0.42$^{*}$}{0.29}{0.53} 
& \valci{0.42$^{*}$}{0.29}{0.53} 
& \valci{0.55}{0.54}{0.55} \\

The label for this image is \{label\} 
& \valci{0.54}{0.46}{0.61} 
& \valci{0.27$^{*}$}{0.11}{0.42} 
& \valci{0.27$^{*}$}{0.12}{0.42} 
& \valci{0.54}{0.53}{0.54} \\

This is a \{label\} 
& \valci{0.58}{0.51}{0.65} 
& \valci{0.36$^{*}$}{0.21}{0.50} 
& \valci{0.32$^{*}$}{0.17}{0.46} 
& \valci{0.55}{0.54}{0.55} \\

This looks like a \{label\} 
& \valci{0.63}{0.56}{0.70} 
& \valci{0.40$^{*}$}{0.25}{0.54} 
& \valci{0.37$^{*}$}{0.23}{0.51} 
& \valci{0.55}{0.55}{0.56} \\

An image of \{label\} 
& \valci{0.57}{0.50}{0.65} 
& \valci{0.49$^{*}$}{0.39}{0.59} 
& \valci{0.49$^{*}$}{0.38}{0.60} 
& \valci{0.56}{0.55}{0.56} \\

This is a \{label\} thing 
& \valci{0.60}{0.53}{0.67} 
& \valci{0.45$^{*}$}{0.32}{0.56} 
& \valci{0.44$^{*}$}{0.32}{0.56} 
& \valci{0.55}{0.55}{0.56} \\

A \{label\} object 
& \valci{0.56}{0.48}{0.63} 
& \valci{0.45$^{*}$}{0.30}{0.56} 
& \valci{0.48$^{*}$}{0.34}{0.59} 
& \valci{0.56}{0.55}{0.56} \\

A \{label\} shape 
& \valci{0.61}{0.54}{0.68} 
& \valci{0.51$^{*}$}{0.38}{0.61} 
& \valci{0.46$^{*}$}{0.33}{0.57} 
& \valci{0.57}{0.57}{0.58} \\

A \{label\} visual 
& \valci{0.53}{0.46}{0.61} 
& \valci{0.32$^{*}$}{0.18}{0.46} 
& \valci{0.32$^{*}$}{0.18}{0.46} 
& \valci{0.53}{0.53}{0.54} \\

A picture of a \{label\} object 
& \valci{0.56}{0.49}{0.64} 
& \valci{0.39$^{*}$}{0.23}{0.53} 
& \valci{0.43$^{*}$}{0.28}{0.55} 
& \valci{0.55}{0.54}{0.55} \\

\midrule
Combined 
& \valci{0.58}{0.50}{0.65} 
& \valci{0.42$^{*}$}{0.30}{0.53} 
& \valci{0.42$^{*}$}{0.29}{0.53} 
& \valci{0.55}{0.55}{0.55} \\
\bottomrule
\end{tabular}

\label{tab:clip_caption_overall_ci}
\end{table*}

\begin{table*}[t]
\centering
\small
\setlength{\tabcolsep}{3pt}
\renewcommand{\arraystretch}{1.15}
\caption{
CLIP (RN50) caption comparison for the pseudo-word subset. 
Each cell reports the point estimate with the 95\% confidence interval below it.
Maj. = majority-choice match per label; $r$ = Pearson correlation; $\rho$ = Spearman correlation; ICA = intended-category agreement.
Maj. and ICA use 95\% Clopper--Pearson confidence intervals; 
correlation intervals use 95\% bootstrap confidence intervals over labels.
Asterisks indicate $p<.05$ under two-sided tests of zero Pearson/Spearman correlation.
}
\begin{tabular}{lcccc}
\toprule
Caption & Maj. & $r$ & $\rho$ & ICA \\
\midrule
\{label\} 
& \valci{0.53}{0.45}{0.60} 
& \valci{0.20}{0.06}{0.33} 
& \valci{0.23$^{*}$}{0.08}{0.36} 
& \valci{0.52}{0.51}{0.52} \\

The label for this image is \{label\} 
& \valci{0.51}{0.43}{0.59} 
& \valci{0.21$^{*}$}{0.06}{0.34} 
& \valci{0.17$^{*}$}{0.02}{0.32} 
& \valci{0.52}{0.51}{0.53} \\

This is a \{label\} 
& \valci{0.54}{0.46}{0.62} 
& \valci{0.22$^{*}$}{0.08}{0.35} 
& \valci{0.18$^{*}$}{0.03}{0.32} 
& \valci{0.52}{0.52}{0.53} \\

This looks like a \{label\} 
& \valci{0.60}{0.53}{0.68} 
& \valci{0.23$^{*}$}{0.08}{0.37} 
& \valci{0.20$^{*}$}{0.06}{0.34} 
& \valci{0.53}{0.52}{0.53} \\

An image of \{label\} 
& \valci{0.53}{0.45}{0.60} 
& \valci{0.31$^{*}$}{0.18}{0.44} 
& \valci{0.33$^{*}$}{0.18}{0.46} 
& \valci{0.53}{0.52}{0.53} \\

This is a \{label\} thing 
& \valci{0.57}{0.49}{0.65} 
& \valci{0.31$^{*}$}{0.16}{0.45} 
& \valci{0.30$^{*}$}{0.16}{0.44} 
& \valci{0.53}{0.53}{0.54} \\

A \{label\} object 
& \valci{0.51}{0.43}{0.59} 
& \valci{0.29$^{*}$}{0.15}{0.44} 
& \valci{0.33$^{*}$}{0.19}{0.47} 
& \valci{0.53}{0.52}{0.53} \\

A \{label\} shape 
& \valci{0.58}{0.50}{0.66} 
& \valci{0.36$^{*}$}{0.23}{0.49} 
& \valci{0.36$^{*}$}{0.23}{0.49} 
& \valci{0.55}{0.54}{0.55} \\

A \{label\} visual 
& \valci{0.49}{0.41}{0.57} 
& \valci{0.15}{-0.00}{0.29} 
& \valci{0.14}{-0.02}{0.28} 
& \valci{0.51}{0.50}{0.51} \\

A picture of a \{label\} object 
& \valci{0.51}{0.43}{0.59} 
& \valci{0.25$^{*}$}{0.09}{0.39} 
& \valci{0.28$^{*}$}{0.12}{0.42} 
& \valci{0.52}{0.51}{0.52} \\

\midrule
Combined 
& \valci{0.53}{0.45}{0.60} 
& \valci{0.20$^{*}$}{0.06}{0.33} 
& \valci{0.23$^{*}$}{0.08}{0.36} 
& \valci{0.52}{0.52}{0.53} \\
\bottomrule
\end{tabular}

\label{tab:clip_caption_pseudoword_ci}
\end{table*}

\begin{table*}[t]
\centering
\small
\setlength{\tabcolsep}{3pt}
\renewcommand{\arraystretch}{1.15}
\caption{
CLIP (RN50) caption comparison for the adjective subset. 
Each cell reports the point estimate with the 95\% confidence interval below it.
Maj. = majority-choice match per label; $r$ = Pearson correlation; $\rho$ = Spearman correlation; ICA = intended-category agreement.
Maj. and ICA use 95\% Clopper--Pearson confidence intervals; 
correlation intervals use 95\% bootstrap confidence intervals over labels.
Asterisks indicate $p<.05$ under two-sided tests of zero Pearson/Spearman correlation.
}
\begin{tabular}{lcccc}
\toprule
Caption & Maj. & $r$ & $\rho$ & ICA \\
\midrule
\{label\} 
& \valci{0.90}{0.68}{0.99} 
& \valci{0.87$^{*}$}{0.76}{0.95} 
& \valci{0.73$^{*}$}{0.43}{0.85} 
& \valci{0.78}{0.77}{0.80} \\

The label for this image is \{label\} 
& \valci{0.75}{0.51}{0.91} 
& \valci{0.67$^{*}$}{0.44}{0.87} 
& \valci{0.53$^{*}$}{0.17}{0.76} 
& \valci{0.69}{0.67}{0.70} \\

This is a \{label\} 
& \valci{0.90}{0.68}{0.99} 
& \valci{0.84$^{*}$}{0.73}{0.93} 
& \valci{0.70$^{*}$}{0.35}{0.86} 
& \valci{0.75}{0.74}{0.76} \\

This looks like a \{label\} 
& \valci{0.90}{0.68}{0.99} 
& \valci{0.92$^{*}$}{0.86}{0.96} 
& \valci{0.77$^{*}$}{0.45}{0.91} 
& \valci{0.79}{0.77}{0.80} \\

An image of \{label\} 
& \valci{0.90}{0.68}{0.99} 
& \valci{0.87$^{*}$}{0.75}{0.97} 
& \valci{0.78$^{*}$}{0.48}{0.89} 
& \valci{0.79}{0.78}{0.81} \\

This is a \{label\} thing 
& \valci{0.85}{0.62}{0.97} 
& \valci{0.83$^{*}$}{0.72}{0.92} 
& \valci{0.68$^{*}$}{0.35}{0.83} 
& \valci{0.74}{0.73}{0.76} \\

A \{label\} object 
& \valci{0.90}{0.68}{0.99} 
& \valci{0.89$^{*}$}{0.78}{0.96} 
& \valci{0.74$^{*}$}{0.40}{0.87} 
& \valci{0.78}{0.77}{0.80} \\

A \{label\} shape 
& \valci{0.90}{0.68}{0.99} 
& \valci{0.90$^{*}$}{0.79}{0.97} 
& \valci{0.81$^{*}$}{0.59}{0.90} 
& \valci{0.79}{0.78}{0.80} \\

A \{label\} visual 
& \valci{0.85}{0.62}{0.97} 
& \valci{0.81$^{*}$}{0.67}{0.93} 
& \valci{0.67$^{*}$}{0.34}{0.81} 
& \valci{0.72}{0.71}{0.74} \\

A picture of a \{label\} object 
& \valci{1.00}{0.83}{1.00} 
& \valci{0.94$^{*}$}{0.89}{0.98} 
& \valci{0.78$^{*}$}{0.54}{0.89} 
& \valci{0.80}{0.78}{0.81} \\

\midrule
Combined 
& \valci{0.90}{0.69}{0.99} 
& \valci{0.87$^{*}$}{0.76}{0.95} 
& \valci{0.73$^{*}$}{0.43}{0.85} 
& \valci{0.76}{0.76}{0.77} \\
\bottomrule
\end{tabular}

\label{tab:clip_caption_adjective_ci}
\end{table*}

\begin{table*}[t]
\centering
\small
\setlength{\tabcolsep}{3pt}
\renewcommand{\arraystretch}{1.15}
\caption{
SigLIP-so400m caption comparison for the overall subset.
Each cell reports the point estimate with the 95\% confidence interval below it.
Maj. = majority-choice match per label; $r$ = Pearson correlation; $\rho$ = Spearman correlation; ICA = intended-category agreement.
Asterisks indicate two-sided approximate tests of zero correlation at $p < .05$.
}
\begin{tabular}{lcccc}
\toprule
Caption & Maj. & $r$ & $\rho$ & ICA \\
\midrule
\{label\}
& \valci{0.67}{0.59}{0.73}
& \valci{0.45$^{*}$}{0.31}{0.56}
& \valci{0.47$^{*}$}{0.34}{0.58}
& \valci{0.59}{0.58}{0.59} \\

The label for this image is \{label\}
& \valci{0.68}{0.61}{0.74}
& \valci{0.55$^{*}$}{0.42}{0.67}
& \valci{0.59$^{*}$}{0.47}{0.68}
& \valci{0.58}{0.57}{0.58} \\

This is a \{label\}
& \valci{0.69}{0.62}{0.75}
& \valci{0.48$^{*}$}{0.34}{0.61}
& \valci{0.54$^{*}$}{0.41}{0.64}
& \valci{0.57}{0.56}{0.58} \\

This looks like a \{label\}
& \valci{0.59}{0.52}{0.66}
& \valci{0.38$^{*}$}{0.23}{0.52}
& \valci{0.43$^{*}$}{0.30}{0.55}
& \valci{0.56}{0.55}{0.56} \\

An image of \{label\}
& \valci{0.73}{0.66}{0.79}
& \valci{0.53$^{*}$}{0.40}{0.65}
& \valci{0.55$^{*}$}{0.42}{0.66}
& \valci{0.60}{0.59}{0.60} \\

This is a \{label\} thing
& \valci{0.62}{0.54}{0.69}
& \valci{0.42$^{*}$}{0.28}{0.56}
& \valci{0.48$^{*}$}{0.34}{0.59}
& \valci{0.56}{0.55}{0.57} \\

A \{label\} object
& \valci{0.54}{0.47}{0.62}
& \valci{0.29$^{*}$}{0.14}{0.43}
& \valci{0.37$^{*}$}{0.22}{0.50}
& \valci{0.55}{0.54}{0.55} \\

A \{label\} shape
& \valci{0.66}{0.58}{0.72}
& \valci{0.39$^{*}$}{0.24}{0.54}
& \valci{0.42$^{*}$}{0.29}{0.56}
& \valci{0.57}{0.57}{0.58} \\

A \{label\} visual
& \valci{0.72}{0.65}{0.78}
& \valci{0.54$^{*}$}{0.42}{0.65}
& \valci{0.57$^{*}$}{0.47}{0.67}
& \valci{0.59}{0.58}{0.59} \\

A picture of a \{label\} object
& \valci{0.61}{0.53}{0.68}
& \valci{0.41$^{*}$}{0.27}{0.54}
& \valci{0.46$^{*}$}{0.32}{0.57}
& \valci{0.56}{0.55}{0.57} \\

\midrule
Combined
& \valci{0.67}{0.59}{0.73}
& \valci{0.45$^{*}$}{0.31}{0.56}
& \valci{0.47$^{*}$}{0.34}{0.58}
& \valci{0.57}{0.57}{0.57} \\
\bottomrule
\end{tabular}

\label{tab:siglip_so400m_caption_overall_ci}
\end{table*}

\begin{table*}[t]
\centering
\small
\setlength{\tabcolsep}{3pt}
\renewcommand{\arraystretch}{1.15}
\caption{
SigLIP-so400m caption comparison for the pseudo-word subset.
Each cell reports the point estimate with the 95\% confidence interval below it.
Maj. = majority-choice match per label; $r$ = Pearson correlation; $\rho$ = Spearman correlation; ICA = intended-category agreement.
Maj. and ICA use 95\% Clopper--Pearson confidence intervals; 
correlation intervals use 95\% bootstrap confidence intervals over labels.
Asterisks indicate $p<.05$ under two-sided tests of zero Pearson/Spearman correlation.
}
\begin{tabular}{lcccc}
\toprule
Caption & Maj. & $r$ & $\rho$ & ICA \\
\midrule
\{label\}
& \valci{0.66}{0.58}{0.73}
& \valci{0.47$^{*}$}{0.35}{0.59}
& \valci{0.47$^{*}$}{0.34}{0.58}
& \valci{0.58}{0.58}{0.59} \\

The label for this image is \{label\}
& \valci{0.67}{0.60}{0.74}
& \valci{0.57$^{*}$}{0.46}{0.66}
& \valci{0.54$^{*}$}{0.40}{0.64}
& \valci{0.56}{0.56}{0.57} \\

This is a \{label\}
& \valci{0.70}{0.62}{0.77}
& \valci{0.53$^{*}$}{0.42}{0.63}
& \valci{0.52$^{*}$}{0.39}{0.62}
& \valci{0.56}{0.56}{0.57} \\

This looks like a \{label\}
& \valci{0.59}{0.51}{0.66}
& \valci{0.37$^{*}$}{0.23}{0.48}
& \valci{0.35$^{*}$}{0.20}{0.49}
& \valci{0.55}{0.54}{0.55} \\

An image of \{label\}
& \valci{0.72}{0.64}{0.78}
& \valci{0.52$^{*}$}{0.40}{0.63}
& \valci{0.52$^{*}$}{0.39}{0.62}
& \valci{0.59}{0.58}{0.59} \\

This is a \{label\} thing
& \valci{0.62}{0.54}{0.69}
& \valci{0.39$^{*}$}{0.27}{0.50}
& \valci{0.38$^{*}$}{0.25}{0.50}
& \valci{0.55}{0.54}{0.55} \\

A \{label\} object
& \valci{0.53}{0.45}{0.61}
& \valci{0.24$^{*}$}{0.09}{0.37}
& \valci{0.30$^{*}$}{0.16}{0.44}
& \valci{0.53}{0.53}{0.54} \\

A \{label\} shape
& \valci{0.64}{0.56}{0.72}
& \valci{0.36$^{*}$}{0.21}{0.50}
& \valci{0.36$^{*}$}{0.21}{0.50}
& \valci{0.56}{0.55}{0.56} \\

A \{label\} visual
& \valci{0.72}{0.65}{0.79}
& \valci{0.60$^{*}$}{0.50}{0.69}
& \valci{0.59$^{*}$}{0.48}{0.69}
& \valci{0.58}{0.58}{0.59} \\

A picture of a \{label\} object
& \valci{0.59}{0.51}{0.66}
& \valci{0.35$^{*}$}{0.21}{0.48}
& \valci{0.36$^{*}$}{0.22}{0.49}
& \valci{0.54}{0.54}{0.55} \\

\midrule
Combined
& \valci{0.66}{0.58}{0.73}
& \valci{0.47$^{*}$}{0.35}{0.59}
& \valci{0.47$^{*}$}{0.34}{0.58}
& \valci{0.56}{0.56}{0.56} \\
\bottomrule
\end{tabular}

\label{tab:siglip_so400m_caption_pseudoword_ci}
\end{table*}

\begin{table*}[t]
\centering
\small
\setlength{\tabcolsep}{3pt}
\renewcommand{\arraystretch}{1.15}
\caption{
SigLIP-so400m caption comparison for the adjective subset.
Each cell reports the point estimate with the 95\% confidence interval below it.
Maj. = majority-choice match per label; $r$ = Pearson correlation; $\rho$ = Spearman correlation; ICA = intended-category agreement.
Maj. and ICA use 95\% Clopper--Pearson confidence intervals; 
correlation intervals use 95\% bootstrap confidence intervals over labels.
Asterisks indicate $p<.05$ under two-sided tests of zero Pearson/Spearman correlation.
}
\begin{tabular}{lcccc}
\toprule
Caption & Maj. & $r$ & $\rho$ & ICA \\
\midrule
\{label\}
& \valci{0.75}{0.51}{0.91}
& \valci{0.64$^{*}$}{0.33}{0.84}
& \valci{0.57$^{*}$}{0.12}{0.82}
& \valci{0.63}{0.61}{0.64} \\

The label for this image is \{label\}
& \valci{0.75}{0.51}{0.91}
& \valci{0.82$^{*}$}{0.68}{0.93}
& \valci{0.75$^{*}$}{0.48}{0.88}
& \valci{0.70}{0.69}{0.72} \\

This is a \{label\}
& \valci{0.65}{0.41}{0.85}
& \valci{0.70$^{*}$}{0.53}{0.83}
& \valci{0.63$^{*}$}{0.31}{0.81}
& \valci{0.63}{0.61}{0.64} \\

This looks like a \{label\}
& \valci{0.65}{0.41}{0.85}
& \valci{0.76$^{*}$}{0.65}{0.88}
& \valci{0.73$^{*}$}{0.46}{0.85}
& \valci{0.63}{0.62}{0.65} \\

An image of \{label\}
& \valci{0.90}{0.68}{0.99}
& \valci{0.80$^{*}$}{0.64}{0.91}
& \valci{0.69$^{*}$}{0.37}{0.85}
& \valci{0.68}{0.66}{0.69} \\

This is a \{label\} thing
& \valci{0.65}{0.41}{0.85}
& \valci{0.83$^{*}$}{0.72}{0.93}
& \valci{0.69$^{*}$}{0.42}{0.83}
& \valci{0.67}{0.65}{0.68} \\

A \{label\} object
& \valci{0.65}{0.41}{0.85}
& \valci{0.68$^{*}$}{0.51}{0.84}
& \valci{0.66$^{*}$}{0.34}{0.83}
& \valci{0.66}{0.65}{0.68} \\

A \{label\} shape
& \valci{0.85}{0.62}{0.97}
& \valci{0.80$^{*}$}{0.66}{0.92}
& \valci{0.73$^{*}$}{0.46}{0.89}
& \valci{0.70}{0.69}{0.72} \\

A \{label\} visual
& \valci{0.75}{0.51}{0.91}
& \valci{0.64$^{*}$}{0.48}{0.80}
& \valci{0.56$^{*}$}{0.18}{0.81}
& \valci{0.63}{0.62}{0.65} \\

A picture of a \{label\} object
& \valci{0.80}{0.56}{0.94}
& \valci{0.86$^{*}$}{0.76}{0.94}
& \valci{0.71$^{*}$}{0.46}{0.84}
& \valci{0.70}{0.68}{0.71} \\

\midrule
Combined
& \valci{0.75}{0.51}{0.91}
& \valci{0.64$^{*}$}{0.33}{0.84}
& \valci{0.57$^{*}$}{0.12}{0.82}
& \valci{0.66}{0.66}{0.67} \\
\bottomrule
\end{tabular}

\label{tab:siglip_so400m_caption_adjective_ci}
\end{table*}

Tables~\ref{tab:image_types_clip_res} and \ref{tab:image_types} show the results split by image types: line-based (original and generated minimal pairs) and model-based (Generated with Gemini), for CLIP (RN50) and SigLIP-so400m.

\begin{table*}[t]
\centering
\scriptsize
\setlength{\tabcolsep}{1.3pt}
\sisetup{
  table-number-alignment = center,
  round-mode = places,
  round-precision = 3
}
\caption{
Model--human alignment by subset for CLIP (RN50), comparing image types. 
Maj. = majority-choice match per label; $r$ = Pearson correlation; $\rho$ = Spearman correlation; ICA = intended-category agreement.
Asterisks indicate $p<.05$ under two-sided tests of zero Pearson/Spearman correlation.
}
\begin{tabular}{
  l
  S[table-format=1.3] S[table-format=-1.3] S[table-format=-1.3] S[table-format=-1.3]
  @{\hspace{7pt}}
  S[table-format=1.3] S[table-format=-1.3] S[table-format=-1.3] S[table-format=-1.3]
  @{\hspace{7pt}}
  S[table-format=1.3] S[table-format=-1.3] S[table-format=-1.3] S[table-format=-1.3]
}
\hline
& \multicolumn{4}{c}{Overall} 
& \multicolumn{4}{c}{Pseudo-word} 
& \multicolumn{4}{c}{Adjective} \\
\cmidrule(lr){2-5}
\cmidrule(lr){6-9}
\cmidrule(lr){10-13}
Image Type 
& \multicolumn{1}{c}{Maj.} 
& \multicolumn{1}{c}{$r$} 
& \multicolumn{1}{c}{$\rho$} 
& \multicolumn{1}{c}{ICA}
& \multicolumn{1}{c}{Maj.} 
& \multicolumn{1}{c}{$r$} 
& \multicolumn{1}{c}{$\rho$} 
& \multicolumn{1}{c}{ICA}
& \multicolumn{1}{c}{Maj.} 
& \multicolumn{1}{c}{$r$} 
& \multicolumn{1}{c}{$\rho$} 
& \multicolumn{1}{c}{ICA} \\
\hline
Lines & 0.551 & 0.425$^{*}$ & 0.423$^{*}$ & 0.558 & 0.512 & 0.277$^{*}$ & 0.269$^{*}$ & 0.530 & 0.857 & 0.868$^{*}$ & 0.701$^{*}$ & 0.787 \\
Realistic & 0.577 & 0.408$^{*}$ & 0.333$^{*}$ & 0.542 & 0.543 & 0.222$^{*}$ & 0.170$^{*}$ & 0.518 & 0.850 & 0.820$^{*}$ & 0.731$^{*}$ & 0.741\\
\hline
Combined      & 0.575 & 0.419$^{*}$ & 0.418$^{*}$ & 0.550 & 0.525 & 0.202$^{*}$ & 0.227$^{*}$ & 0.524 & 0.900 & 0.868$^{*}$ & 0.726$^{*}$ & 0.764 \\
\hline
\end{tabular}

\label{tab:image_types_clip_res}
\end{table*}


\begin{table*}[t]
\centering
\scriptsize
\setlength{\tabcolsep}{1.3pt}
\sisetup{
  table-number-alignment = center,
  round-mode = places,
  round-precision = 3
}
\caption{
Model--human alignment by subset for SigLIP-so400m, comparing image types. 
Maj. = majority-choice match per label; $r$ = Pearson correlation; $\rho$ = Spearman correlation; ICA = intended-category agreement.
Asterisks indicate $p<.05$ under two-sided tests of zero Pearson/Spearman correlation.
}
\begin{tabular}{
  l
  S[table-format=1.3] S[table-format=-1.3] S[table-format=-1.3] S[table-format=-1.3]
  @{\hspace{7pt}}
  S[table-format=1.3] S[table-format=-1.3] S[table-format=-1.3] S[table-format=-1.3]
  @{\hspace{7pt}}
  S[table-format=1.3] S[table-format=-1.3] S[table-format=-1.3] S[table-format=-1.3]
}
\hline
& \multicolumn{4}{c}{Overall} 
& \multicolumn{4}{c}{Pseudo-word} 
& \multicolumn{4}{c}{Adjective} \\
\cmidrule(lr){2-5}
\cmidrule(lr){6-9}
\cmidrule(lr){10-13}
Image Type 
& \multicolumn{1}{c}{Maj.} 
& \multicolumn{1}{c}{$r$} 
& \multicolumn{1}{c}{$\rho$} 
& \multicolumn{1}{c}{ICA}
& \multicolumn{1}{c}{Maj.} 
& \multicolumn{1}{c}{$r$} 
& \multicolumn{1}{c}{$\rho$} 
& \multicolumn{1}{c}{ICA}
& \multicolumn{1}{c}{Maj.} 
& \multicolumn{1}{c}{$r$} 
& \multicolumn{1}{c}{$\rho$} 
& \multicolumn{1}{c}{ICA} \\
\hline
Lines & 0.647 & 0.393$^{*}$ & 0.418$^{*}$ & 0.561 & 0.648 & 0.362$^{*}$ & 0.370$^{*}$ & 0.550 & 0.667 & 0.661$^{*}$ & 0.644$^{*}$ & 0.654
\\
Realistic & 0.769 & 0.663$^{*}$ & 0.657$^{*}$ & 0.583 & 0.753 & 0.604$^{*}$ & 0.603$^{*}$ & 0.572 & 0.900 & 0.846$^{*}$ & 0.763$^{*}$ & 0.672 \\
\hline
Combined & 0.667 & 0.452$^{*}$ & 0.473$^{*}$ & 0.572 & 0.660 & 0.469$^{*}$ & 0.471$^{*}$ & 0.561 & 0.750 & 0.639$^{*}$ & 0.569$^{*}$ & 0.663 \\
\hline
\end{tabular}

\label{tab:image_types}
\end{table*}

\section{Full Statistics for Table~\ref{tab:model_results}}
\label{sec:appendix_full_stats}
Tables~\ref{tab:appendix_overall_stats}--\ref{tab:appendix_adjective_stats} report the same cleaned-cohort point estimates as Table~\ref{tab:model_results}, with 95\% confidence intervals.
The six embedding models, six completed original-prompt generative models, and the two neutral-prompt runs each cover 8{,}208 trials from 53 participants: 7{,}141 pseudo-word trials (162 labels), 910 adjective trials (20 labels), and 157 canonical trials (4 labels).
Overall includes all 186 labels; the pseudo-word subset excludes canonical labels.
For embedding models, each human trial is matched to all ten captions, and ICA and label-level round-choice rates use the same trial--caption weighting.
Maj. compares human and model majority choices per label, treating a rate of 0.5 as a tie; ties match only when both rates are tied. Generative model round-choice rates average hard A/B decisions derived from FP32 logits over the corresponding human trials; an exact A/B logit tie contributes 0.5. These rates are not averages of $P(\mathrm{round})$.
Confidence intervals use 10,000 participant-cluster bootstrap resamples with seed 20260922, applying the same participant weights to all models and retaining the observed labels, images, and captions. Replicates missing any observed label are omitted for Maj., $r$, and $\rho$; undefined correlations are also omitted. These intervals are conditional on the observed stimulus inventory and checkpoint and do not measure generalization to new labels or variation across training seeds.
Because a resample repeats some participants and omits others, label-level human rates are noisier within resamples than in the full sample, which attenuates label-level correlations; 
percentile bootstrap intervals need not contain the observed point estimate, as illustrated by Qwen3-VL-30B's pseudo-word correlation ($r=.571$, 95\% CI $[.414,.566]$).

\begin{table*}[t]
\centering
\scriptsize
\setlength{\tabcolsep}{3pt}
\caption{
Uncertainty statistics for the overall subset: 8,208 trials and 186 labels.
Maj. = majority-choice match per label; $r$ = Pearson correlation;
$\rho$ = Spearman correlation; ICA = intended-category agreement.
All intervals are 95\% percentile participant-cluster bootstrap intervals (10,000 resamples of the 53 participants). 
Human and model label rates and ICA are recomputed in each resample.
Asterisks indicate nominal $p<.05$ in two-sided label-level correlation tests, unadjusted for multiple comparisons; these tests differ from the bootstrap intervals. Dashes indicate unavailable runs.
}
\begin{tabular}{
  @{}
  l
  c c
  c c
  c c
  c c
  @{}
}
\toprule
Model 
& Maj. & 95\% CI
& $r$ & 95\% CI
& $\rho$ & 95\% CI
& ICA & 95\% CI \\
\midrule
CLIP RN50 & 0.618 & [0.559, 0.640] & 0.452$^{*}$ & [0.374, 0.474] & 0.424$^{*}$ & [0.334, 0.446] & 0.548 & [0.541, 0.555] \\

CLIP ViT-B/32 & 0.597 & [0.548, 0.629] & 0.375$^{*}$ & [0.296, 0.402] & 0.344$^{*}$ & [0.252, 0.377] & 0.534 & [0.527, 0.540] \\

SigLIP-base & 0.613 & [0.565, 0.640] & 0.381$^{*}$ & [0.314, 0.411] & 0.372$^{*}$ & [0.296, 0.408] & 0.548 & [0.540, 0.556] \\

SigLIP-so400m & 0.731 & [0.667, 0.753] & 0.534$^{*}$ & [0.438, 0.547] & 0.603$^{*}$ & [0.474, 0.601] & 0.574 & [0.566, 0.582] \\

SigLIP2-base & 0.495 & [0.446, 0.516] & 0.324$^{*}$ & [0.240, 0.354] & 0.162$^{*}$ & [0.110, 0.262] & 0.531 & [0.524, 0.538] \\

SigLIP2-so400m & 0.478 & [0.446, 0.559] & 0.018 & [-0.079, 0.110] & -0.043 & [-0.109, 0.074] & 0.502 & [0.493, 0.510] \\

\midrule
Gemma4-E4B & 0.565 & [0.527, 0.624] & 0.447$^{*}$ & [0.350, 0.464] & 0.326$^{*}$ & [0.259, 0.395] & 0.554 & [0.545, 0.564] \\

Gemma4-26B & 0.758 & [0.694, 0.763] & 0.696$^{*}$ & [0.626, 0.704] & 0.695$^{*}$ & [0.612, 0.701] & 0.663 & [0.654, 0.672] \\

Qwen3-VL-8B & 0.613 & [0.543, 0.651] & 0.527$^{*}$ & [0.421, 0.531] & 0.456$^{*}$ & [0.314, 0.471] & 0.566 & [0.556, 0.575] \\

Qwen3-VL-30B & 0.715 & [0.661, 0.747] & 0.707$^{*}$ & [0.612, 0.702] & 0.693$^{*}$ & [0.563, 0.689] & 0.613 & [0.603, 0.623] \\

Molmo2-4B & 0.570 & [0.500, 0.613] & 0.417$^{*}$ & [0.251, 0.412] & 0.349$^{*}$ & [0.173, 0.367] & 0.530 & [0.519, 0.541] \\

Molmo2-8B & 0.656 & [0.597, 0.694] & 0.622$^{*}$ & [0.514, 0.618] & 0.580$^{*}$ & [0.429, 0.577] & 0.584 & [0.573, 0.594] \\

\midrule
\multicolumn{9}{@{}l}{\emph{Neutral prompt}} \\
Gemma4-26B & 0.726 & [0.677, 0.753] & 0.678$^{*}$ & [0.599, 0.683] & 0.656$^{*}$ & [0.566, 0.663] & 0.625 & [0.614, 0.636] \\
Qwen3-VL-30B & 0.624 & [0.608, 0.677] & 0.665$^{*}$ & [0.587, 0.667] & 0.644$^{*}$ & [0.550, 0.657] & 0.609 & [0.600, 0.618] \\
\bottomrule
\end{tabular}

\label{tab:appendix_overall_stats}
\end{table*}

\begin{table*}[t]
\centering
\scriptsize
\setlength{\tabcolsep}{3pt}
\caption{
Uncertainty statistics for the pseudo-word subset: 7,141 trials and 162 labels.
Maj. = majority-choice match per label; $r$ = Pearson correlation; 
$\rho$ = Spearman correlation; ICA = intended-category agreement. 
All intervals are 95\% percentile participant-cluster bootstrap intervals (10,000 resamples of the 53 participants). Human and model label rates and ICA are recomputed in each resample.
Asterisks indicate nominal $p<.05$ in two-sided label-level correlation tests, unadjusted for multiple comparisons; these tests differ from the bootstrap intervals. Dashes indicate unavailable runs.
}
\begin{tabular}{
  @{}
  l
  c c
  c c
  c c
  c c
  @{}
}
\toprule
Model 
& Maj. & 95\% CI
& $r$ & 95\% CI
& $\rho$ & 95\% CI
& ICA & 95\% CI \\
\midrule
CLIP RN50 & 0.580 & [0.518, 0.611] & 0.285$^{*}$ & [0.192, 0.318] & 0.274$^{*}$ & [0.168, 0.310] & 0.520 & [0.513, 0.528] \\

CLIP ViT-B/32 & 0.549 & [0.494, 0.586] & 0.116 & [0.036, 0.170] & 0.132 & [0.031, 0.192] & 0.505 & [0.497, 0.512] \\

SigLIP-base & 0.574 & [0.519, 0.611] & 0.209$^{*}$ & [0.136, 0.256] & 0.197$^{*}$ & [0.114, 0.259] & 0.521 & [0.512, 0.529] \\

SigLIP-so400m & 0.716 & [0.648, 0.747] & 0.560$^{*}$ & [0.427, 0.566] & 0.561$^{*}$ & [0.419, 0.568] & 0.565 & [0.557, 0.573] \\

SigLIP2-base & 0.438 & [0.383, 0.463] & -0.095 & [-0.165, 0.038] & -0.121 & [-0.174, 0.042] & 0.497 & [0.489, 0.506] \\

SigLIP2-so400m & 0.500 & [0.451, 0.580] & 0.011 & [-0.101, 0.120] & -0.019 & [-0.115, 0.103] & 0.503 & [0.494, 0.512] \\

\midrule
Gemma4-E4B & 0.512 & [0.469, 0.580] & 0.153 & [0.031, 0.206] & 0.085 & [0.002, 0.193] & 0.511 & [0.501, 0.522] \\

Gemma4-26B & 0.728 & [0.654, 0.741] & 0.580$^{*}$ & [0.493, 0.594] & 0.579$^{*}$ & [0.470, 0.592] & 0.622 & [0.612, 0.632] \\

Qwen3-VL-8B & 0.562 & [0.488, 0.605] & 0.262$^{*}$ & [0.078, 0.279] & 0.242$^{*}$ & [0.054, 0.275] & 0.516 & [0.505, 0.526] \\

Qwen3-VL-30B & 0.679 & [0.617, 0.716] & 0.571$^{*}$ & [0.414, 0.566] & 0.562$^{*}$ & [0.387, 0.564] & 0.566 & [0.554, 0.577] \\

Molmo2-4B & 0.525 & [0.451, 0.574] & 0.131 & [-0.021, 0.186] & 0.143 & [-0.025, 0.189] & 0.507 & [0.496, 0.518] \\

Molmo2-8B & 0.605 & [0.543, 0.654] & 0.413$^{*}$ & [0.218, 0.410] & 0.390$^{*}$ & [0.190, 0.400] & 0.532 & [0.521, 0.543] \\
\midrule
\multicolumn{9}{@{}l}{\emph{Neutral prompt}} \\
Gemma4-26B & 0.704 & [0.642, 0.728] & 0.534$^{*}$ & [0.426, 0.549] & 0.516$^{*}$ & [0.400, 0.535] & 0.582 & [0.569, 0.594] \\
Qwen3-VL-30B & 0.574 & [0.562, 0.642] & 0.535$^{*}$ & [0.410, 0.541] & 0.524$^{*}$ & [0.400, 0.546] & 0.566 & [0.556, 0.575] \\
\bottomrule
\end{tabular}

\label{tab:appendix_pseudoword_stat}
\end{table*}

\begin{table*}[t]
\centering
\scriptsize
\setlength{\tabcolsep}{3pt}
\caption{
Uncertainty statistics for the adjective subset: 910 trials and 20 labels.
Maj. = majority-choice match per label; $r$ = Pearson correlation; 
$\rho$ = Spearman correlation; ICA = intended-category agreement. 
All intervals are 95\% percentile participant-cluster bootstrap intervals (10,000 resamples of the 53 participants). Human and model label rates and ICA are recomputed in each resample.
Asterisks indicate nominal $p<.05$ in two-sided label-level correlation tests, unadjusted for multiple comparisons; these tests differ from the bootstrap intervals. Dashes indicate unavailable runs.
}
\begin{tabular}{
  @{}
  l
  c c
  c c
  c c
  c c
  @{}
}
\toprule
Model 
& Maj. & 95\% CI
& $r$ & 95\% CI
& $\rho$ & 95\% CI
& ICA & 95\% CI \\
\midrule
CLIP RN50 & 0.950 & [0.850, 1.000] & 0.879$^{*}$ & [0.829, 0.901] & 0.715$^{*}$ & [0.667, 0.810] & 0.750 & [0.733, 0.768] \\

CLIP ViT-B/32 & 1.000 & [0.900, 1.000] & 0.918$^{*}$ & [0.863, 0.942] & 0.752$^{*}$ & [0.702, 0.853] & 0.760 & [0.739, 0.781] \\

SigLIP-base & 0.950 & [0.850, 1.000] & 0.917$^{*}$ & [0.872, 0.935] & 0.847$^{*}$ & [0.750, 0.903] & 0.762 & [0.744, 0.780] \\

SigLIP-so400m & 0.900 & [0.700, 0.950] & 0.819$^{*}$ & [0.704, 0.861] & 0.685$^{*}$ & [0.568, 0.815] & 0.657 & [0.634, 0.679] \\

SigLIP2-base & 0.950 & [0.850, 1.000] & 0.923$^{*}$ & [0.884, 0.940] & 0.756$^{*}$ & [0.713, 0.847] & 0.786 & [0.766, 0.807] \\

SigLIP2-so400m & 0.350 & [0.300, 0.600] & -0.022 & [-0.259, 0.211] & -0.032 & [-0.225, 0.256] & 0.497 & [0.477, 0.517] \\

\midrule
Gemma4-E4B & 1.000 & [0.950, 1.000] & 0.992$^{*}$ & [0.973, 0.994] & 0.854$^{*}$ & [0.780, 0.915] & 0.909 & [0.891, 0.927] \\

Gemma4-26B & 1.000 & [1.000, 1.000] & 0.994$^{*}$ & [0.981, 0.996] & 0.775$ƒp^{*}$ & [0.734, 0.875] & 0.970 & [0.960, 0.980] \\

Qwen3-VL-8B & 1.000 & [0.950, 1.000] & 0.990$^{*}$ & [0.973, 0.993] & 0.829$^{*}$ & [0.750, 0.891] & 0.942 & [0.928, 0.955] \\

Qwen3-VL-30B & 0.950 & [0.950, 1.000] & 0.985$^{*}$ & [0.964, 0.992] & 0.905$^{*}$ & [0.818, 0.940] & 0.937 & [0.925, 0.950] \\

Molmo2-4B & 0.950 & [0.800, 1.000] & 0.866$^{*}$ & [0.760, 0.901] & 0.817$^{*}$ & [0.674, 0.876] & 0.716 & [0.688, 0.745] \\

Molmo2-8B & 1.000 & [1.000, 1.000] & 0.987$^{*}$ & [0.970, 0.992] & 0.825$^{*}$ & [0.761, 0.888] & 0.936 & [0.920, 0.951] \\
\midrule
\multicolumn{9}{@{}l}{\emph{Neutral prompt}} \\
Gemma4-26B & 0.950 & [0.950, 1.000] & 0.982$^{*}$ & [0.960, 0.991] & 0.771$^{*}$ & [0.735, 0.852] & 0.945 & [0.930, 0.958] \\
Qwen3-VL-30B & 0.950 & [0.950, 1.000] & 0.983$^{*}$ & [0.962, 0.991] & 0.869$^{*}$ & [0.809, 0.911] & 0.945 & [0.933, 0.957] \\
\bottomrule
\end{tabular}

\label{tab:appendix_adjective_stats}
\end{table*}

\subsection{Neutral-prompt robustness and paired changes}
\label{sec:appendix_neutral_results}
Both large models were evaluated on the same 7,583 unique inputs, mapped to all 8,208 human trials, under original and neutral prompts. The checkpoint revision, processor, $512\times512$ input, eager backend, and FP32 A/B projection were identical within each model; Gemma's output softcap was retained. 

\begin{figure}[ht]
    \centering
    \begin{tcolorbox}[colback=gray!5!white, colframe=gray!75!black]
        You will see one image containing two options:\\
        - Image A is on the LEFT.\\
        - Image B is on the RIGHT.\\
        
        Label: ``\{word\}''\\
        
        Choose the image that better matches this word.\\
        
        Choose exactly one option, A or B. Return only valid JSON:\\
        \{``choice'':``A''\}\\ 
        or\\ 
        \{``choice'':``B''\}
    \end{tcolorbox}
    \caption{Neutral prompt used for forced choice analysis.}
    \label{fig:llm-prompt-forced-neutral}
\end{figure}


This replaces the original task/rule block and omits the label type, including the instruction to judge pseudowords by ``visual/sound-symbolic fit.'' It is a comparison of two prompt formulations, not an isolated intervention on one phrase. Neutral runs record choices only; they do not provide a neutral-prompt spatial-alignment comparison.

Table~\ref{tab:model_results} and Tables~\ref{tab:appendix_overall_stats}--\ref{tab:appendix_adjective_stats} report the completed neutral point estimates and participant-bootstrap intervals. On pseudo-words, neutral Pearson correlations are $.534$ for Gemma4-26B (95\% CI $[.426,.549]$) and $.535$ for Qwen3-VL-30B ($[.410,.541]$). Neutral adjective ICA remains high at $.945$ for both models (original: $.970$ and $.937$, respectively).

For the exploratory paired contrasts in Table~\ref{tab:neutral_paired_changes}, we resample the 53 participants 10,000 times with replacement (seed 20260922), applying identical weights to original and neutral trials and to both models. We recompute each prompt's human/model label rates and trial-average metrics before taking the neutral-minus-original difference. These are nominal 95\% percentile intervals, conditional on the observed labels, images, and checkpoints; they are not adjusted for multiple comparisons. They differ from the word--pair--participant bootstrap used for the spatial Center contrasts. Marginal confidence-interval overlap is not used as a test of prompt differences.

Gemma4-26B's pseudo-word ICA changes by $-3.99$ pp (paired 95\% CI $[-5.29,-2.74]$), while its agreement with the actual participant response changes from $57.30\%$ to $57.74\%$ ($+0.43$ pp, $[-0.96,1.84]$). Qwen3-VL-30B's ICA has the same point estimate under both prompts, but its actual-response agreement changes from $55.73\%$ to $57.72\%$ ($+1.99$ pp, $[0.76,3.22]$). Neither Pearson-$r$ change has an interval excluding zero; these intervals do not establish equivalence. The original/neutral choice differs on 33.23\% of the 6,720 unique pseudo-word inputs for Gemma4-26B and 20.55\% for Qwen3-VL-30B (trial-weighted: 33.26\% and 20.70\%). In particular, unchanged aggregate Qwen ICA does not mean unchanged individual decisions. Together, these results support persistence of positive choice alignment under the neutral formulation, while qualifying any claim of prompt invariance.

\begin{table}[t]
\centering\footnotesize
\setlength{\tabcolsep}{5pt}
\caption{Exploratory neutral-minus-original changes on the same 7,141 pseudo-word trials. Intervals use paired participant bootstrap and are nominal 95\% intervals without multiplicity adjustment. Maj., ICA, and human agreement differences are percentage points (pp); correlation differences are on their original scale. Human agreement scores the actual participant response on each trial.}
\label{tab:neutral_paired_changes}
\begin{tabular}{@{}llrr@{}}
\toprule
Model & Metric & Difference & Paired 95\% CI \\
\midrule
Gemma4-26B & Maj. (pp) & $-2.47$ & $[-6.79, 4.94]$ \\
Gemma4-26B & $r$ & $-0.047$ & $[-0.113, 0.001]$ \\
Gemma4-26B & $\rho$ & $-0.063$ & $[-0.123, -0.002]$ \\
Gemma4-26B & ICA (pp) & $-3.99$ & $[-5.29, -2.74]$ \\
Gemma4-26B & Human agreement (pp) & $+0.43$ & $[-0.96, 1.84]$ \\
\midrule
Qwen3-VL-30B & Maj. (pp) & $-10.49$ & $[-12.35, -1.85]$ \\
Qwen3-VL-30B & $r$ & $-0.037$ & $[-0.068, 0.039]$ \\
Qwen3-VL-30B & $\rho$ & $-0.038$ & $[-0.066, 0.058]$ \\
Qwen3-VL-30B & ICA (pp) & $+0.00$ & $[-0.93, 0.93]$ \\
Qwen3-VL-30B & Human agreement (pp) & $+1.99$ & $[0.76, 3.22]$ \\
\bottomrule
\end{tabular}
\end{table}

\section{Attention Alignment}
\label{sec:attention}

\subsection{Choice evaluation}
Generative models use the original forced-choice prompt (Appendix~\ref{sec:appendix_prompt_forced}), as in fine-tuning, with the composite panel resized to $512\times512$ before native processing. 
All six completed generative runs use $512\times512$ inputs (resolution sensitivity: \emph{Input resolution} below).
Generative choices and attention come from one eager forward pass per input. 
A/B probabilities use FP32 projections of the final hidden state, with Gemma's output softcap retained; remaining exact ties count as half in trial agreement measures.
ICA and same-trial human agreement describe individual responses; label-level majority match and correlations compare response distributions. 
Overall, pseudo-word, and adjective summaries contain 186, 162, and 20 labels, respectively; the four canonical labels enter Overall only. 

\subsection{Trial-wise native-grid spatial evaluation}
\label{sec:appendix_human_model_alignment_methods}
\label{sec:appendix_molmo_gemma}
The full-cohort evaluation contains 8,208 human trials and 7,583 unique model inputs; the main spatial table uses 7,141 constructed pseudo-word trials (6,720 unique inputs). 
Each model map is matched back to every corresponding human trial. We do not average maps across participants or image--category cells before computing the metrics.

\paragraph{Generative readout and layers.}
Each generative model receives the original forced-choice prompt, and we read attention from the last prompt position, the position whose output predicts the A/B answer; the answer token itself is not part of the input.
This is the \dta{} readout, and it is the same attention query used in fine-tuning.
Maps are averaged over heads and over full-attention layers with zero-based index at least $\lfloor L/2\rfloor$; for Gemma, only the global-attention layers in that half contribute.
The resulting layers are L18--35 for Qwen3-VL-8B, Molmo2-4B, and Molmo2-8B; L24--47 for Qwen3-VL-30B; L23/29/35/41 for Gemma4-E4B; and L17/23/29 for Gemma4-26B.

Inputs are resized to $512\times512$ before each model's standard image preprocessing.
At this size, Qwen and Gemma produce a $16\times16$ image-token grid; Molmo uses its global and high-resolution tokens together (725 image tokens), as in fine-tuning, with each token mapped to its actual image region.
Model maps are compared on these native cells without upsampling, and tokens that straddle the boundary between the two images are excluded.

A/B logits are computed in FP32 from the final hidden state using the two corresponding rows of the output projection, with Gemma's final-logit softcap retained; exact ties count as half in agreement measures.
Because choices on near-indifferent items can depend on the numerical backend, choices and attention maps for each input come from a single forward pass with a fixed eager backend, and all generative entries in Table~\ref{tab:model_results}, the Choice $r$ entries in Table~\ref{tab:human_model_attention_alignment_combined}, and the participant-bootstrap intervals in Appendix~\ref{sec:appendix_full_stats} are computed from these choices.
The neutral-prompt runs (Gemma4-26B and Qwen3-VL-30B; Section~\ref{sec:neutral_prompt}) use the same input size and backend but record choices only, so all spatial results use the original prompt.

\paragraph{Embedding readouts and coordinates.}
All six embedding models use the fixed caption ``The label for this image is \{word\}'' and the original composite panel resized to $512\times512$ before native image processing. CLIP RN50 uses layer4 feature Grad-CAM; CLIP ViT-B/32 uses positive final-layer attention--gradient products, head-averaged at CLS-to-patch positions. SigLIP and SigLIP2 use feature Grad-CAM at vision post-layernorm with gradients of their native pooled image--text cosine score. Native positive maps and coordinates are saved directly. CLIP uses normalized feature-cell bins, not receptive-field masks; SigLIP coordinates follow the actual patch convolution kernel, stride, padding, and processor size, including uncovered edge strips where applicable. Trials whose Grad-CAM map is identically zero after retaining positive evidence are excluded and counted explicitly; such maps cannot support a spatial correlation. These spatial readouts use one caption, whereas the embedding choice survey retains ten-caption averaging.

\paragraph{Common evaluation and references.}
For each trial, sample-density gaze maps are smoothed with $\sigma=40$ screen pixels, integrated over each model's native token regions, and normalized within each valid image with the same target construction as in fine-tuning. Spearman correlation is computed separately within images A and B, averaged over valid image correlations within trial, and then over trials with a defined correlation. Constant maps yield undefined correlations. The center reference (the center baseline of Section~\ref{sec:references}) is an isotropic Gaussian with $\sigma=0.125$ in normalized composite-panel coordinates, centered at $(0.25,0.5)$ for A and $(0.75,0.5)$ for B. Its mass is integrated over the same cells and evaluated with the same human masks; valid image and correlation denominators are reported alongside every summary. Relative mass on A/B and the gap cannot determine this within-image statistic.

As a consistency check, the 1,030 \Test{} trials of each small-model run reproduce the zero-shot rows of Table~\ref{tab:main}: the same trial identities, human masks, 2,028 valid images, valid-correlation counts, and metrics within a prespecified tolerance of $0.0005$ (Gemma4-E4B $\rho=.349$, $\Delta$KL $=-2.398$ bits; Qwen3-VL-8B $.032$, $-3.776$; Molmo2-8B $.010$, $-5.548$). These \Test{} values differ from the full-cohort pseudo-word means in Table~\ref{tab:human_model_attention_alignment_combined} because the latter use all splits.

\paragraph{Completed embedding grids and denominators.}
Table~\ref{tab:embedding_spatial_denominators} lists the denominators for the pseudo-word entries in Table~\ref{tab:human_model_attention_alignment_combined}. CLIP RN50 excludes 42 pseudo-word trials with zero positive-evidence maps. Of the remaining 7,099 trials, 22 have no defined model correlation in either valid image, leaving 7,077 trials for model $\rho$; Center remains defined on 7,099 trials. Undefined image-level correlations are omitted before the within-trial average, so model and Center image counts can also differ (including seven images for SigLIP2-so400m). These are descriptive means with the displayed denominators, not a paired significance test against Center.

\begin{table}[t]
\centering\footnotesize
\setlength{\tabcolsep}{4pt}
\caption{Native grids and defined-correlation denominators for the embedding pseudo-word spatial results. Counts follow zero-map exclusions; image-level values are averaged within each trial before averaging trials.}
\label{tab:embedding_spatial_denominators}
\begin{tabular}{@{}llrrrr@{}}
\toprule
 & & \multicolumn{2}{c}{Trials} & \multicolumn{2}{c}{Images} \\
\cmidrule(lr){3-4}\cmidrule(lr){5-6}
Model & Grid & $\rho$ & Center & $\rho$ & Center \\
\midrule
CLIP RN50 & $7\times7$ & 7,077 & 7,099 & 13,233 & 13,987 \\
CLIP ViT-B/32 & $7\times7$ & 7,141 & 7,141 & 14,070 & 14,070 \\
SigLIP-base & $14\times14$ & 7,141 & 7,141 & 14,070 & 14,070 \\
SigLIP-so400m & $27\times27$ & 7,141 & 7,141 & 14,070 & 14,070 \\
SigLIP2-base & $14\times14$ & 7,141 & 7,141 & 14,070 & 14,070 \\
SigLIP2-so400m & $32\times32$ & 7,141 & 7,141 & 14,063 & 14,070 \\
\bottomrule
\end{tabular}
\end{table}

\paragraph{Input resolution.}
To check sensitivity to input size, we compared the two large models at $512\times512$ with a $1000\times1000$ input on 500 stratified pseudo-word inputs (539 human trials).
For Qwen3-VL-30B, ICA changed by $-0.93$ pp and human-choice agreement by $+0.56$ pp, and input-level $P(\mathrm{round})$ correlated at Spearman .828 across resolutions; for Gemma4-26B, the changes were $+2.04$ and $+2.23$ pp, with correlation .892.
Both changes are within a tolerance of 3 pp fixed before the comparison.
The $1000\times1000$ outputs use native BF16 logits rather than the FP32 A/B projection, so these differences bound the combined effect of resolution and numerical backend rather than isolating resolution.

\paragraph{Generative results and denominators.}
For all six generative models, each pseudo-word spatial mean uses 7,141 trials (6,720 unique model inputs, 162 labels) and 14,070 valid image-level correlations for both the model and Center, so the two denominators match.
Qwen and Gemma use $16\times16$ grids; both Molmo models use the combined 725-token global and high-resolution geometry.
The model/Center means are .340/.637 (Gemma4-E4B), .265/.637 (Gemma4-26B), .022/.637 (Qwen3-VL-8B), $-.003$/.637 (Qwen3-VL-30B), $-.074$/.601 (Molmo2-4B), and .004/.601 (Molmo2-8B).
For the three fine-tuned models, the \Test{} subset of these runs reproduces the zero-shot rows of Table~\ref{tab:main} (consistency check above); Molmo2-4B is not fine-tuned, so this check does not apply to it.
Paired comparisons against Center for all six models are reported below.

\section{Probability-based Analysis}
\label{sec:appendix_prob_analysis}

Tables~\ref{tab:prob-overall}, \ref{tab:prob-pseudo}, and \ref{tab:prob-adjective} report the probability-based analysis following \citet{kouwenhovencross}.
This method produces a model choice for each image by comparing the probabilities assigned to candidate labels.
However, it differs substantially from the human forced-choice task: human participants choose between two images for a single label, whereas the probability-based method chooses among all labels for a given image.
As a result, this analysis is not directly comparable to the human study.
We therefore use it only to report ICA, rather than model--human agreement.

This distinction is important because the probability-based method can introduce label-side biases.
In particular, we observe a strong tendency toward sharp labels.
Even when overall accuracy is close to chance, round and sharp accuracy can be highly asymmetric.
If a model is slightly more likely to choose sharp labels regardless of the image, then sharp-label accuracy will be inflated while round-label accuracy will be correspondingly reduced.
Such an asymmetry does not necessarily indicate human-aligned sensitivity to the round--sharp contrast; instead, it may reflect a bias induced by the label-selection procedure.
For this reason, we treat the probability-based results as a diagnostic for the ICA rather than as an approximation of human forced-choice behavior.

\newcommand{\accCI}[2]{\begin{tabular}{@{}c@{}}#1\\{\scriptsize #2}\end{tabular}}
 
\begin{table}[t]
\centering
\scriptsize
\setlength{\tabcolsep}{3pt}
\renewcommand{\arraystretch}{1.18}
\caption{Probability-based top-choice shape-match accuracy, overall, following \citet{kouwenhovencross}. Values in brackets are Wilson 95\% confidence intervals.}
\begin{tabular}{l ccc}
\hline
Caption & Total & Round & Sharp \\
\hline
CLIP (RN50)
& \accCI{0.712}{[0.702, 0.722]}
& \accCI{0.644}{[0.629, 0.660]}
& \accCI{0.780}{[0.766, 0.793]} \\
CLIP (ViT)
& \accCI{\bfseries 0.720}{[0.709, 0.730]}
& \accCI{\bfseries 0.701}{[0.686, 0.715]}
& \accCI{0.739}{[0.724, 0.753]} \\
SigLIP-base
& \accCI{0.521}{[0.510, 0.533]}
& \accCI{0.464}{[0.448, 0.480]}
& \accCI{0.579}{[0.563, 0.595]} \\
SigLIP-so400m
& \accCI{0.635}{[0.624, 0.646]}
& \accCI{0.467}{[0.451, 0.483]}
& \accCI{0.804}{[0.790, 0.816]} \\
SigLIP2-base
& \accCI{0.516}{[0.504, 0.527]}
& \accCI{0.142}{[0.131, 0.153]}
& \accCI{\bfseries 0.890}{[0.879, 0.900]} \\
SigLIP2-so400m
& \accCI{0.511}{[0.500, 0.523]}
& \accCI{0.345}{[0.329, 0.360]}
& \accCI{0.678}{[0.662, 0.693]} \\
\hline
\end{tabular}
\label{tab:prob-overall}
\end{table}

\begin{table}[t]
\centering
\scriptsize
\setlength{\tabcolsep}{3pt}
\renewcommand{\arraystretch}{1.18}
\caption{Probability-based top-choice shape-match accuracy, pseudo-word captions only. Values in brackets are Wilson 95\% confidence intervals}
\begin{tabular}{l ccc}
\hline
Caption & Total & Round & Sharp \\
\hline
CLIP (RN50)
& \accCI{0.552}{[0.536, 0.568]}
& \accCI{0.537}{[0.515, 0.560]}
& \accCI{0.566}{[0.543, 0.588]} \\
CLIP (ViT)
& \accCI{0.583}{[0.567, 0.599]}
& \accCI{\bfseries 0.666}{[0.644, 0.687]}
& \accCI{0.499}{[0.477, 0.522]} \\
SigLIP-base
& \accCI{0.523}{[0.507, 0.539]}
& \accCI{0.291}{[0.270, 0.312]}
& \accCI{0.756}{[0.736, 0.775]} \\
SigLIP-so400m
& \accCI{\bfseries 0.604}{[0.588, 0.619]}
& \accCI{0.573}{[0.550, 0.595]}
& \accCI{0.634}{[0.612, 0.656]} \\
SigLIP2-base
& \accCI{0.532}{[0.515, 0.548]}
& \accCI{0.283}{[0.263, 0.304]}
& \accCI{\bfseries 0.780}{[0.760, 0.798]} \\
SigLIP2-so400m
& \accCI{0.513}{[0.497, 0.529]}
& \accCI{0.366}{[0.345, 0.389]}
& \accCI{0.660}{[0.638, 0.682]} \\
\hline
\end{tabular}
\label{tab:prob-pseudo}
\end{table}

\begin{table}[t]
\centering
\scriptsize
\setlength{\tabcolsep}{3pt}
\renewcommand{\arraystretch}{1.18}
\caption{Probability-based top-choice shape-match accuracy, adjective captions only. Values in brackets are Wilson 95\% confidence intervals}
\begin{tabular}{l ccc}
\hline
Caption & Total & Round & Sharp \\
\hline
CLIP (RN50)
& \accCI{\bfseries 0.872}{[0.861, 0.883]}
& \accCI{\bfseries 0.751}{[0.731, 0.770]}
& \accCI{\bfseries 0.993}{[0.989, 0.996]} \\
CLIP (ViT)
& \accCI{0.857}{[0.845, 0.868]}
& \accCI{0.735}{[0.715, 0.755]}
& \accCI{0.978}{[0.971, 0.984]} \\
SigLIP-base
& \accCI{0.519}{[0.503, 0.535]}
& \accCI{0.636}{[0.614, 0.658]}
& \accCI{0.402}{[0.380, 0.425]} \\
SigLIP-so400m
& \accCI{0.667}{[0.651, 0.682]}
& \accCI{0.361}{[0.339, 0.383]}
& \accCI{0.973}{[0.964, 0.979]} \\
SigLIP2-base
& \accCI{0.856}{[0.844, 0.867]}
& \accCI{0.744}{[0.724, 0.763]}
& \accCI{0.968}{[0.959, 0.975]} \\
SigLIP2-so400m
& \accCI{0.509}{[0.493, 0.525]}
& \accCI{0.323}{[0.302, 0.345]}
& \accCI{0.695}{[0.674, 0.716]} \\
\hline
\end{tabular}
\label{tab:prob-adjective}
\end{table}

\section{Choice agreement and majority-vote reference}
\label{app:choice_agreement_intervals}
Table~\ref{tab:choice_score_intervals} reports individual 95\% confidence intervals for the two percentage-valued scores in Table~\ref{tab:main}: ICA and human-choice agreement.
The human majority-vote reference predicts each participant's response from the other participants' responses to the same word on \Test{}; ties receive an expected agreement score of .5.
All rows use the same 1{,}030 test trials.

Model score intervals are percentile intervals from 20{,}000 word$\times$pair$\times$participant product-cluster bootstrap draws, with the selected checkpoints held fixed.
The reference score and paired model--reference differences use 20{,}000 draws with the original majority-vote predictions held fixed; their intervals are centered on the observed estimate using the 95th percentile of absolute bootstrap errors.
Within each analysis, resampling weights are shared across conditions.
All intervals are individual and unadjusted; they exclude uncertainty from training runs, checkpoint selection, and, for the reference, estimation of the majority-vote predictions.

\begin{table}[htbp]
\centering\footnotesize
\setlength{\tabcolsep}{4pt}
\caption{Individual ICA and human-choice agreement scores (\%) with 95\% confidence intervals on \Test{}. These are intervals for each score, not for differences between scores. ICA measures agreement with the designed category; human-choice agreement measures agreement with the observed participant response. The reference row reports the human majority-vote predictor's agreement with participant responses. Model and reference intervals use the respective bootstrap constructions described above.}
\label{tab:choice_score_intervals}
\begin{tabular}{llll}
\toprule
Model / reference & Condition & ICA (\%) [95\% CI] & Human agr.\ (\%) [95\% CI] \\
\midrule
\gemma & \condZ & 53.69 [44.95, 62.21] & 52.91 [44.90, 61.16] \\
 & \condC & 92.72 [86.36, 97.33] & 72.72 [64.13, 80.61] \\
 & \condG & 92.04 [85.45, 96.92] & 72.82 [64.78, 80.45] \\
 & \condA & 94.37 [88.45, 98.34] & 73.01 [64.90, 80.60] \\
 & \condS & 94.56 [88.86, 98.42] & 72.62 [64.39, 80.39] \\
\midrule
\qwen & \condZ & 49.90 [41.56, 58.16] & 48.54 [40.43, 56.50] \\
 & \condC & 94.08 [88.16, 98.18] & 72.91 [64.49, 80.62] \\
 & \condG & 92.43 [86.25, 96.99] & 72.82 [64.41, 80.46] \\
 & \condA & 89.61 [81.87, 95.50] & 71.75 [63.32, 79.50] \\
 & \condS & 90.97 [84.29, 96.00] & 71.94 [63.35, 79.74] \\
\midrule
\molmo & \condZ & 53.50 [44.86, 61.99] & 50.00 [41.90, 57.94] \\
 & \condC & 92.52 [86.66, 96.89] & 71.94 [63.45, 79.76] \\
 & \condG & 89.03 [81.36, 95.00] & 73.11 [65.00, 80.67] \\
 & \condA & 89.71 [82.42, 95.52] & 72.43 [64.21, 80.15] \\
 & \condS & 89.22 [81.52, 95.30] & 72.72 [64.52, 80.43] \\
\midrule
Human majority-vote & --- & --- & 72.86 [64.76, 80.96] \\
\bottomrule
\end{tabular}
\end{table}

Table~\ref{tab:choice_majority_contrasts} separately reports the paired differences between \condC{} and the human majority-vote reference.
These intervals are computed directly from differences on matched trials, rather than by subtracting the bounds of the individual score intervals.
They describe uncertainty in the comparison and do not establish equivalence or a ceiling on achievable agreement.

\begin{table}[htbp]
\centering\footnotesize
\caption{Paired differences in human-choice agreement between \condC{} and the human majority-vote reference, in percentage points. Positive values favor the model. Individual 95\% intervals are conditional on the fixed model and reference predictions.}
\label{tab:choice_majority_contrasts}
\begin{tabular}{lrr}
\toprule
Model & Model minus reference (pp) & 95\% CI (pp) \\
\midrule
\gemma & $-0.15$ & $[-5.29, +5.00]$ \\
\qwen & $+0.05$ & $[-4.80, +4.90]$ \\
\molmo & $-0.92$ & $[-6.20, +4.36]$ \\
\bottomrule
\end{tabular}
\end{table}

\section{Gaze controls and allocation comparisons}
\label{app:gaze_controls}

\subsection{Training details}
\label{sec:optim}\label{app:training_details}

\paragraph{Optimization and checkpoint selection.}
All conditions train LoRA adapters (rank 16, scaling 32, dropout .05) on the language decoder's attention and MLP projections, with the vision encoder and projector frozen, using AdamW ($2\times10^{-5}$, 5\% warmup, cosine decay, effective batch 8, bf16) for at most six epochs of 242 updates with patience two, at a single optimization seed and a single gaze-loss weight.
Every condition selects its checkpoint by \Val{} agreement with the actual human choice, never by gaze fit or test performance. 

\paragraph{Gaze targets.}
We construct sample-density gaze maps from the verified image-display interval using Gaussian smoothing ($\sigma=40$ screen pixels), integrate them over each model's image-token regions, and normalize separately within each valid image.
For \condG, $H_{t,i}$ is the map of the participant who supplied trial $t$'s choice; for \condA, it is replaced by the fixed \Train{} average for the same image position $i\in\{A,B\}$, with the original trial masks retained.
\condS{} instead jointly normalizes the human token masses and the readout over the union of both images, excluding boundary-crossing tokens.
Its joint KL decomposes into a KL between the two-image mass distributions and within-image KL terms weighted by the corresponding human mass.
Unlike \condG{}, \condS{} does not mask an image solely because it contains no raw gaze samples; any mass contributed by Gaussian smoothing is retained.

\paragraph{Loss details.}
The gaze term is zero if no image is valid; the choice loss is retained, and the training pool has at least one valid image per trial.
KL uses natural logarithms during training, with zero human bins contributing zero.
Because gaze mass and choice coincide on most trials, \condS{} partly re-supplies the choice label through the gaze channel; a choice gain under \condS{} would therefore not isolate gaze information beyond the label, whereas its absence is the stronger result (\cref{sec:res_gaze_choice}).
 
\paragraph{Alignment measures.}
We evaluate every condition on \Test{} along three groups of measures.
\emph{Choice:} \humanagr{} against the actual choice in the same trial, and mean human-response negative log-likelihood (NLL) from the saved A/B probabilities (\cref{sec:protocol}).
NLL averages $-\log_2 p(y_{\mathrm{human}})$ across trials, where $p(y_{\mathrm{human}})$ is the probability assigned to the participant's actual response; lower values indicate better probability predictions, and assigning A and B equal probability gives 1 bit.
\emph{Within-image distribution:} \rhoin{} and \gazeKL{}.
We compute KL using base-2 logarithms and report \dKL{} in bits. Image-level values are averaged within each trial and then across trials; positive \dKL{} indicates lower divergence than the fixed training-mean map (\cref{eq:delta_kl}).
Equation~\ref{eq:delta_kl} defines the KL reduction relative to the train-mean reference.
KL matches the direction of the supervised objective.
To assess whether the findings extend to other aspects of spatial alignment, we also compare Jensen--Shannon divergence (JS; symmetric distribution mismatch),  histogram intersection (overlap in probability mass), and center-of-mass distance (separation between the maps' spatial centers).
Lower JS and center-of-mass distance, and higher histogram intersection, indicate closer alignment.
\emph{Allocation between images:} 
We measure how human gaze and model attention are distributed between images A and B.
First, the \sideJS{} compares the proportions of total gaze and attention assigned to the two images; lower values indicate more similar allocation.
We also report two agreement rates:
\cattn{}, whether the model chooses the image receiving more model attention, and \gattn{}, whether the image receiving more model attention is also the one receiving more human gaze.
These agreement rates can be high simply because both quantities favor the same side across trials.
For each rate, we therefore compute the agreement expected if the two quantities were independent, using their respective overall A/B frequencies:
\begin{equation}
E = p_B\,q_B + (1-p_B)(1-q_B),
\label{eq:side_expected}
\end{equation}
where $p_B$ and $q_B$ are the rates at which the two compared quantities (e.g., the model's choice and the image with more model attention) favor B.
The observed-minus-expected gap indicates whether they agree more often than these side preferences alone would predict. 
It does not establish that attention to an image causes the model to choose it.

\subsection{Reference maps and uncertainty}
\label{sec:app_ref_maps_and_uncert}
\paragraph{Reference maps.}
Alongside the train-mean map in \cref{eq:delta_kl}, we use the two gaze references of Section~\ref{sec:references}: a center baseline (the Gaussian defined in Appendix~\ref{sec:appendix_human_model_alignment_methods}, centered within each image) and a population-mean map (the average gaze map of the other \Test{} trials, computed separately for A and B, excluding the evaluated trial).
These assess alignment with a central spatial prior and a shared gaze pattern, respectively.
Reference maps use each model's native image-token geometry, with direct comparisons evaluated on the same valid images; they are empirical references, not performance ceilings.

\paragraph{Uncertainty.}
For model-condition comparisons, we use paired word$\times$pair$\times$participant cluster bootstraps with shared trial weights across conditions.
The 95\% confidence intervals are individual and unadjusted, conditional on the fitted seed and selected checkpoints; they do not capture variability from retraining or checkpoint selection.
We apply Holm correction to exploratory p-values separately within two families: the 18 \condG--\condA{} comparisons (three models, six measures) and the nine choice-NLL comparisons among \condC, \condG, and \condA{} (three models, three condition pairs).

\subsection{Trial-specific versus average-gaze supervision}
\label{app:gaze_mean}
The within-image correlations under \condG{} (.733/.729/.673) and \condA{} (.733/.731/.679) are close to the population-mean reference (.731/.731/.680) on the corresponding native token geometries.
A fixed training-average target thus reproduces high correlation without trial-specific gaze targets, although similar correlations do not establish statistical equivalence.
Table~\ref{tab:gaze_kl_levels} reports mean gaze KL under \condA{} and \condG{}, providing the denominator for the relative reductions of 7.64/6.97/6.84\%.
Here, ``mean gaze KL'' denotes image-level KL averaged within and then across trials; the denominator is the KL of the \condA{} model's attention maps, not the KL of the fixed train-mean target or its $\Delta$KL value.
Improvements are not uniform across metrics: for \molmo{}, within-image correlation is higher under \condA{}, while KL point estimates favor \condG{}.
Because \condA{} uses one map per image position, the remaining \condG{} advantage in KL may reflect image-specific gaze structure shared across participants rather than information specific to individual trials; a per-image average target would separate the two, and these single-seed comparisons do not.

\begin{table}[htbp]
\centering\footnotesize
\setlength{\tabcolsep}{4pt}
\caption{Mean gaze KL under \condA{} and \condG{} and the reduction relative to \condA{}. KL is computed per valid image, averaged within each trial and then across trials, as in \cref{sec:measures}. Relative reduction is $100\times(\mathrm{KL}_{\text{\condA}}-\mathrm{KL}_{\text{\condG}})/\mathrm{KL}_{\text{\condA}}$, calculated from unrounded values. These are point estimates; paired intervals and significance tests are in Table~\ref{tab:gaze_control_contrasts}.}
\label{tab:gaze_kl_levels}
\begin{tabular}{lrrrr}
\toprule
Model & \condA{} KL (bits) & \condG{} KL (bits) & Reduction (bits) & Reduction (\%) \\
\midrule
\gemma & 1.0624 & 0.9812 & 0.0812 & 7.64 \\
\qwen & 1.0608 & 0.9869 & 0.0739 & 6.97 \\
\molmo & 1.1211 & 1.0444 & 0.0767 & 6.84 \\
\bottomrule
\end{tabular}
\end{table}

Table~\ref{tab:gaze_control_contrasts} reports all 18 exploratory \condG--\condA{} comparisons on the full \Test{} set (1{,}030 trials; 2{,}028 valid images).
The two-sided tests use centered errors from 1{,}000{,}000 paired word$\times$pair$\times$participant product-cluster bootstrap resamples, with weights shared across models and metrics.
Holm correction is applied across the 18 tests; the centered 95\% intervals are individual and unadjusted.
The selected checkpoints are held fixed, so these results exclude training-seed and checkpoint-selection uncertainty.

The stars in Table~\ref{tab:main} refer specifically to the \condG--\condA{} contrast in $\Delta$KL.
Since both conditions use the same fixed train-mean reference, it cancels in this difference:
\[
\Delta\operatorname{KL}_{\text{\condG}}-\Delta\operatorname{KL}_{\text{\condA}}
=\operatorname{KL}(H\,\Vert\,A_{\text{\condA}})
-\operatorname{KL}(H\,\Vert\,A_{\text{\condG}}),
\]
where $A_{\text{\condG}}$ and $A_{\text{\condA}}$ denote the two models' attention maps, with the same averaging as in \cref{eq:delta_kl}.
Thus, the starred values indicate lower gaze KL than the \condA{} model, not significance against the fixed train-mean map or a test of $\Delta$KL against zero.
Markers are assigned only to the three \condG{} rows for this contrast.

\begin{table}[htbp]
\centering\footnotesize
\setlength{\tabcolsep}{4pt}
\caption{Exploratory \condG--\condA{} differences with individual 95\% intervals, two-sided bootstrap $p$-values, and Holm-adjusted $p$-values across all 18 tests. Positive differences favor \condG{} for human agreement, $\rho$, histogram intersection, and $\Delta$KL; negative differences favor \condG{} for JS and center-of-mass distance. Human agreement is in percentage points and $\Delta$KL in bits.}
\label{tab:gaze_control_contrasts}
\resizebox{\linewidth}{!}{%
\begin{tabular}{llrrr}
\toprule
Model & Measure & Difference [95\% CI] & $p$ & $p_{\mathrm{Holm}}$ \\
\midrule
\gemma & Human agreement (pp) & -0.19 [-3.89, +3.50] & 0.9145 & 1.0000 \\
 & Gaze $\rho$ & -0.0004 [-0.0035, +0.0027] & 0.8012 & 1.0000 \\
 & Gaze JS & -0.0147 [-0.0215, -0.0079] & 0.0003 & 0.0048 \\
 & Histogram intersection & +0.0232 [+0.0128, +0.0335] & 0.0001 & 0.0023 \\
 & Center-of-mass distance & -0.0034 [-0.0054, -0.0014] & 0.0020 & 0.0286 \\
 & $\Delta$KL (bits) & +0.0812 [+0.0298, +0.1326] & 0.0040 & 0.0478 \\
\midrule
\qwen & Human agreement (pp) & +1.07 [-3.41, +5.54] & 0.6063 & 1.0000 \\
 & Gaze $\rho$ & -0.0022 [-0.0045, +0.0000] & 0.0504 & 0.2520 \\
 & Gaze JS & -0.0105 [-0.0171, -0.0039] & 0.0034 & 0.0447 \\
 & Histogram intersection & +0.0176 [+0.0074, +0.0278] & 0.0014 & 0.0211 \\
 & Center-of-mass distance & -0.0028 [-0.0049, -0.0008] & 0.0083 & 0.0660 \\
 & $\Delta$KL (bits) & +0.0739 [+0.0254, +0.1223] & 0.0048 & 0.0497 \\
\midrule
\molmo & Human agreement (pp) & +0.68 [-2.92, +4.28] & 0.6778 & 1.0000 \\
 & Gaze $\rho$ & -0.0061 [-0.0089, -0.0034] & 0.0003 & 0.0048 \\
 & Gaze JS & -0.0096 [-0.0170, -0.0022] & 0.0133 & 0.0801 \\
 & Histogram intersection & +0.0158 [+0.0049, +0.0266] & 0.0062 & 0.0557 \\
 & Center-of-mass distance & -0.0031 [-0.0052, -0.0011] & 0.0045 & 0.0497 \\
 & $\Delta$KL (bits) & +0.0767 [+0.0202, +0.1331] & 0.0109 & 0.0762 \\
\bottomrule
\end{tabular}}
\end{table}

\subsection{Allocation between images}
\label{app:gaze_side}
Similar choice scores can coexist with different choice--attention relationships (Table~\ref{tab:main}).
Under \condC{}, \cattn{} is 93.11/59.32/47.48\%, with gaps over independence of $+43.12$/$+9.59$/$+0.51$ pp.
For \molmo{} under \condZ{}, the observed 80.68\% agreement is below the 82.54\% expected from its shared side preferences (\cref{eq:side_expected}).
These associations describe the measured readout and do not identify which image regions caused a choice.
 
For \gemma{}, \condG{} and \condA{} both round to $\rho=.733$, yet \cattn{} is 92.14\% versus 53.20\% (gaps over independence: $+42.11$ versus $+3.83$ pp).
Their \gattn{} is 64.37\% versus 46.21\%, a difference of $+18.16$ pp [7.40, 29.03].
Under \condA{}, attention favors B on 96.12\% of trials in this run.
The corresponding \sideJS{} difference is $-.00311$ [$-.00870$, $+.00250$] nats, so the directional-agreement result does not extend to every measure of allocation.
For \qwen{} and \molmo{}, both \condG{} and \condA{} increase \cattn{} relative to \condC{}; this change is therefore not unique to trial-specific gaze supervision.

\condS{} reduces \sideJS{} relative to \condC{} in all three models, with individual paired intervals excluding zero (Table~\ref{tab:full_side_contrasts}).
Relative to \condG{}, only \gemma{} has an interval entirely below zero.
These allocation improvements are distinct from the uncertain choice differences reported in \cref{sec:res_gaze_choice}.
The side comparisons use 20{,}000 paired word$\times$pair$\times$participant product-cluster bootstrap resamples, with shared weights across conditions and individual, unadjusted 95\% percentile intervals conditional on the selected checkpoints.
They are not included in the 18-test Holm family above.

\begin{table}[htbp]
\centering\footnotesize
\setlength{\tabcolsep}{4pt}
\caption{Paired differences in Side JS (nats), with individual 95\% intervals. Negative values indicate closer alignment with the human two-image mass split under \condS{}. These intervals are unadjusted and conditional on a single fitted seed.}
\label{tab:full_side_contrasts}
\begin{tabular}{lrr}
\toprule
Model & \condS{} minus \condC{} [95\% CI] & \condS{} minus \condG{} [95\% CI] \\
\midrule
\gemma & -0.00805 [-0.01350, -0.00328] & -0.00368 [-0.00664, -0.00094] \\
\qwen & -0.02213 [-0.02997, -0.01477] & -0.00052 [-0.00237, +0.00127] \\
\molmo & -0.04174 [-0.05647, -0.02834] & -0.00078 [-0.00201, +0.00035] \\
\bottomrule
\end{tabular}
\end{table}

\section{Position-swap evaluation}
\label{app:swap}
We evaluate the 15 small-model conditions and two additional larger zero-shot models on the same 1{,}030 \Test{} trials.
For each trial, we exchange the two image panels without mirroring either panel, keep the word and the A/B position labels fixed, and match the original and exchanged predictions by trial.
Both layouts use the same $512\times512$ input size, prompt, and greedy decoding protocol.
Responses are accepted as a JSON choice, an exact single A/B, or a whole JSON object enclosed in a code fence; otherwise, we use the A/B probability fallback.
The same response rule is applied to all 17 conditions and both layouts.
 
Swap consistency is the fraction of trials selecting the same shape in both layouts, which requires opposite A/B answers after the exchange.
An always-A or always-B strategy scores 0\%; independent uniform random answers have expected consistency of 50\%.
Position-balanced ICA is the mean of original-layout and exchanged-layout ICA, with the intended side reversed in the exchanged layout.
Each original trial has equal weight, and its two layouts contribute equally; the two layouts are not treated as independent trials.
These controls assess sensitivity to position: even perfect swap consistency does not establish a word-dependent association, and averaging layouts does not remove every possible shortcut.
The larger models are additional zero-shot references; comparisons with the smaller models do not isolate the effect of parameter count.

Table~\ref{tab:swap_all_conditions} reports swap consistency and position-balanced ICA for all conditions.
Table~\ref{tab:swap_human_side} additionally reports position-balanced agreement with the original human choice, together with the model's B-choice rate in each layout.
For the exchanged layout, the human reference is the originally selected shape moved to its new position; it is not a newly observed human response to the exchanged stimulus.
We do not evaluate human-gaze alignment in the exchanged layout because no human gaze was recorded for that presentation.
 
All intervals use 20{,}000 product-cluster bootstrap resamples over words, image pairs, and participants (seed 20260921), with weights shared across conditions and across the two layouts of each trial.
The reported intervals are individual, unadjusted 95\% percentile intervals, conditional on the fixed checkpoints; they exclude training-seed and checkpoint-selection uncertainty.
 
Under \condC{}, swap consistency rises to 90.10/96.89/88.45\% (from 30.97/8.25/19.51\% under \condZ{}), while position-balanced ICA reaches 92.72/94.56/91.50\%.
These results show that the learned choices largely persist across positions and cannot be explained by always choosing A or always choosing B.
The two larger zero-shot models, Gemma4-26B and Qwen3-VL-30B, reach swap consistency of 57.38/54.56\% and position-balanced ICA of 61.99/60.10\%, respectively.
Swap consistency alone does not establish a word-dependent shape association.

\begin{table}[htbp]
\centering\footnotesize
\setlength{\tabcolsep}{4pt}
\caption{Position-swap results for all 17 conditions on \Test{}. Values are percentages with individual 95\% confidence intervals. Balanced ICA averages intended-category agreement across the original and exchanged layouts.}
\label{tab:swap_all_conditions}
\begin{tabular}{llrr}
\toprule
Model & Condition & Swap consistency [95\% CI] & Balanced ICA [95\% CI] \\
\midrule
\gemma & \condZ & 30.97 [20.97, 41.73] & 52.48 [47.34, 57.86] \\
 & \condC & 90.10 [82.76, 95.66] & 92.72 [87.39, 96.72] \\
 & \condG & 93.01 [87.26, 97.20] & 91.84 [85.76, 96.46] \\
 & \condA & 94.95 [89.43, 98.63] & 94.76 [89.70, 98.26] \\
 & \condS & 94.66 [88.92, 98.43] & 94.42 [89.25, 98.05] \\
\midrule
\qwen & \condZ & 8.25 [2.65, 15.88] & 51.31 [48.70, 54.61] \\
 & \condC & 96.89 [93.06, 99.30] & 94.56 [88.97, 98.32] \\
 & \condG & 89.22 [81.30, 95.42] & 92.18 [86.79, 96.43] \\
 & \condA & 91.26 [85.09, 95.97] & 90.39 [83.46, 95.70] \\
 & \condS & 87.38 [79.46, 93.70] & 91.36 [85.92, 95.68] \\
\midrule
\molmo & \condZ & 19.51 [10.90, 29.37] & 54.22 [49.91, 58.97] \\
 & \condC & 88.45 [80.57, 94.58] & 91.50 [85.92, 95.90] \\
 & \condG & 82.72 [72.77, 91.07] & 87.57 [80.68, 93.35] \\
 & \condA & 82.91 [72.99, 91.34] & 88.35 [81.65, 94.06] \\
 & \condS & 84.37 [74.69, 92.43] & 88.30 [81.40, 94.16] \\
\midrule
Gemma4-26B-A4B & \condZ & 57.38 [46.33, 67.85] & 61.99 [54.48, 69.33] \\
\midrule
Qwen3-VL-30B-A3B & \condZ & 54.56 [43.74, 65.16] & 60.10 [53.04, 67.10] \\
\bottomrule
\end{tabular}
\end{table}

\begin{table}[htbp]
\centering\footnotesize
\setlength{\tabcolsep}{3pt}
\caption{Position-balanced human-reference agreement and B-choice rates for all 17 conditions. Values are percentages with individual 95\% confidence intervals. Human-reference agreement uses the participant's originally selected shape in both layouts; the exchanged-layout reference is not a new human observation. Original B and swapped B are the model's B-choice rates before and after exchanging the images.}
\label{tab:swap_human_side}
\resizebox{\linewidth}{!}{%
\begin{tabular}{llrrr}
\toprule
Model & Condition & Balanced human ref. & Original B & Swapped B \\
\midrule
\gemma & \condZ & 51.99 [47.39, 56.80] & 18.74 [11.42, 27.42] & 13.40 [7.63, 20.42] \\
 & \condC & 72.62 [64.52, 80.14] & 45.15 [36.67, 53.48] & 46.31 [38.11, 54.77] \\
 & \condG & 72.52 [64.64, 79.89] & 49.13 [40.92, 57.24] & 49.71 [41.73, 57.90] \\
 & \condA & 72.82 [64.75, 80.26] & 49.32 [41.14, 57.20] & 49.71 [41.63, 57.99] \\
 & \condS & 72.48 [64.49, 80.02] & 47.96 [39.68, 56.00] & 48.25 [40.17, 56.63] \\
\midrule
\qwen & \condZ & 50.92 [48.65, 53.71] & 96.12 [91.59, 99.12] & 95.63 [90.66, 98.96] \\
 & \condC & 72.91 [64.59, 80.55] & 48.83 [40.77, 56.79] & 50.00 [42.06, 57.99] \\
 & \condG & 72.18 [64.28, 79.54] & 45.44 [37.19, 53.52] & 45.15 [36.80, 53.73] \\
 & \condA & 71.75 [63.79, 79.21] & 51.94 [43.76, 59.88] & 52.14 [43.96, 60.38] \\
 & \condS & 72.62 [64.77, 79.89] & 56.21 [47.82, 64.26] & 56.02 [47.79, 64.29] \\
\midrule
\molmo & \condZ & 52.28 [48.40, 56.54] & 89.81 [83.25, 95.11] & 90.10 [83.66, 95.20] \\
 & \condC & 71.02 [62.98, 78.63] & 46.70 [37.92, 55.48] & 47.77 [39.25, 56.45] \\
 & \condG & 71.55 [64.00, 78.72] & 41.84 [33.18, 50.46] & 41.84 [33.26, 50.79] \\
 & \condA & 71.46 [63.82, 78.73] & 42.14 [33.62, 50.85] & 41.55 [32.99, 50.61] \\
 & \condS & 71.70 [63.92, 79.08] & 42.43 [33.69, 51.17] & 42.72 [34.20, 51.59] \\
\midrule
Gemma4-26B & \condZ & 56.84 [50.33, 63.36] & 29.03 [21.16, 37.44] & 30.29 [22.08, 39.27] \\
\midrule
Qwen3-VL-30B & \condZ & 57.86 [51.65, 64.24] & 69.32 [60.21, 77.68] & 68.54 [59.95, 76.62] \\
\bottomrule
\end{tabular}}
\end{table}

\end{document}